\documentclass[lettersize,journal]{IEEEtran}
\usepackage{amsmath,amsfonts}
\usepackage{algorithmic}
\usepackage{algorithm}
\usepackage{array}
\usepackage{color}
\usepackage{textcomp}
\usepackage{stfloats}
\usepackage{url}
\usepackage{verbatim}
\usepackage[caption=false,font=normalsize,labelfont=sf,textfont=sf]{subfig}
\usepackage{graphicx}
\usepackage{cite}
\usepackage{rotating}
\usepackage{arydshln}
\usepackage{mathrsfs}
\usepackage{booktabs}  
\usepackage{array}     
\usepackage{adjustbox} 
\usepackage{hyperref}
\hypersetup{hypertex=true,
colorlinks=true,
linkcolor=blue,
anchorcolor=blue,
citecolor=blue}
\usepackage{booktabs}
\usepackage{bbding}
\usepackage{amssymb}
\usepackage{pifont}
\usepackage{array}
\newcolumntype{C}[1]{>{\centering\arraybackslash}p{#1}}
\newcommand{\PanelLabel}{}
\usepackage{xcolor}
\usepackage{colortbl}
\usepackage{multirow}

\begin{document}

\title{Degradation-Aware Prompt Learning with Cross-Modal Compensation for\\ Adverse Weather Removal
}

\author{\
Wanshu Fan, Yunzhe Zhang, Yue Shen, Liyan Wang, Jing Qin, Kin-Man Lam, Cong Wang, Jinshan Pan\

\thanks{
This work was supported in part by the National Natural Science Foundation of China (Grant No. 62502064); Liaoning Provincial Key Research and Development Joint Program (Grant No. 2025110219-JH2/1018); Educational Department of Liaoning Province, China (Grant No. LJ222511258003); Interdisciplinary project of Dalian University (Grant No. DLUXK-2025-QN-020);
111 Center (Grant No. D23006). (Corresponding authors: Jing Qin; Cong Wang)}

\thanks{Wanshu Fan, Yunzhe Zhang, Yue Shen, Jing Qin are with the School of Software Engineering, Dalian University, Dalian, China (E-mail: fanwanshu@dlu.edu.cn;
zhangyunzhe@s.dlu.edu.cn;
chenyue@s.dlu.edu.cn;
qinjing@dlu.edu.cn).
}
\thanks{Liyan Wang is with the School of Mathematical Sciences, Dalian University of Technology, Dalian, China (E-mail: wangliyan@mail.dlut.edu.cn).}

\thanks{Kin-Man Lam is with The Hong Kong Polytechnic University, Hong Kong, China (E-mail: kin.man.lam@polyu.edu.hk).}

\thanks{Cong Wang is with the Department of Radiology and Biomedical Imaging, University of California, San Francisco, 94107, USA (e-mail: supercong94@gmail.com).}

\thanks{Jinshan Pan is with the School of Computer Science and Engineering, Nanjing University of Science and Technology, Nanjing, China (E-mail: sdluran@gmail.com).}

}

\markboth{Journal of \LaTeX\ Class Files,~Vol.~14, No.~8, August~2021}%
{Shell \MakeLowercase{\textit{et al.}}: A Sample Article Using IEEEtran.cls for IEEE Journals}

\IEEEpubid{}

\maketitle

\begin{abstract}
Adverse weather causes diverse and complex image degradations, severely compromising the reliability of computer vision systems.
Existing all-in-one restoration models attempt to address multiple degradation types within a unified framework, but often lack explicit spatial and semantic modeling of degradation characteristics, limiting their adaptability to diverse weather conditions.
To address this limitation, we propose a Degradation-Aware Cross-Modal Prompt Compensation Network (DCMPC-Net) that 
leverages cross-modal degradation cues from
a pre-trained vision-language model to condition restoration features within a unified backbone.
Specifically, our DCMPC-Net mainly consists of the Cross-Modal Prompt Generator (CMPG), Prompt-Guided Attention Alignment Module (PGAAM), and Dual Feature Compensation Module (DFCM).
%
The CMPG integrates textual embeddings with visual features to produce degradation-aware prompts that encode degradation-related semantic and contextual cues.
These prompts are injected into the decoder via a PGAAM, which adaptively aligns semantic information with degraded regions to facilitate context-aware restoration.
To further enhance structural fidelity, DFCM is introduced that disentangles degradation artifacts from scene structures, thereby improving the reconstruction of fine textures and detailed content.
By integrating cross-modal semantic guidance with spatial alignment and structural enhancement, DCMPC-Net achieves robust and perceptually consistent restoration across diverse weather conditions.
Extensive experiments show that DCMPC-Net outperforms state-of-the-art methods in both task-specific and unified settings, achieving superior accuracy and visual fidelity.
The code is available at 
\href{https://github.com/fanamber831/DCMPC-Net}{https://github.com/fanamber831/DCMPC-Net}.
\end{abstract}


\begin{figure}[t]
\centering
\includegraphics[width=1.025\linewidth]{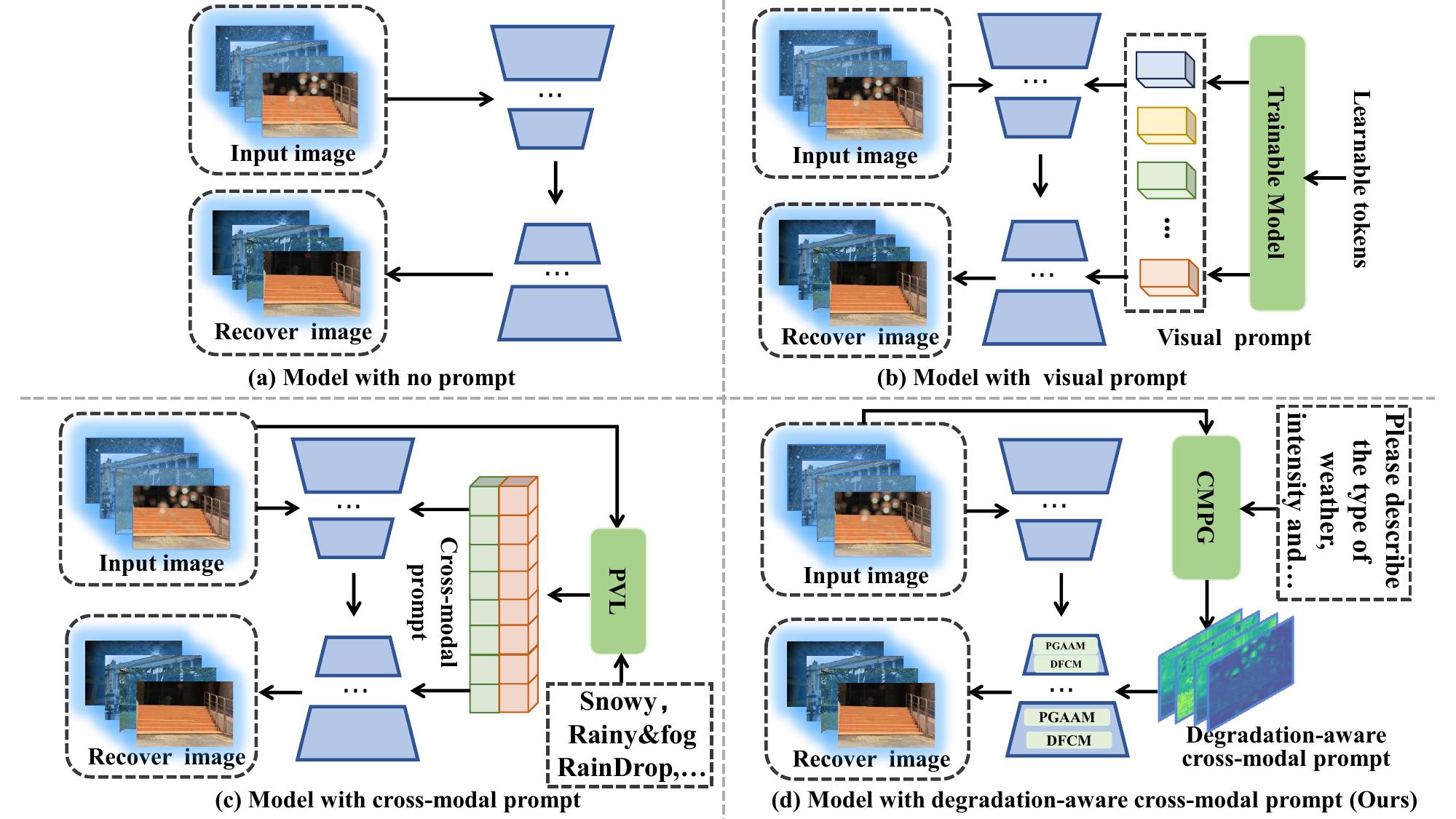}  
\caption{
Conceptual comparison of restoration frameworks for adverse weather. 
(a) All-in-one models use a shared backbone without explicit degradation modeling.
(b) Visual prompt models use handcrafted or learned visual tokens without semantic guidance. 
(c) Cross-modal prompt models introduce semantic priors from vision-language models as high-level or auxiliary guidance, but their feature-level interaction with locally degraded visual features remains limited.
(d) Our DCMPC-Net uses a Cross-Modal Prompt Generator (CMPG) to construct latent degradation-aware cross-modal prompt representations, a Prompt-Guided Attention Alignment Module (PGAAM) 
to align them with degradation-related visual features, and a Dual Feature Compensation Module (DFCM) for structural compensation, enabling robust and detail-preserving restoration across diverse weather conditions.}
\label{fig:p1}
\end{figure}

\section{Introduction}\label{sec: Introduction}

\IEEEPARstart{A}{dverse} weather removal is a fundamental low-level vision task that aims to recover high-quality images degraded by environmental conditions such as rain, fog, and snow.
While substantial progress has been made in this field, the complexity and diversity of real-world degradation patterns present significant challenges for accurate and robust restoration.
Early works have primarily focused on task-specific networks tailored for single degradation types, including deraining~\cite{AttentiveGAN,dcsfn,jdnet,online_derain_aaai22,DRSformer,wang_gragh_derain_ijcai24,wang_acmmm24_derain,PPTformer}, dehazing~\cite{Gated_Fusion_Network,SelfPromer,uhdformer,wang2026neural}, and desnowing~\cite{Online_Rain/Snow_Removal, DDMSNet}.
%
However, these models require retraining or structural modifications for different scenarios, limiting their scalability in real-world applications.

To improve generalization, recent works have explored all-in-one restoration frameworks~\cite{AirNet, All-in-one, Restormer}, which aim to handle diverse weather degradations within a unified model (Fig.~\ref{fig:p1}(a)). 
Most of these methods adopt an encoder-decoder architecture, where the encoder captures multi-scale hierarchical features from degraded images, and the decoder progressively reconstructs clean outputs. 
While effective, these frameworks generally rely on implicit feature learning without explicitly modeling the underlying degradation conditions, resulting in limited adaptability under diverse weather scenarios.
%
%

%
Motivated by the success of prompt learning in natural language processing, 
recent studies have explored prompt-based frameworks for image restoration~\cite{GenLV,weather_self_prompt} (Fig.~\ref{fig:p1}(b)). 
These methods rely on learnable prompts, where degradation representations are encoded as task-aware tokens to guide a shared backbone for image restoration.
%
{
Representative prompt-based image restoration methods~\cite{PromptRestorer,PromptIR,AWRCP} encode degradation cues as learnable prompts or implicit conditioning vectors to modulate a shared restoration backbone. While effective, these approaches typically generate holistic, image-level representations that lack spatial awareness, implicitly assuming that degradation is distributed uniformly across the scene. Consequently, they struggle with real-world adverse weather where artifacts like raindrops or fog density are spatially heterogeneous, leading to sub-optimal restoration in regions with complex, non-uniform degradation.
}
%
%
%
\begin{figure}[t]\footnotesize
    \centering
    \begin{minipage}[t]{0.05\linewidth}
        \centering
        \vspace{1.5mm} 
        \scriptsize \rotatebox{90}{Snow}
    \end{minipage}%
    \begin{minipage}[t]{0.23\linewidth}
        \centering
        \adjustbox{valign=t}{\includegraphics[width=\linewidth, height=1.5cm]{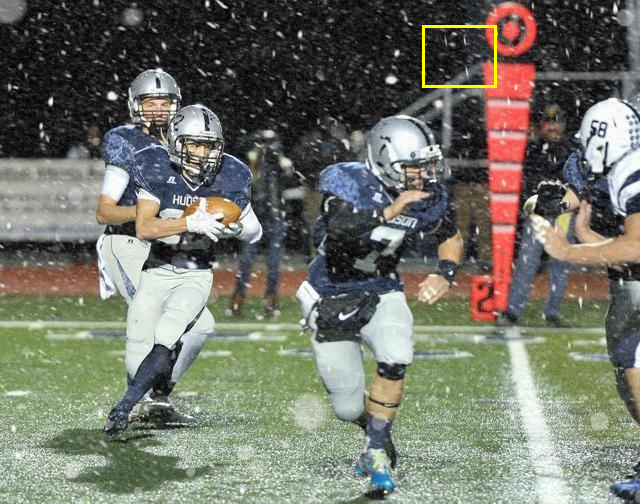}}
    \end{minipage}
    \begin{minipage}[t]{0.23\linewidth}
        \centering
        \adjustbox{valign=t}{\includegraphics[width=\linewidth, height=1.5cm]{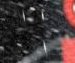}}
    \end{minipage}
    \begin{minipage}[t]{0.23\linewidth}
        \centering
        \adjustbox{valign=t}{\includegraphics[width=\linewidth, height=1.5cm]{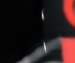}}
    \end{minipage}
    \begin{minipage}[t]{0.23\linewidth}
        \centering
        \adjustbox{valign=t}{\includegraphics[width=\linewidth, height=1.5cm]{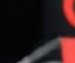}}
    \end{minipage}

\vspace{0.5mm}
    \begin{minipage}[t]{0.05\linewidth}
        \centering
        \vspace{1.5mm} 
        \scriptsize \rotatebox{90}{Rain\&Fog}
    \end{minipage}%
    \begin{minipage}[t]{0.23\linewidth}
        \centering
        \adjustbox{valign=t}{\includegraphics[width=\linewidth, height=1.5cm]{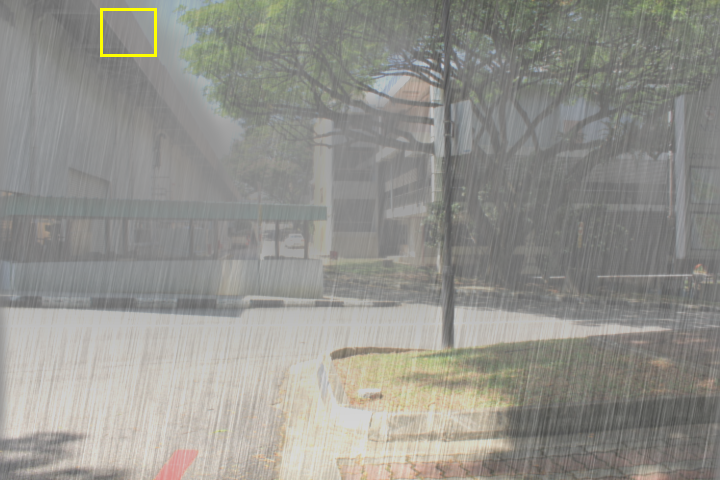}}
    \end{minipage}
    \begin{minipage}[t]{0.23\linewidth}
        \centering
        \adjustbox{valign=t}{\includegraphics[width=\linewidth, height=1.5cm]{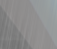}}
    \end{minipage}
    \begin{minipage}[t]{0.23\linewidth}
        \centering
        \adjustbox{valign=t}{\includegraphics[width=\linewidth, height=1.5cm]{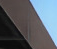}}
    \end{minipage}
    \begin{minipage}[t]{0.23\linewidth}
        \centering
        \adjustbox{valign=t}{\includegraphics[width=\linewidth, height=1.5cm]{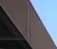}}
    \end{minipage}

\vspace{0.5mm}
    \begin{minipage}[t]{0.05\linewidth}
        \centering
        \vspace{1.5mm} 
        \scriptsize \rotatebox{90}{RainDrop}
    \end{minipage}%
    \begin{minipage}[t]{0.23\linewidth}
        \centering
        \adjustbox{valign=t}{\includegraphics[width=\linewidth, height=1.5cm]{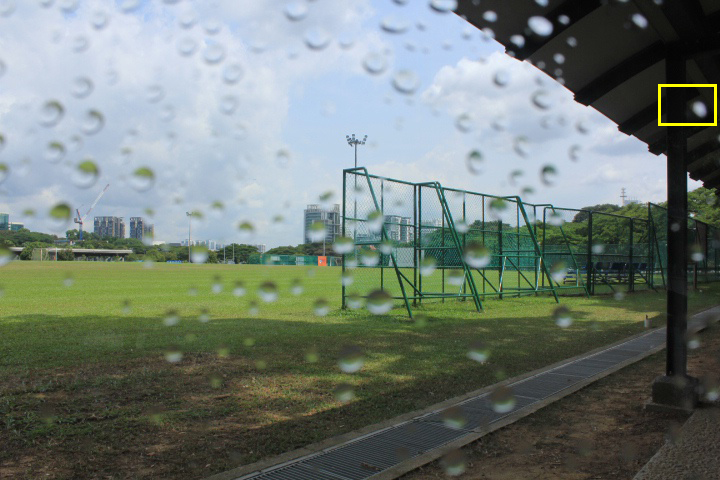}}\\[0.3em]
        \PanelLabel{Degrade Image}
    \end{minipage}
    \begin{minipage}[t]{0.23\linewidth}
        \centering
        \adjustbox{valign=t}{\includegraphics[width=\linewidth, height=1.5cm]{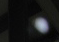}}\\[0.3em]
        \PanelLabel{Input}
    \end{minipage}
    \begin{minipage}[t]{0.23\linewidth}
        \centering
        \adjustbox{valign=t}{\includegraphics[width=\linewidth, height=1.5cm]{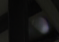}}\\[0.3em]
        \PanelLabel{Histoformer}
    \end{minipage}
    \begin{minipage}[t]{0.23\linewidth}
        \centering
        \adjustbox{valign=t}{\includegraphics[width=\linewidth, height=1.5cm]{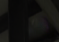}}\\[0.3em]
        \PanelLabel{Ours}
    \end{minipage}
    \caption{A comparative analysis demonstrates that our proposed approach consistently enhances performance across various all-in-one image restoration tasks and yields superior visual results.}
    \label{fig:p2}
\end{figure}

More recently, cross-modal prompt strategies based on pre-trained vision-language models (PVLs) have shown promise in guiding restoration with semantic knowledge~(Fig.~\ref{fig:p1}(c)).
These methods aim to introduce semantic priors to guide the restoration process from a language perspective.  
%
Conde et al.~\cite{InstructIR} utilize human-written textual instructions to provide explicit semantic guidance.
%
Luo et al.~\cite{DACLIP} introduce a vision-language prior-based framework for incorporating degradation-related semantics into image restoration.
%
%
Ai et al.~\cite{GenIR} propose a dual-prompt strategy that leverages contrasting textual semantics to guide the restoration process.
%
Ai et al.~\cite{MPerceiver} further enrich both image and text representations via multimodal prompt learning.
%
%
Yang et al.~\cite{Language-driven} further propose the Language-driven Restoration framework (LDR), which queries PVLs to obtain degradation-related semantics and transforms them into degradation maps for MoE-based expert routing.


Despite these advances, how to incorporate such semantic priors into restoration networks remains insufficiently explored. 
Previous cross-modal restoration studies have demonstrated the value of semantic guidance,  but the interaction between semantic cues and locally degraded visual features is still limited.  
LDR~\cite{Language-driven} further converts vision-language priors into explicit degradation maps for MoE-based expert routing. 
While this design enables adaptive expert selection, the semantic prior mainly serves as a routing cue rather than a continuous feature-conditioning signal for refining restoration features. 
Therefore, the direct interaction between degradation-aware prompts and restoration features within a unified backbone remains underexplored.

To address the above limitations, we propose a Degradation-Aware Cross-Modal Prompt Compensation Network (DCMPC-Net) for adverse weather removal.
{Rather than using language-derived degradation semantics only as external descriptions or routing cues, DCMPC-Net couples semantic priors with degraded visual features and formulates them as latent degradation-aware cross-modal prompt representations for feature-level conditioning within a unified restoration backbone~(Fig.~\ref{fig:p1}(d)).
%
Specifically, our DCMPC-Net mainly consists of the Cross-Modal Prompt Generator (CMPG), Prompt-Guided Attention Alignment Module (PGAAM), and Dual Feature Compensation Module (DFCM).
%
CMPG integrates textual embeddings extracted by frozen semantic prior models with visual features to generate degradation-aware prompts.
To further refine this semantic guidance and establish spatial correspondence between degradation-aware cross-modal prompts and degraded regions, we develop the PGAAM, 
which adaptively modulates feature responses and highlights restoration-relevant regions.
In addition, the DFCM is used to enhance structural fidelity by disentangling degradation artifacts from scene content and reinforcing critical visual details. 
By effectively aligning degradation-aware cross-modal prompts with visual representations, DCMPC-Net enhances spatial precision and semantic coherence in the restoration process.
As illustrated in Fig.~\ref{fig:p2}, DCMPC-Net not only restores clearer structures but also preserves fine textures under challenging conditions such as snow, rain\&fog, and raindrop scenes.

Our contributions can be summarized as follows.
\begin{itemize}
\item We propose a Degradation-Aware Cross-Modal Prompt Compensation Network 
for adverse weather removal, which introduces degradation-aware cross-modal prompts into a unified restoration backbone for prompt-guided feature interaction and compensation.

\item We design a Cross-Modal Prompt Generator to extract degradation-aware cross-modal prompts by integrating textual and visual features, and introduce a Prompt-Guided Attention Alignment Module to align these prompts with degraded regions by establishing spatial correspondence.
\item We develop a Dual Feature Compensation Module to enhance structural fidelity by disentangling degradation artifacts from scene content, effectively preserving fine textures and structural details.
\item We conduct comprehensive evaluations on multiple benchmarks, demonstrating that our method consistently achieves leading performance over existing restoration baselines.
\end{itemize}


\section{Related Work}\label{sec: Related Work}
\subsection{Adverse Weather Removal} 
Adverse weather conditions lead to complex low-level degradations that significantly impair image visibility and compromise the performance of subsequent vision systems.
Recent efforts~\cite{All-in-one,TransWeather,AirNet,AIRformer,Restormer,AdaIR_Adaptive} have focused on all-in-one restoration frameworks to handle diverse weather degradations using a single model. 
%
Li et al.~\cite{All-in-one} propose an all-in-one image restoration framework that employs multiple encoders to model different degradation types, improving restoration performance across diverse conditions.
To improve computational efficiency, 
Valanarasu et al.~\cite{TransWeather} introduce TransWeather, a unified transformer-based restoration network that incorporates weather-type embeddings within a single encoder–decoder architecture.
Different from weather-specific designs, 
Li et al.~\cite{AirNet} develop AirNet, which leverages contrastive coding-based degradation representations to enhance generalization to unknown corruption types.
Focusing on efficiency in high-resolution restoration,
Zamir et al.~\cite{Restormer} introduce Restormer, an efficient transformer that models global dependencies via channel-wise self-attention while incorporating convolutional operations.
%
From a frequency-domain perspective,
Cui et al.~\cite{AdaIR_Adaptive} further explore adaptive all-in-one image restoration, introducing AdaIR to jointly exploit spatial and spectral information for improved separation of degradation artifacts from clean scene content.
Similarly, 
Gao et al.~\cite{AIRformer} develop AIRformer, a frequency-oriented transformer framework that leverages frequency priors to guide image restoration under diverse weather degradations.
%
Sun et al.~\cite{Histoformer} further extend statistical prior modeling by introducing Histoformer, which incorporates histogram priors into a transformer-based framework to facilitate global structure modeling and adaptive feature modulation.
{More recently, Wang et al.~\cite{MOERL} propose MOERL, a reinforcement learning-based framework that integrates a mixture-of-experts module into a vision–language architecture, casting all-in-one adverse weather image restoration as a sequential decision-making problem.} 

Despite these advancements, existing all-in-one models often rely on static priors and insufficiently exploit semantic information, which limits their adaptability to diverse and complex degradations. Therefore, our research is dedicated to exploring degradation-aware semantic guidance that enables adaptive restoration across adverse weather conditions.
\subsection{Vision-Language Model} 
The success of large language models (LLMs)~\cite{Bert}, such as ChatGPT~\cite{GPT}, LLaMA~\cite{Llama3}, and so on, has inspired increasing efforts to transfer their reasoning capabilities to visual domains. 
Pre-trained vision-language models (PVLs), such as CLIP~\cite{clip} and BLIP~\cite{BLIP2}, are pre-trained on large-scale image-text pairs to learn shared semantic representations across modalities. 
These models typically consist of separate encoders for image and text inputs, and support joint embedding, cross-modal alignment, and prompt-based conditioning. Due to their strong generalization and semantic understanding abilities, PVLs have been widely applied to tasks such as captioning~\cite{caption} and visual question answering~\cite{VQA1}. 
Recently, several studies~\cite{DACLIP, Language-driven, GenIR, Textual_Removal} have explored the integration of vision-language models (PVLs) into image restoration pipelines.
Luo et al.~\cite{DACLIP} propose DA-CLIP to align textual embeddings with visual features for condition-aware prompt generation.
To enhance prompt effectiveness, 
Zhou et al.~\cite{GenIR} design a dual-prompt learning strategy that leverages both positive and negative samples to provide complementary guidance for image restoration.
In contrast, 
Lin et al. ~\cite{Textual_Removal}~focus on optimizing textual inputs to convey image content.
While these methods demonstrate the potential of PVLs to introduce high-level semantic priors, they are mainly tailored for single-type degradations and exhibit limited generalization to real-world scenarios involving diverse weather conditions.
%
%
Specifically, Yang et al.~\cite{Language-driven} propose a Language-driven Restoration framework (LDR), which converts vision-language priors into a pixel-wise degradation map and then into an expert-selection score map for Top-\(K\) expert activation within a Mixture-of-Experts (MoE) framework. 
Thus, the vision-language prior in LDR acts as a routing signal that determines the executed expert path for each input. 
By contrast, DCMPC-Net does not construct an explicit degradation map, generate routing scores, or select experts. 
Instead, it encodes degradation semantics as latent cross-modal prompts and injects them into a unified restoration backbone through prompt-guided attention alignment, using the vision-language prior to refine restoration features without changing the computation path.

\subsection{Prompt Learning} 
Prompt learning~\cite{MAE-VQGAN,SelfPromer} has achieved remarkable success in the natural language processing (NLP) ~\cite{NLP_prompt,NLP1,NLP2}, providing a lightweight mechanism to condition model behavior via task-relevant cues. 
Inspired by its effectiveness, recent studies~\cite{VPT,Prompt-in-prompt,TextualIR,InstructIR} extend prompt learning to computer vision tasks, particularly in the context of image restoration. 
%
Jia et al.~\cite{VPT} initially design visual prompts as an efficient alternative to full fine-tuning for large-scale vision transformers. 
To capture more semantic information and details, 
Jia et al.~\cite{Prompt-in-prompt} extract degradation-aware and restoration visual prompts from encoded features, thereby enhancing the quality of recovered images. 
However, the emphasis on visual prompts for identifying degradation types overlooks the crucial role of textual information, resulting in a semantic gap that impedes the precise identification of these types.
%
Yan et al.~\cite{TextualIR} propose a textual prompt-guided image restoration framework, demonstrating the feasibility of leveraging language cues to guide restoration tasks. 
Conde et al.~\cite{InstructIR} adopt a textual format similar to that of TextualIR~\cite{TextualIR} to represent degradation information by embedding natural language instructions into traditional image restoration pipelines.
To enrich the semantic representation of image textures, recent works tend to explore the power of pre-trained visual language (PVL) models for learning the diversity of weather-specific knowledge.
%
Zhang et al.~\cite{SSP-IR} leverage a PVL to generate explicit text embedding for improving the details of the model. 
Besides, 
Yu et al.~\cite{Scaling-Up-to-Excellence} employ a PVL to provide image content prompts, greatly improving the accuracy and intelligence. 
Compared with the conventional question-answering paradigm described above, 
Zhou et al.~\cite{GPP_LLIE} 
reformulate the problem paradigm by providing explicit degradation options, each accompanied by corresponding textual descriptions. 
Although these approaches have demonstrated the potential of prompt-based guidance in visual restoration, several limitations remain unresolved, which rely on static or ambiguous prompts and naive cross-modal fusion, limiting adaptability and weakening restoration guidance due to poor feature alignment.

\begin{figure*}[!t] 
\includegraphics[width=\linewidth]{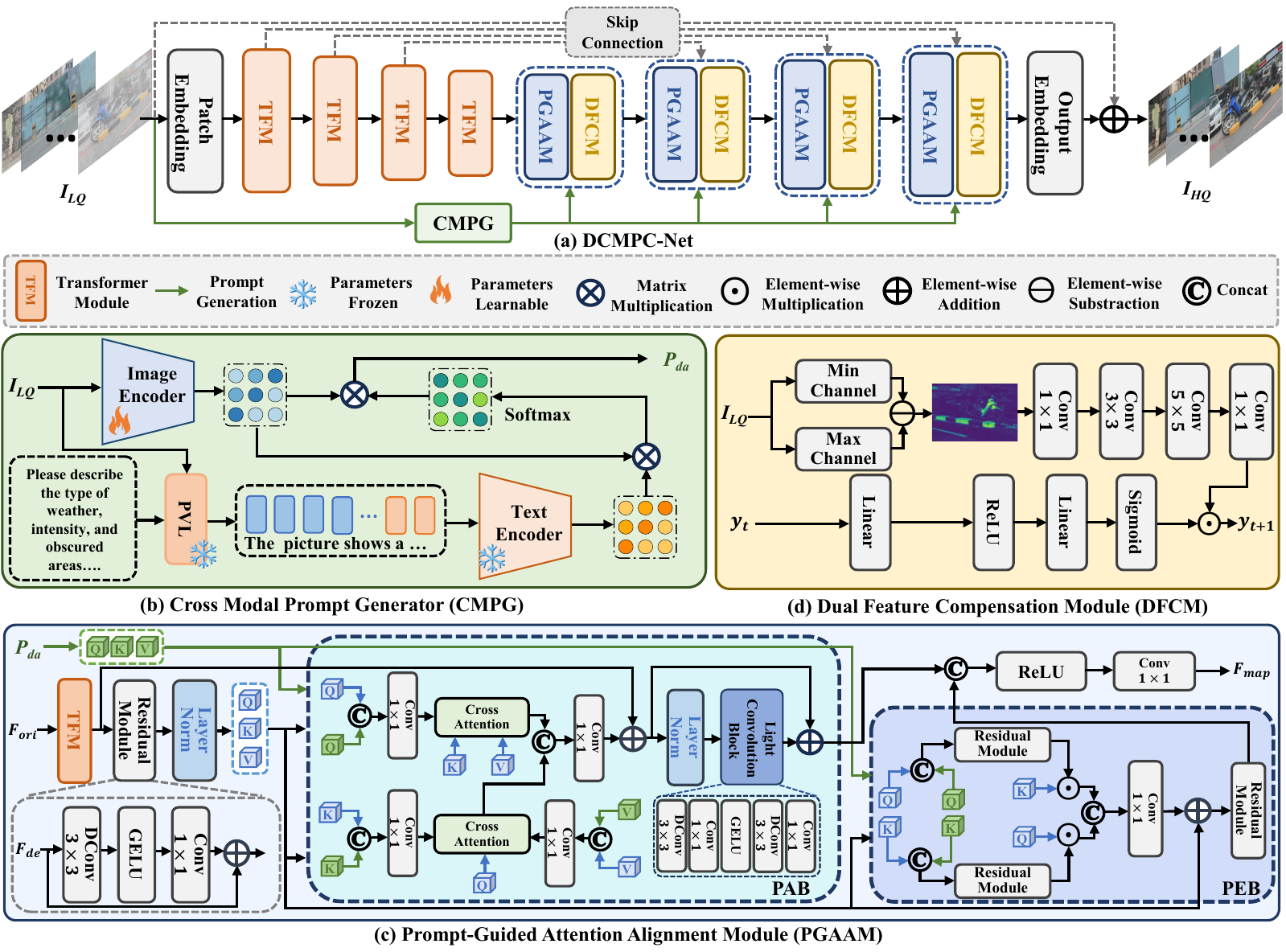}
\vspace{-3mm}
\caption{ 
{Overview of the proposed DCMPC-Net architecture.
The framework consists of three core components: Cross-Modal Prompt Generator (CMPG), Prompt-Guided Attention Alignment Module (PGAAM), and Dual Feature Compensation Module (DFCM). 
Given an adverse weather image, 
CMPG integrates the semantic embedding obtained from frozen semantic prior models with visual features} to generate multi-scale, degradation-aware cross-modal prompts.
These prompts are injected into the restoration backbone via PGAAM, which adaptively aligns semantic guidance with degraded regions. Meanwhile, DFCM enhances structural fidelity by disentangling degradation artifacts from scene content, leading to improved detail preservation and robustness under complex weather conditions.
}
\label{fig:p3}
\end{figure*}
\section{Method}
\label{sec: Method}
Our goal is to restore high-quality images degraded by adverse weather conditions.
%
To this end, we propose DCMPC-Net, a Degradation-Aware Cross-Modal Prompt Compensation Network that incorporates high-level semantic knowledge to enhance degradation-aware restoration.
We introduce a Cross-Modal Prompt Generator (CMPG) that generates degradation-aware cross-modal prompts by integrating semantic embeddings from a pre-trained vision-language model LLaMA~\cite{Llama3} with visual features (see Section~\ref{sec: CMPG}).
%
These prompts are incorporated into a Prompt-Guided Attention Alignment Module (PGAAM), which progressively refines attention to degraded regions through an alignment strategy (see Section~\ref{sec: PGAAM}).
To enhance structural fidelity, we further design a Dual Feature Compensation Module (DFCM) that disentangles degradation artifacts from clean scene content for accurate detail restoration (see Section~\ref{sec: DFCM}).
%

\subsection{Overall Pipeline}\label{sec: overall}

Fig.~\ref{fig:p3} illustrates the overall pipeline of DCMPC-Net, which comprises two main stages: prompt generation and image restoration.
In the prompt generation stage, semantic descriptions are extracted from a pre-trained vision-language model and integrated with visual features to construct degradation-aware cross-modal prompts.
These prompts are then fed into the image restoration stage to guide the decoder in recovering degraded regions.
%



\subsubsection{Prompt generation}
%
%
To enable dynamic perception of degraded regions, we propose a Cross-Modal Prompt Generator (CMPG) that produces degradation-aware cross-modal prompts (see Section~\ref{sec: CMPG} for details). 
These prompts are subsequently integrated into the image restoration process to selectively guide the decoder's attention toward degraded areas, thereby enhancing the overall restoration quality.
%
\subsubsection{Image restoration}
DCMPC-Net adopts a four-level encoder-decoder architecture to learn hierarchical representations for progressive image restoration.
Given a degraded image $I_{LQ}\in\mathbb{R}^{H\times W\times3}$, where $3$ is the number of channels and $H\times W$ represents the spatial dimensions, a $3\times 3$ convolutional layer is first applied to adjust the number of channels and extract the initial feature $X_{t}\in\mathbb{R}^{H\times W\times C}$, where $C$ is the number of channels.
%
%
The feature $X_t$ is subsequently processed by a four-level encoder, in which each level incorporates a transformer module~\cite{Histoformer} to capture hierarchical representations. 
%
%
The resulting encoded features, together with degradation-aware cross-modal prompts generated from the prompt generation process and degraded feature information from the original input, are then fed into a four-level decoder composed of multiple PGAAM and DFCM modules. 
The decoder progressively reconstructs high-resolution feature representations, producing sharper and more accurate restoration results.
Throughout the process, decoder features are concatenated with corresponding encoder features to facilitate reconstruction, followed by a $1\times1$ convolution to reduce channel dimensions.
%
%
Finally, a global residual connection is employed by adding the network output to the degraded input image, yielding the restored result $I_{HQ}\in\mathbb{R}^{H \times W\times3}$ 
\subsection{Cross Modal Prompt Generator}\label{sec: CMPG}
Multimodal foundation models often suffer from embedding inconsistency, where semantically similar textual descriptions yield divergent representations.
To address this issue, we propose a Cross-Modal Prompt Generator (CMPG) that integrates textual semantic information with visual features to generate degradation-aware cross-modal prompts.
As shown in Fig.~\ref{fig:p3}, CMPG comprises four components: (1) a pre-trained vision-language model, (2) a pre-trained text encoder, (3) a trainable image encoder, and (4) a spatial cross-attention module.
{In our framework, the vision-language model and text encoder are kept frozen to preserve their generalized representations, while the image encoder and spatial cross-attention module are jointly optimized with the restoration backbone to capture task-specific features.}
%
\subsubsection{Pre-trained vision-language model}
To leverage high-level semantic knowledge, we adopt LLaMA~\cite{Llama3} as the backbone vision-language model. 
LLaMA is a state-of-the-art open-source model pre-trained on $15$ trillion multimodal tokens. 
Its strong semantic grounding capability enables accurate identification and modeling of degradation patterns across diverse weather conditions.
%
We combine the question $Q_{t}$  and the input image $I_{LQ}$ as input to the model to generate the descriptive answer $T_{text}$, defined as follows:
\begin{equation}
\textit{T}_{\textit{text}} = \textit{PVL}(Q_{t}, {I_{LQ}}),  \quad T_{\textit{text}} \in \mathbb{R}^{L \times N}, 
\end{equation}
where \textit{PVL}($\cdot$) denotes the pre-trained vision-language model LLaMA,~\textit{L} denotes the description length, and~\textit{N} is the channel dimension.
\subsubsection{Pre-trained text encoder}
Additionally, Multilingual-E5~\cite{Text_Encoder} with frozen parameters serves as the text encoder, encoding input descriptions into 1024-dimensional embeddings via tokenization and embedding projection layers.
This design preserves linguistic integrity and mitigates catastrophic forgetting during joint training.
The frozen text encoder maps $T_{text}$ to an embedding $T_{emb}$, which is further projected to match the channel dimension $C$ of the restoration model.
This process can be expressed as:
\begin{equation}
T_{\textit{emb}} = {Enc_{text}}(T_{\textit{text}}), \quad T_{\textit{emb}} \in \mathbb{R}^{L \times C},
\end{equation}
where ${Enc}_{text}$ represents the text encoder.

\subsubsection{Trainable image encoder}
At the same time, we adopt the image encoder from the baseline architecture~\cite{Histoformer} to extract visual features $I_{emb}$ from the input image $I_{LQ}$.

The parameters of this encoder are jointly optimized within our network, thereby retaining its hierarchical representation capabilities and enabling end-to-end training. 
This process is described as follows:

\begin{equation}
I_{\textit{emb}} = {Enc_{img}}(I_{\textit{LQ}}), \quad I_{emb} \in \mathbb{R}^{H \times W \times C}, 
\end{equation}
where \({Enc}_{img}\) represents the trainable image encoder. 
\subsubsection{Spatial cross-attention module}
The image distribution conditioned on textual descriptions exhibits significant multimodality, which poses challenges for effective training.
To address this issue, 
we employ a spatial cross-attention mechanism to integrate $T_{emb}$ and $I_{emb}$, thereby facilitating the modeling of relationships between different modalities. 

Here, the query $Q$ is projected from the text features, while the key $K$ and value $V$ are projected from the image features. 
The computation is defined as follows:
\begin{equation}
Q = \textit{T}_{\textit{emb}} W_q^1, \quad K =\textit{I}_\textit{emb} W_k^1, \quad V = I_{emb} W_v^1,
\end{equation}
where $W_q^1, W_k^1, W_v^1 \in \mathbb{R}^{C_t \times C}$ are the corresponding projection matrices, and $Q, K, V \in \mathbb{R}^{H \times W \times C}$.
The spatial cross-attention is then computed as:
\begin{equation}
P_{da}=V{\cdot}Softmax(\frac{K{\cdot}Q}{\beta}), 
\end{equation}
where $P_{da}$ denotes the degradation-aware cross-modal prompt and $\beta$ represents the scale factor. 
Here, \(P_{da}\) is a latent prompt representation that encodes degradation-related semantic cues and provides conditioning information for subsequent restoration feature refinement.

\subsection{Prompt-Guided Attention Alignment Module}\label{sec: PGAAM}
The Prompt-Guided Attention Alignment Module (PGAAM) incorporates degradation-aware cross-modal prompts into the restoration process following a prompt-driven attention paradigm~\cite{PromptRestorer}.  
As illustrated in Fig.~\ref{fig:p3}, PGAAM comprises two core components: the Prompt Alignment Block (PAB) and the Prompt Enhancement Block (PEB). 
This design enables the model to capture global semantic context while enhancing local structural representations under the guidance of degradation-aware cross-modal prompts. 
Given the intermediate feature $F_{ori}$ from the encoder, the degradation feature $F_{de}$ is extracted using a transformer block~\cite{Histoformer} as $F_{de}=TFM(F_{ori})$, where $TFM(\cdot)$ denotes the transformer operation.

With the degradation feature $F_{de}$ and the prompt feature $P_{da}$, the PGAAM workflow is defined as follows:

\begin{equation}
\mathcal{S} = \mathcal{T}(\varphi_{PAB}(\mathcal{Y}(F_{de}), P_{da},F_{de}), \varphi_{PEB}(\mathcal{Y}(F_{de}), P_{da})),
\end{equation}
where $\mathcal{T}(\cdot,\cdot)=\mathcal{C}_1(\mathcal{W}[\cdot,\cdot])$, $\mathcal{Y}(\cdot)=LN(Res(\cdot))$. 
Here,  $\mathcal{W}[\cdot]$ means the concatenation at channel dimension, $C_1(\cdot)$ represents the operation of $1\times1$ convolution, $Res(\cdot)$ refers to the residual module, and $LN(\cdot)$ means the operation of layer normalization~\cite{layernorm}.
In addition, $\varphi_{PAB}(\cdot,\cdot)$ and $\varphi_{PEB}(\cdot,\cdot)$ denote the operations of PAB and PEB, respectively. 

\subsubsection{Prompt Alignment Block}
The Prompt Alignment Block is responsible for global degradation perception and feature alignment of prompt $P_{da}$ and degradation feature $F_{de}$. 
PAB contains the dual cross-attention mechanism followed by a light convolution module. 
By applying $1\times1$ convolution $\mathcal{C}_1(\cdot)$ and $3\times3$ depth-wise convolution $\mathcal{D}_3(\cdot)$, yielding the degradation query $Q_d=\mathcal{D}_3\mathcal{C}_1(F_{de})$, degradation key $K_d=\mathcal{D}_3\mathcal{C}_1(F_{de})$, degradation value $V_d=\mathcal{D}_3\mathcal{C}_1(F_{de})$, similarity; prompt query $Q_p=\mathcal{D}_3\mathcal{C}_1(P_{da})$, prompt key $K_p=\mathcal{D}_3\mathcal{C}_1(P_{da})$ and prompt value $V_p=\mathcal{D}_3\mathcal{C}_1(P_{da})$ from the prompt.
The dual cross-attention mechanism considers reforming the query, key, and value vector of degradation features and prompts to build representative vectors to perform cross-attention~\cite{Restormer}, we have:

\begin{subequations}
\begin{align}
F_{m} &= CA\left( C_1\left( \mathcal{W} \left[ Q_{d}, Q_p \right] \right), K_d, V_d \right), \\
F_{n} &= CA\left( Q_d, C_1\left( \mathcal{W} \left[ K_d, K_p \right] \right), C_1\left( \mathcal{W} \left[ V_d, V_p \right] \right) \right), \\
F_{i}  &= C_1\left( \mathcal{W} \left[F_{m}, F_{n}\right] \right) + F_{de},
\end{align}
\end{subequations}
where $CA(\cdot,\cdot,\cdot)$ means the operation of cross-attention~\cite{Restormer}. 
Then, the light convolution module considers refining the degradation tensor. 
We format that:
\begin{equation}
F_{t_1} = C_1 D_3 \phi C_1 D_3 \bigl(LN(F_i)\bigr) + F_i,
\end{equation}
where $ C_1 D_3 \phi C_1 D_3 \bigl(\cdot)$ represents the operation of light convolution module shown in the Fig.~\ref{fig:p3}.
\subsubsection{Prompt Enhancement Block}

The Prompt Enhancement Block (PEB) is designed to enhance local structural details by exploiting degradation-related visual features.
Specifically, PEB uses the degradation-aware prompt to guide the refinement of degradation-related visual features through a series of residual modules.
Given the prompt $P_{da}$ and the degradation feature $F_{de}$, we first concatenate their projected representations along the channel dimension. 
The resulting fused features are then processed by multiple residual modules to selectively recover fine structures and details. 
The overall operation of the PEB is defined as:
%
\begin{subequations}
\begin{align}
F_{u} &= Res\left(\mathcal{W}[Q_d, Q_p]\right)\odot K_d, \\
F_{v} &= Res\left(\mathcal{W}[K_d, K_p]\right)\odot Q_d, \\
F_{t_2}  &= Res\left(C_1\left(\mathcal{W}[F_{u}, F_{v}]\right) + \mathcal{Y}(F_{de})\right),
\end{align}
\end{subequations}
where $\odot$ represents the element-wise multiplication and $\mathcal{Y}(\cdot)=LN(Res(\cdot))$. 
The outputs of PAB and PEB, denoted as $F_{t_1}$ and $F_{t_2}$ respectively, are concatenated and further refined by a $1 \times 1$ convolutional layer with ReLU activation:
%
\begin{equation}
F_{map}=C_1(\mathcal{R}(\mathcal{W}[F_{t_{1}},F_{t_{2}}])),
\end{equation}
%
where $\mathcal{R}(\cdot)$ is the ReLU activation.
\(F_{map}\) denotes the prompt-conditioned feature representation obtained by integrating the degradation-aware cross-modal prompt \(P_{da}\) with the degradation feature \(F_{de}\) in PGAAM.

As shown in Fig.~\ref{fig:p4}, PGAAM enables precise focus on degraded regions, facilitating accurate and robust restoration. 
By integrating parallel processing pathways, PGAAM allows the model to effectively handle both globally distributed degradations and fine-grained local details.


\begin{figure}[!t]\footnotesize
\begin{center}
\begin{tabular}{ccccc}
\hspace{-1.5mm}\includegraphics[width = 0.19\linewidth]{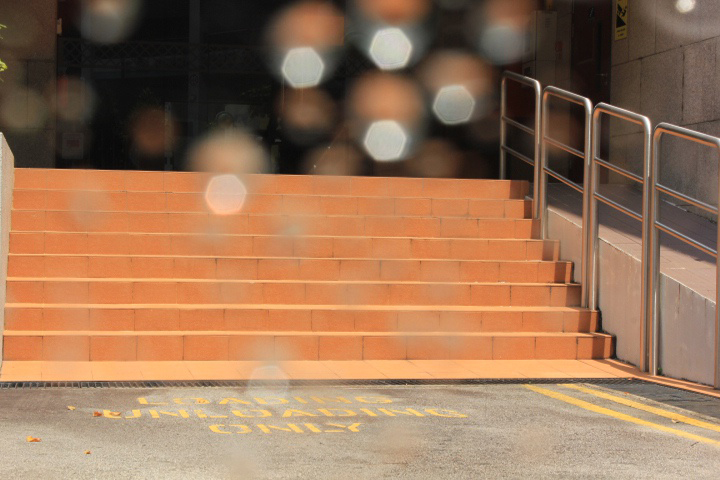} &\hspace{-4.5mm}
\includegraphics[width = 0.19\linewidth]{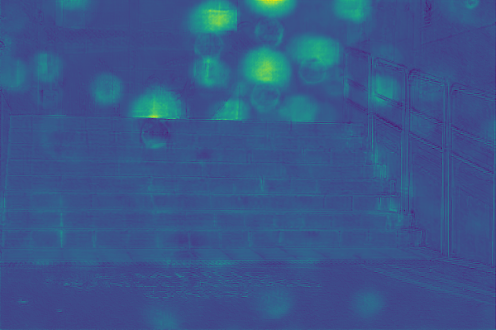}  &\hspace{-4.5mm}
\includegraphics[width = 0.19\linewidth]{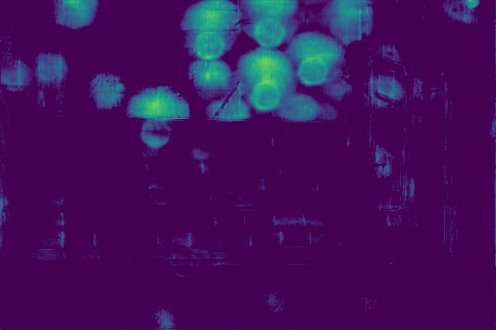} &\hspace{-4.5mm}
\includegraphics[width = 0.19\linewidth]{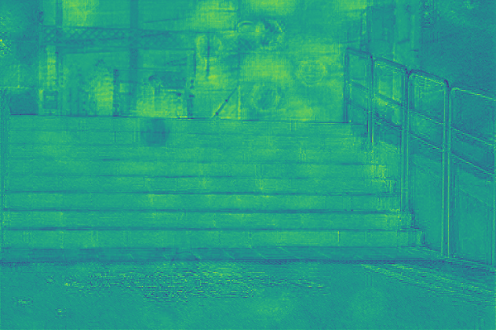} &\hspace{-4.5mm}
\includegraphics[width = 0.19\linewidth]{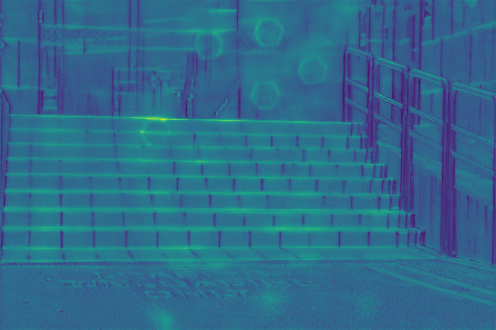} 
\\
\hspace{-1.5mm}\includegraphics[width = 0.19\linewidth]{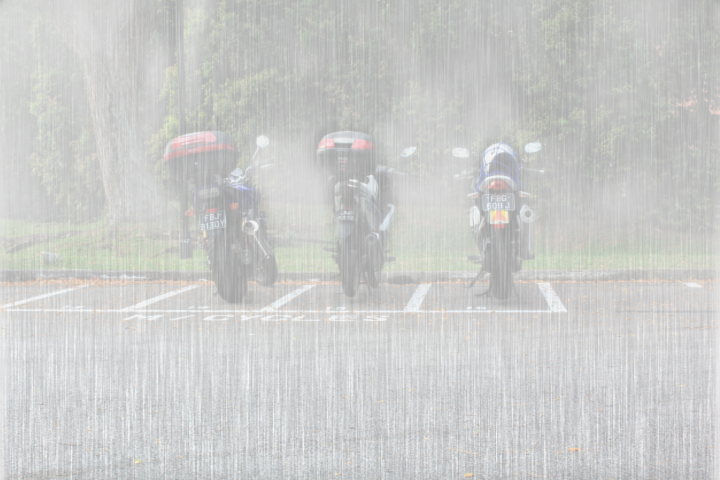} &\hspace{-4.5mm}
\includegraphics[width = 0.19\linewidth]{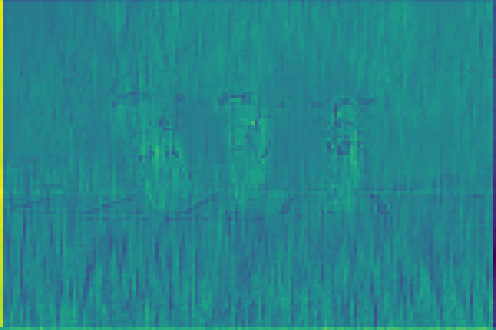} &\hspace{-4.5mm}
\includegraphics[width = 0.19\linewidth]{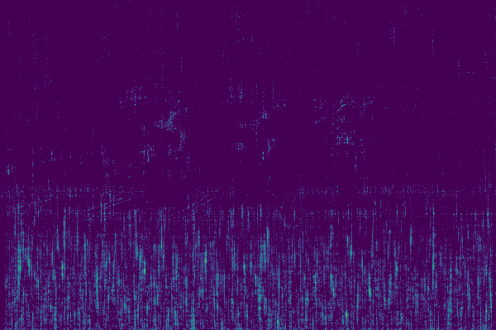} &\hspace{-4.5mm}
\includegraphics[width = 0.19\linewidth]{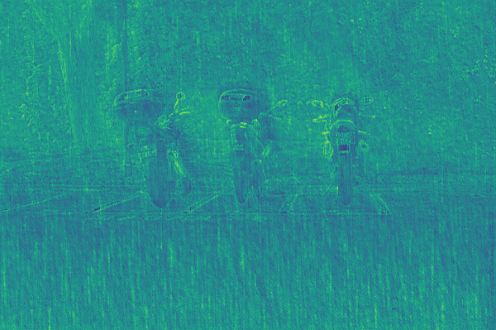} &\hspace{-4.5mm}
\includegraphics[width = 0.19\linewidth]{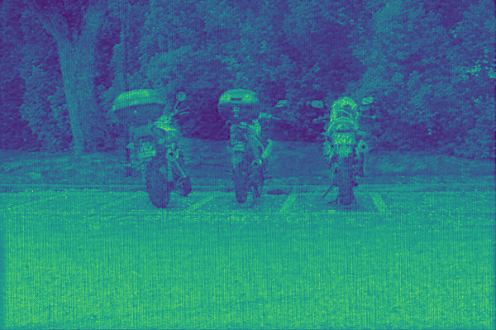}  
\\
\hspace{-1.5mm}\includegraphics[width = 0.19\linewidth]{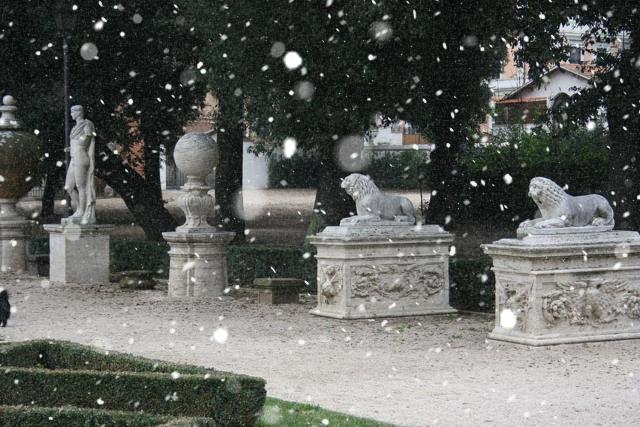} &\hspace{-4.5mm}
\includegraphics[width = 0.19\linewidth]{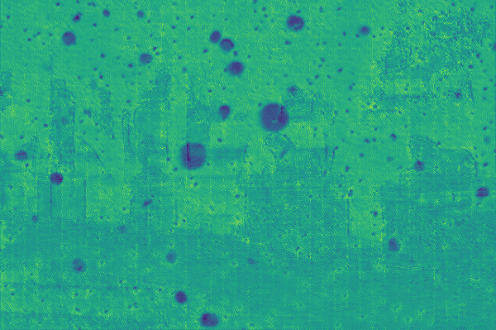}  &\hspace{-4.5mm}
\includegraphics[width = 0.19\linewidth]{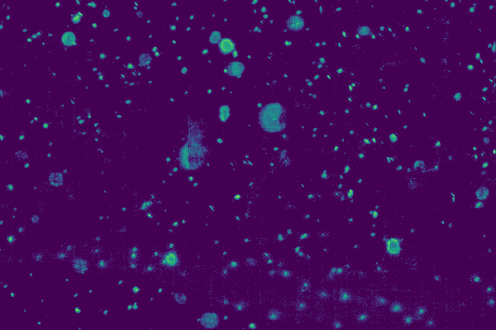} &\hspace{-4.5mm}
\includegraphics[width = 0.19\linewidth]{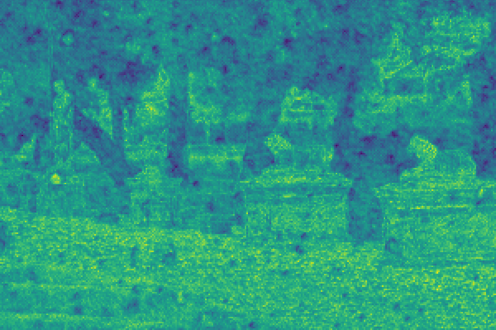} &\hspace{-4.5mm}
\includegraphics[width = 0.1975\linewidth]{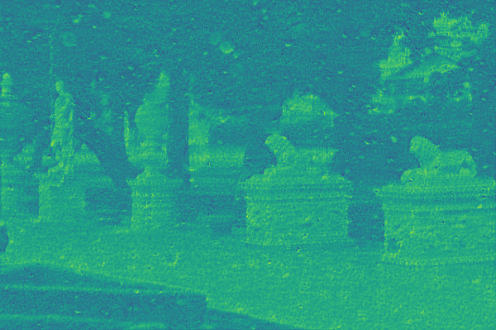} 
\\
\hspace{-1.5mm}$I_{LQ}$   &\hspace{-4.5mm}  $\textit{P}_{da}$ &\hspace{-4.5mm} $F_{map}$ &\hspace{-4.5mm}  $y_{t}$&\hspace{-4.5mm} $y_{t+1}$ 
\end{tabular}
\end{center}
\vspace{-2mm}
\caption{Visualization of $\textit{P}_{da}$, $F_{map}$, $y_{t}$, and $y_{t+1}$. From $\textit{P}_{da}$ to $F_{map}$, the degraded regions become more prominent, demonstrating the effectiveness of the proposed PGAAM in highlighting degradation cues. Furthermore, the transition from $y_{t}$ to $y_{t+1}$ reveals enhanced structural details, underscoring the contribution of DFCM to structural refinement.
}
\label{fig:p4}
\end{figure}

\subsection{Dual Feature Compensation Module}\label{sec: DFCM}

Conventional image restoration networks~\cite{Restormer,AIRformer} usually process channel and pixel features uniformly, which limits their ability to distinguish foreground content from severe background degradation. 
Moreover, shallow textual and visual cues may be gradually weakened as the network depth increases.
Inspired by prior work~\cite{Residual_channel}, which refines degraded images by exploiting channel-wise variance between maximum and minimum responses without introducing additional learnable parameters, 
we design a Dual Feature Compensation Module (DFCM) to effectively distinguish visually similar background object features in degraded images. 
The process is defined as follows: 
\begin{equation}
C_{map}=\max _{c \in \{r, g, b\}} I^{c}_{LQ}-\min _{c \in \{r, g, b\}} I^{c}_{LQ}.
\end{equation}

As illustrated in Fig.~\ref{fig:p3}, DFCM comprises two parallel branches: the feature compensation branch and the degradation restoration branch. 
A convolutional neural network and a linear transformation layer are used, respectively. 
Given the $C_{map}$, we first increase the channel dimension via a $1\times1$ convolution and then extract high-level information via a $3\times3$ convolution and a $5\times5$ convolution. 

Finally, local details are refined using $1\times1$ convolution. 
Meanwhile, the degradation restoration branch introduces linear transformations and the $ReLU$ activation function to further refine information. 
This addition significantly strengthens and reconstructs background information disrupted by adverse weather conditions. 
The description is as follows:
\begin{subequations}
\begin{align}
C_{out} &= C_1\left(C_5\left(C_3\left(C_1\left(C_{map}\right)\right)\right)\right),  \\
y_{out} &= Sigmoid\left(LE\left(\mathcal{R}\left(LE\left(y_{t}\right)\right)\right)\right),\\
y_{t+1} &= y_{out}\odot C_{out},
\end{align}
\end{subequations}
where $LE$ denotes the linear transform and $Sigmoid$ represents the sigmoid activation. 
The visualization of feature maps, as illustrated in Fig.~\ref{fig:p4}, demonstrates that the features processed by the DFCM have refined the details of the image structure. 

\subsection{Loss Function}
To train the network, three objective loss functions are employed, including image reconstruction loss ($\mathcal{L}_1$) for pixel recovery, patch-level correlation loss ($\mathcal{L}_{cor}$) for structural consistency~\cite{Histoformer}, and frequency-domain loss ($\mathcal{L}_{fre}$) for detail enhancement~\cite{FFTloss}:
\begin{equation}
\mathcal{L} = \mathcal{L}_1 + \mathcal{L}_{cor} + \mathcal{L}_{fre},
\end{equation}
where
\begin{align}
\mathcal{L}_1 &= \|\widehat{H} - H\|_1, \\
\mathcal{L}_{cor} &= \frac{1}{2} \left[ 1 - \rho(\widehat{H}, H) \right], \\
\mathcal{L}_{fre} &= \|\mathcal{F}(\widehat{H}) - \mathcal{F}(H)\|_1,
\end{align}
where $H$ denotes the ground truth image, $\widehat{H}$ denotes the restored image; $\rho(\cdot)$ is the Pearson correlation coefficient~\cite{pearson} and $\mathcal{F}(\cdot)$ represents the Fast Fourier Transform.

\section{Experiments} 
\subsection{Datasets}
Several benchmark datasets are widely adopted under adverse weather removal tasks, including Snow100K~\cite{DesnowNet}, Outdoor-Rain~\cite{Test1}, and RainDrop~\cite{RainDropAttn}. 
Outdoor-Rain consists of 9,000 training samples and 1,500 validation samples. Snow100K provides 100k synthetic snow images paired with corresponding snow-free ground truth images. RainDrop comprises 1,119 pairs of real-world adherent raindrop images and their corresponding clean background references. 
We conduct experiments on the All-Weather dataset~\cite{All-in-one}, which primarily contains degradations caused by four typical adverse weather conditions, such as rain, fog, raindrops, and snow. 
This dataset includes 18069 images as the training data, comprising 9001 snowy images from Snow100K~\cite{DesnowNet}, 818 images with raindrops from RainDrop~\cite{RainDropAttn}, and 8250 heavy rain images from Outdoor-Rain~\cite{Test1} for rain\&fog. 
For testing, there are also 750 images from Outdoor-Rain Test1~\cite{Test1}, 58 images from RainDrop TestA~\cite{RainDropAttn}, 16,611 images from Snow100K-S~\cite{DesnowNet}, and 16,801 images from Snow100K-L~\cite{DesnowNet}. 

\begin{table*}
\centering
\caption{{Quantitative results on widely-used benchmarks, including Snow100K-S~\cite{DesnowNet}, Snow100K-L~\cite{DesnowNet}, Outdoor-Rain~\cite{Test1}, Raindrop~\cite{RainDropAttn} in terms of PSNR and SSIM. The \textbf{best} and the \underline{second} best results are highlighted in bold and underlined, respectively. Higher values of PSNR and SSIM indicate better performance.}}
\label{tab:t1}
\renewcommand{\arraystretch}{1.1}
\resizebox{\textwidth}{!}{%
\begin{tabular}{cccc} 
\hline
\begin{tabular}{lcccc}
\multirow{2}{*}{Method} & \multicolumn{2}{c}{Snow100K-S} & \multicolumn{2}{c}{Snow100K-L} \\ 
& PSNR & SSIM & PSNR & SSIM \\
\hline
SPANet$_{\text{[CVPR'19]}}$\cite{SPANet}  & 29.92 & 0.8260 & 23.70 & 0.7930 \\
JSTASR$_{\text{[ECCV'20]}}$\cite{JSTASR}  & 31.40 & 0.9012 & 25.32 & 0.8076 \\
RESCAN$_{\text{[ECCV'18]}}$\cite{RESCAN}  & 31.51 & 0.9032 & 26.08 & 0.8408 \\
DesnowNet$_{\text{[TIP'18]}}$\cite{DesnowNet}  & 32.33 & 0.9500 & 27.17 & 0.8983 \\
DDMSNet$_{\text{[TIP'21]}}$\cite{DDMSNet}  & 34.34 & 0.9445 & 28.85 & 0.8772 \\
NAFNet$_{\text{[ECCV'22]}}$\cite{NAPNet}  & 34.79 & 0.9497 & 30.06 & 0.9017 \\
\hline
Restormer$_{\text{[CVPR'22]}}$\cite{Restormer} & 36.01 & 0.9579 & 30.36 & 0.9068 \\
All-in-One$_{\text{[CVPR'20]}}$\cite{All-in-one}  & - & - & 28.33 & 0.8820 \\
TransWeather$_{\text{[CVPR'22]}}$\cite{TransWeather}  & 32.51 & 0.9341 & 29.31 & 0.8879 \\
Chen~\textit{et al.}$_{\text{[CVPR'22]}}$~\cite{Chen}  & 34.42 & 0.9469 & 30.22 & 0.9071 \\
WGWSNet$_{\text{[CVPR'23]}}$\cite{WGWS}  & 34.31 & 0.9516 & 30.16 & 0.9007 \\
AWRCP$_{\text{[ICCV'23]}}$\cite{AWRCP}   & 36.92 & 0.9652 & 31.92 & 0.9341 \\
Histormer$_{\text{[ECCV'24]}}$\cite{Histoformer} & 37.41 & 0.9658 & 32.16 & 0.9261 \\
{GridFormer$_{\text{[IJCV'24]}}$}\cite{GridFormer}& 37.46 & 0.9640 & 31.71 & 0.9231 \\
{T3-DiffWeather$_{\text{[ECCV'24]}}$}\cite{T3-DiffWeather} & \underline{37.51} & \underline{0.9664} & \textbf{32.37} & \textbf{0.9355} \\
{CyclicPrompt$_{\text{[TIP'25]}}$}\cite{CyclicPrompt} & 37.50 & 0.9655 & 32.16 & 0.9265 \\
\textbf{DCMPC-Net (Ours)} & \textbf{38.21} & \textbf{0.9692} & \underline{32.35} & \underline{0.9302} \\
\end{tabular}
&
\begin{tabular}{lcc}
\multirow{2}{*}{Method} & \multicolumn{2}{c}{Outdoor-Rain} \\
& PSNR & SSIM \\
\hline
CycleGAN$_{\text{[ICCV'17]}}$\cite{CycleGAN}  & 17.62 & 0.6560 \\
pix2pix$_{\text{[CVPR'17]}}$\cite{pix2pix}  & 19.09 & 0.7100 \\
HRGAN$_{\text{[CVPR'19]}}$\cite{Test1}  & 21.56 & 0.8550 \\
PCNet$_{\text{[CVPR'21]}}$\cite{PCNet}  & 26.19 & 0.9015 \\
MPRNet$_{\text{[CVPR'21]}}$\cite{MPRNet}  & 28.03 & 0.9192 \\
NAFNet$_{\text{[ECCV'22]}}$\cite{NAPNet}  & 29.59 & 0.9027 \\
\hline
Restormer\cite{Restormer} & 30.03 & 0.9215 \\
All-in-One\cite{All-in-one}  & 24.71 & 0.8980 \\
TransWeather\cite{TransWeather}  & 28.83 & 0.9000 \\
Chen \textit{et al.}\cite{Chen}  & 29.27 & 0.9147 \\
WGWSNet\cite{WGWS}  & 29.32 & 0.9207 \\
AWRCP\cite{AWRCP} & 31.39 & 0.9329 \\
Histormer\cite{Histoformer} & 32.08 & \underline{0.9389} \\
{GridFormer}\cite{GridFormer} & 31.87 & 0.9335 \\
{T3-DiffWeather}\cite{T3-DiffWeather} & 31.99 & 0.9365 \\
{CyclicPrompt}\cite{CyclicPrompt} & \textbf{32.81} & 0.9371 \\
\textbf{DCMPC-Net (Ours)} & \underline{32.49} & \textbf{0.9477} \\
\end{tabular}
&
\begin{tabular}{lcc}
\multirow{2}{*}{Method} & \multicolumn{2}{c}{RainDrop} \\
& PSNR & SSIM \\
\hline
pix2pix$_{\text{[CVPR'17]}}$\cite{pix2pix}  & 28.02 & 0.8547 \\
DuRN$_{\text{[CVPR'19]}}$\cite{DuRN}  & 31.24 & 0.9259 \\
RainDropAttn$_{\text{[ICCV'19]}}$\cite{RainDropAttn} & 31.44 & 0.9264 \\
AttentiveGAN$_{\text{[CVPR'18]}}$\cite{AttentiveGAN}  & 31.59 & 0.9170 \\
IPT$_{\text{[CVPR'21]}}$\cite{IPT}  & 31.87 & 0.9313 \\
MAXIM$_{\text{[CVPR'22]}}$\cite{MAXIM}  & 31.87 & 0.9352 \\
\hline
Restormer\cite{Restormer}  & 32.18 & 0.9408 \\
All-in-One\cite{All-in-one}  & 31.12 & 0.9268 \\
TransWeather\cite{TransWeather}  & 30.17 & 0.9157 \\
Chen \textit{et al.}\cite{Chen}  & 31.81 & 0.9309 \\
WGWSNet\cite{WGWS}  & 32.38 & 0.9378 \\
AWRCP\cite{AWRCP} & 31.93 & 0.9314 \\
Histormer\cite{Histoformer}    & \underline{33.06} & 0.9441 \\
{GridFormer}\cite{GridFormer} & 32.39 & 0.9362 \\
{T3-DiffWeather}\cite{T3-DiffWeather} & 32.66 & 0.9411 \\
{CyclicPrompt}\cite{CyclicPrompt} & 32.57 & \underline{0.9454} \\
\textbf{DCMPC-Net (Ours)} & \textbf{33.08} & \textbf{0.9474} \\
\end{tabular}
&
\begin{tabular}{cc}
\multicolumn{2}{c}{Average} \\
PSNR & SSIM \\
\hline
- & - \\
- & - \\
- & - \\
- & - \\
- & - \\
- & - \\
\hline
32.14 & 0.9318    \\
- & -     \\
30.21 & 0.9094  \\
31.43 & 0.9249  \\
31.54 & 0.9277  \\
33.04 & 0.9409  \\
33.68 & 0.9437  \\
33.36 & 0.9392  \\
33.63 & \underline{0.9449}  \\
\underline{33.76} & 0.9436 \\
\textbf{34.03} & \textbf{0.9486} \\
\end{tabular}
\\
\hline
\end{tabular}%
}
\end{table*}

\begin{figure*}[!t]\footnotesize
\begin{center}
\begin{tabular}{cccccc}
\hspace{-1.5mm}\includegraphics[width = 0.163\linewidth]{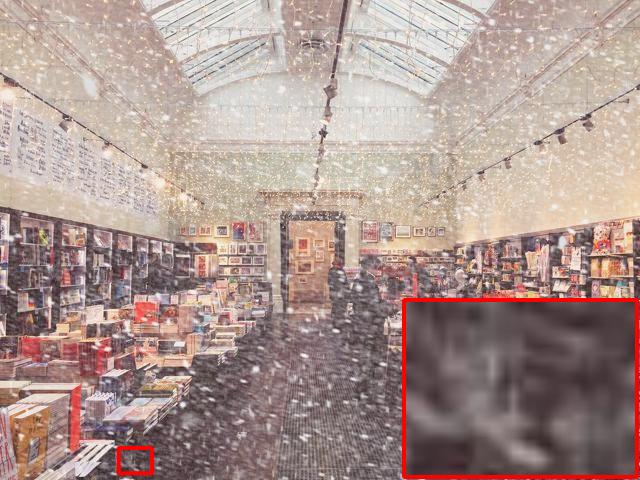} &\hspace{-4.5mm}
\includegraphics[width = 0.163\linewidth]{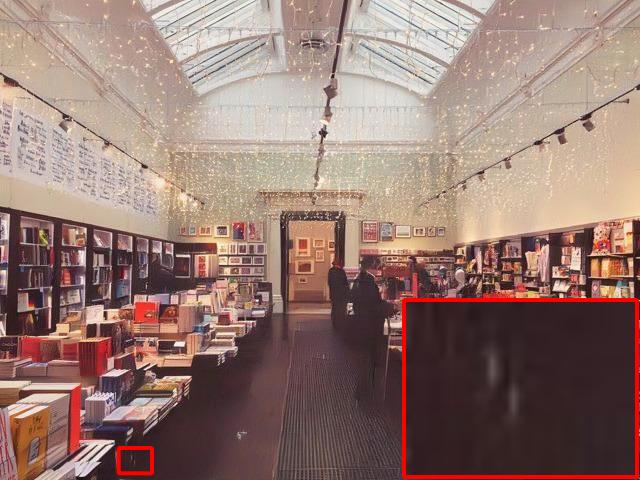}  &\hspace{-4.5mm}
\includegraphics[width = 0.163\linewidth]{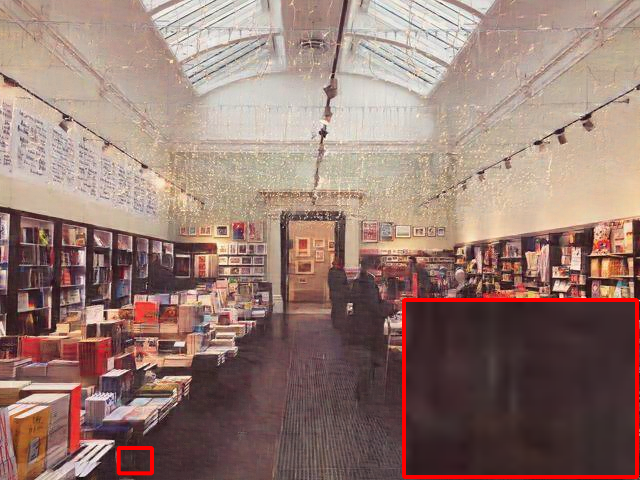} &\hspace{-4.5mm}
\includegraphics[width = 0.163\linewidth]{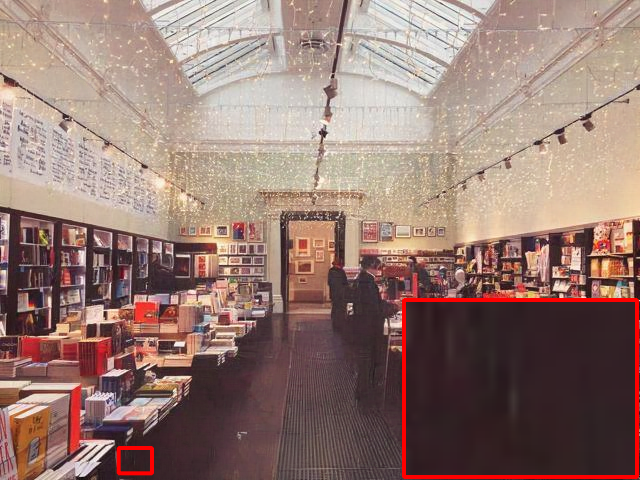} &\hspace{-4.5mm}
\includegraphics[width = 0.163\linewidth]{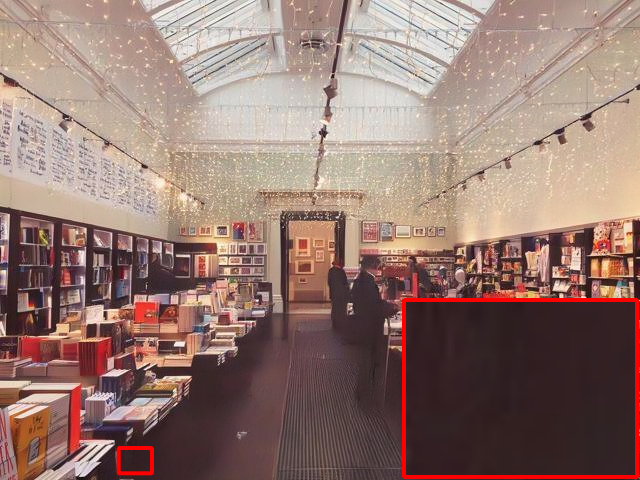} &\hspace{-4.5mm}
\includegraphics[width = 0.163\linewidth]{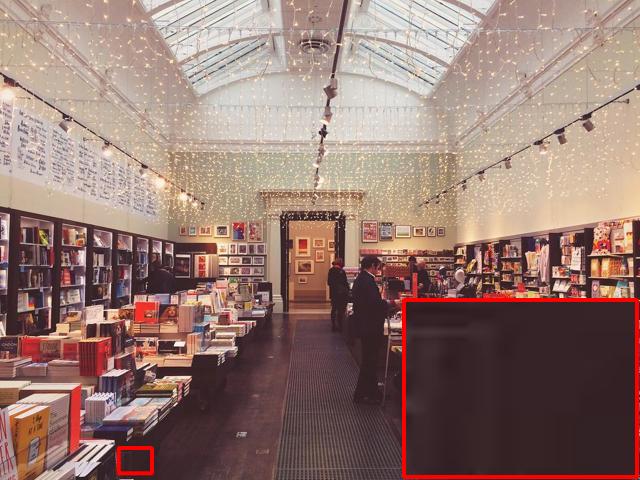} 
\\
\hspace{-1.5mm}\includegraphics[width = 0.163\linewidth]{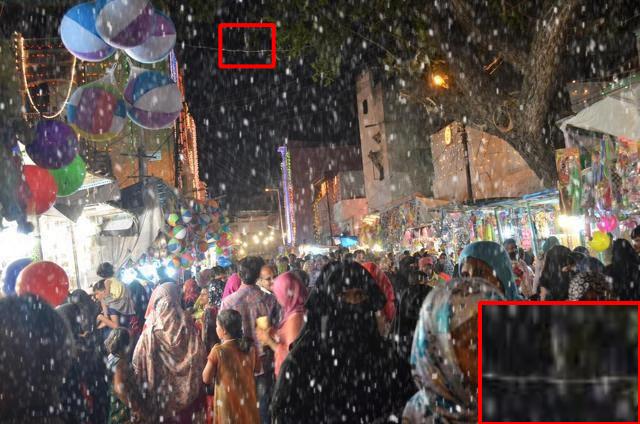} &\hspace{-4.5mm}
\includegraphics[width = 0.163\linewidth]{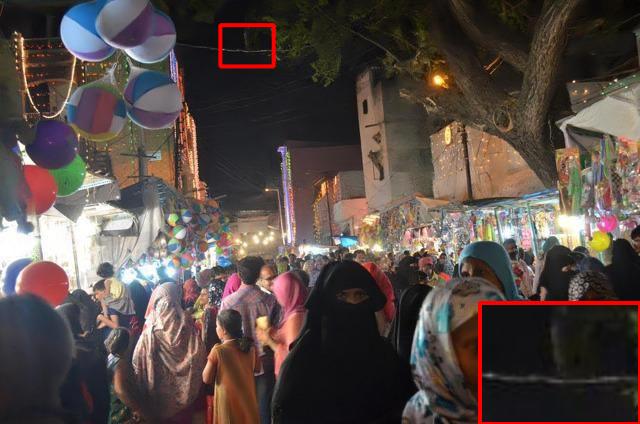} &\hspace{-4.5mm}
\includegraphics[width = 0.163\linewidth]{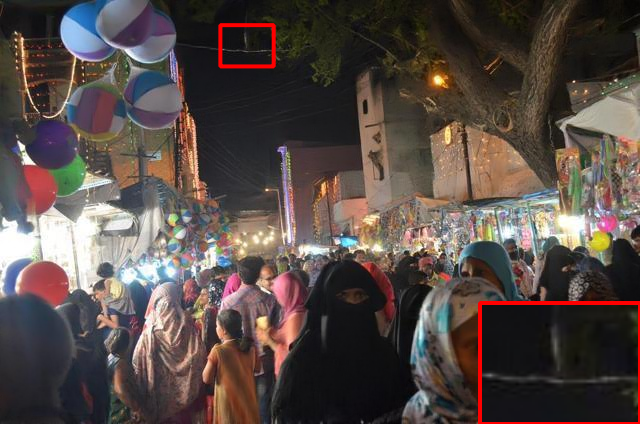} &\hspace{-4.5mm}
\includegraphics[width = 0.163\linewidth]{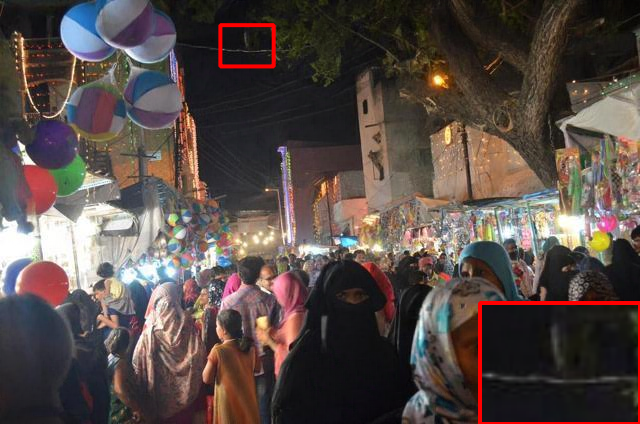}  &\hspace{-4.5mm}
\includegraphics[width = 0.163\linewidth]{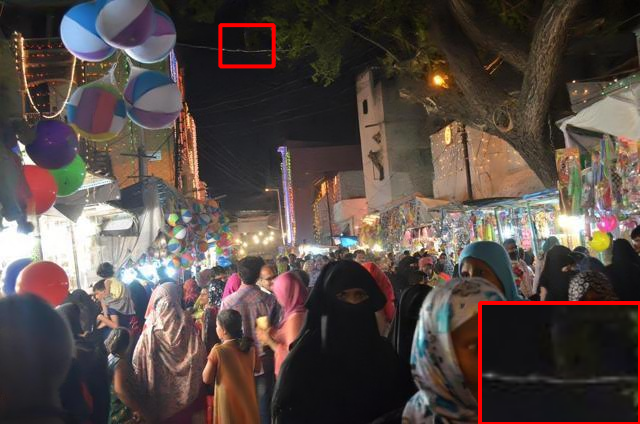} &\hspace{-4.5mm}
\includegraphics[width = 0.163\linewidth]{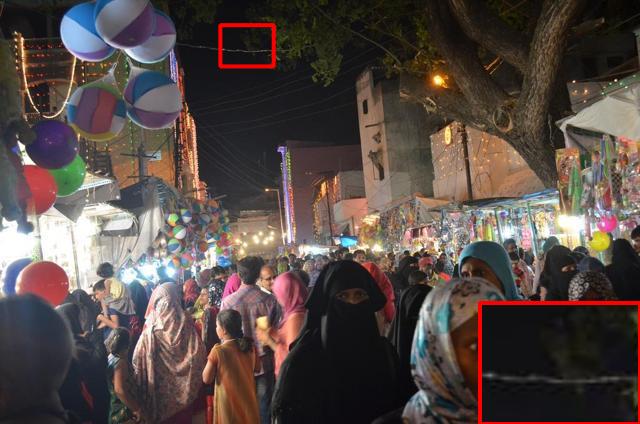} 
\\
\hspace{-1.5mm}\includegraphics[width = 0.163\linewidth]{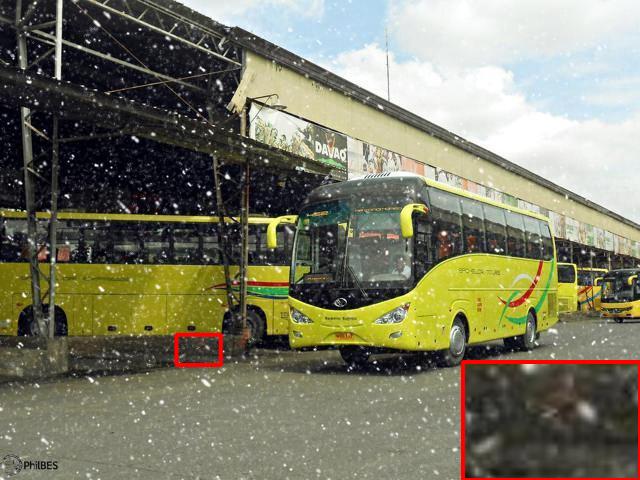} &\hspace{-4.5mm}
\includegraphics[width = 0.163\linewidth]{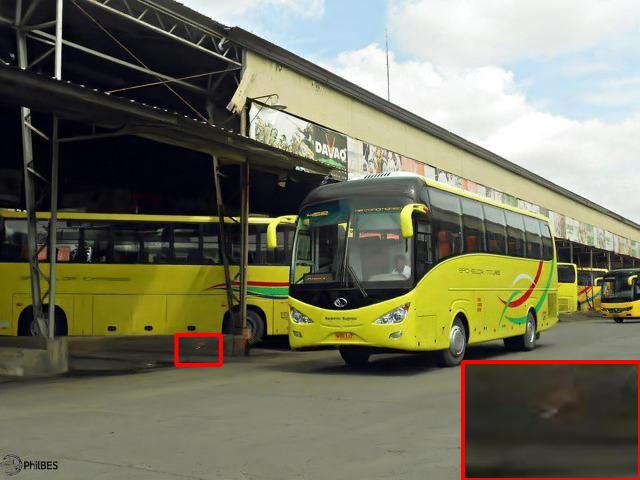} &\hspace{-4.5mm}
\includegraphics[width = 0.163\linewidth]{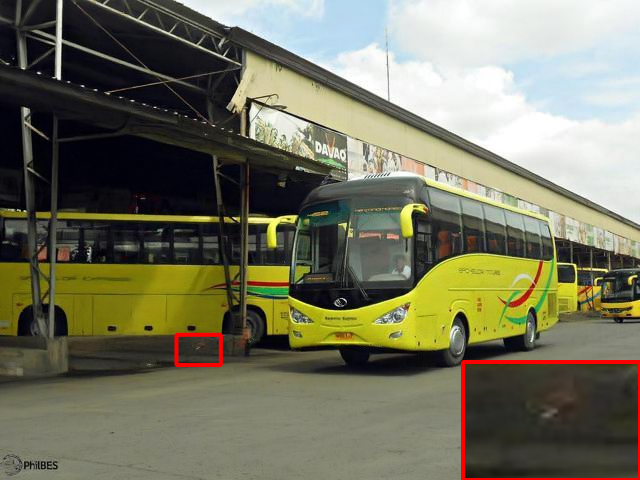} &\hspace{-4.5mm}
\includegraphics[width = 0.163\linewidth]{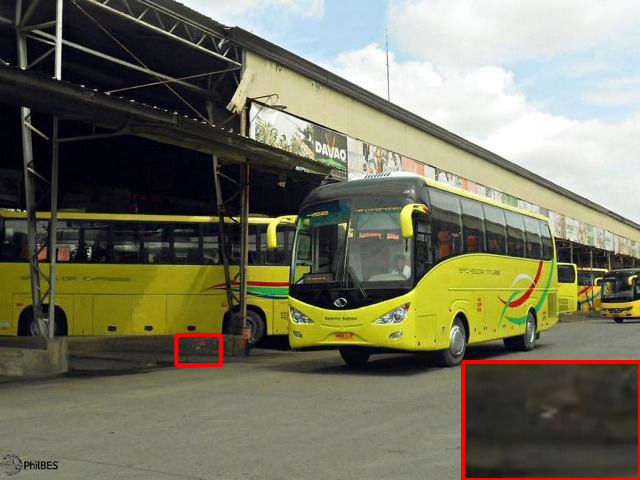} &\hspace{-4.5mm}
\includegraphics[width = 0.163\linewidth]{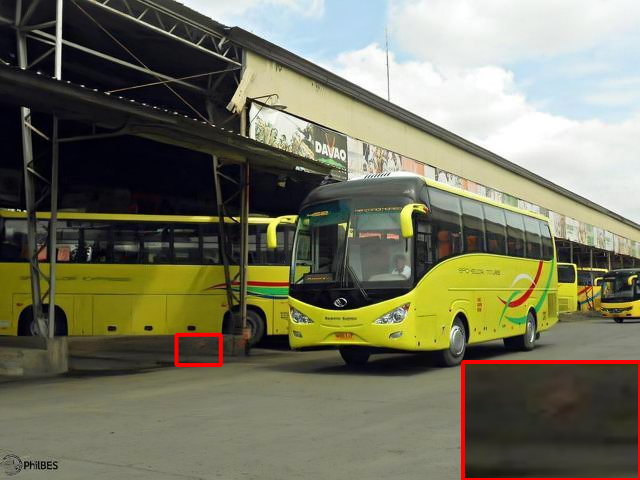}  &\hspace{-4.5mm}
\includegraphics[width = 0.163\linewidth]{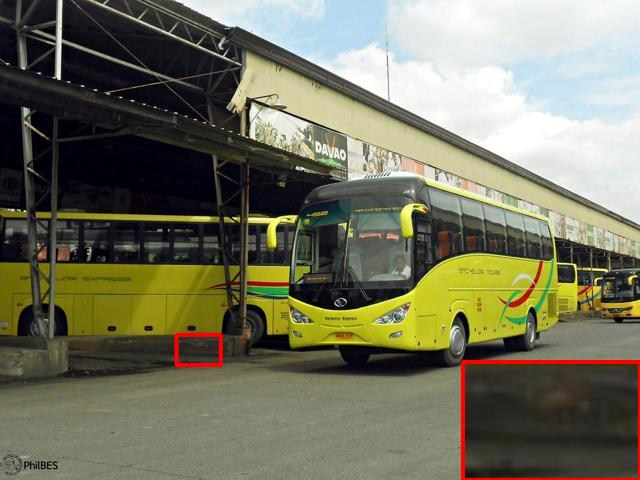} 
\\
\hspace{-1.5mm} 
\includegraphics[width = 0.163\linewidth]{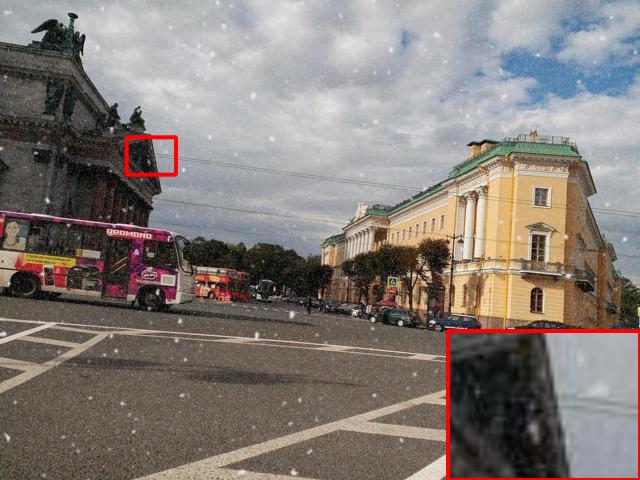} &\hspace{-4.5mm}
\includegraphics[width = 0.163\linewidth]{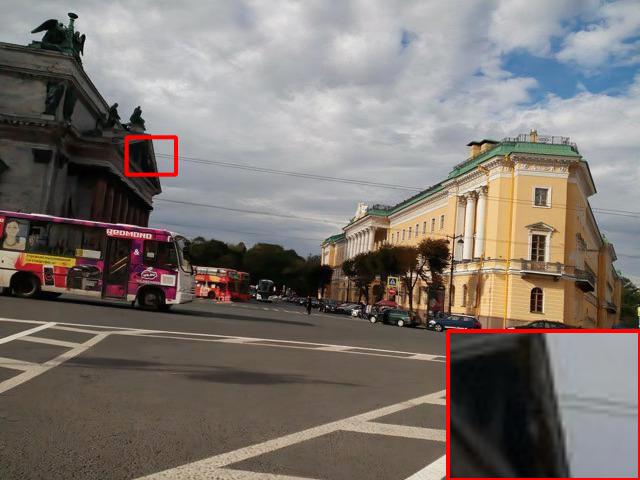} &\hspace{-4.5mm}
\includegraphics[width = 0.163\linewidth]{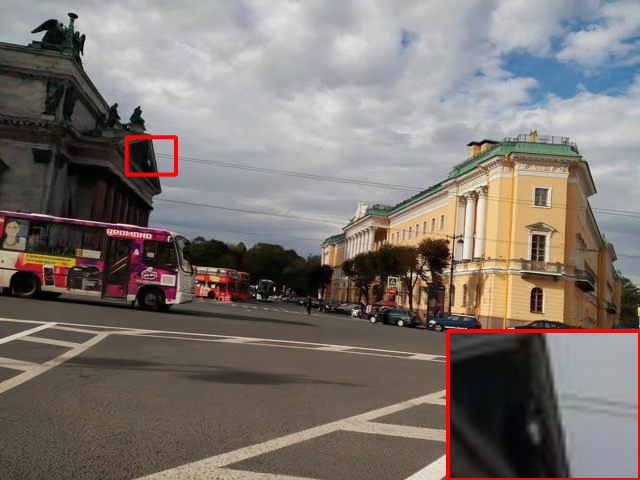} &\hspace{-4.5mm}
\includegraphics[width = 0.163\linewidth]{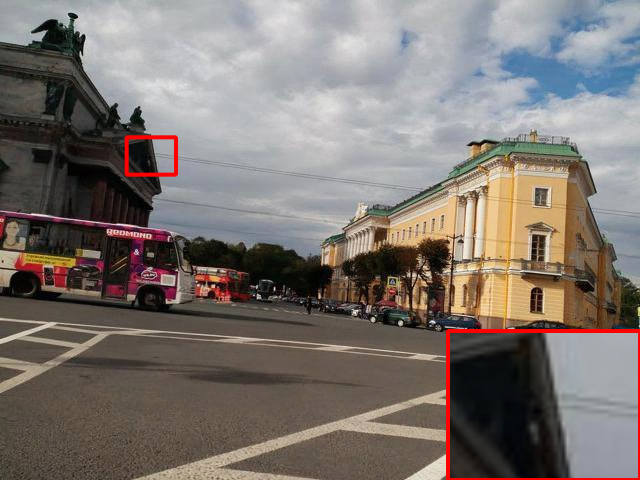}  &\hspace{-4.5mm}
\includegraphics[width = 0.163\linewidth]{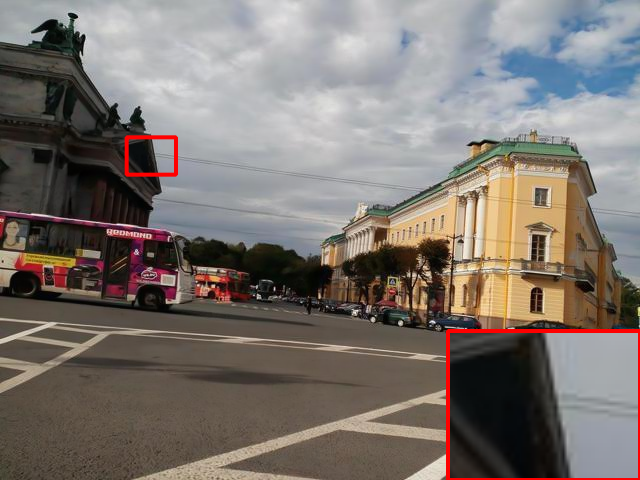} &\hspace{-4.5mm}
\includegraphics[width = 0.163\linewidth]{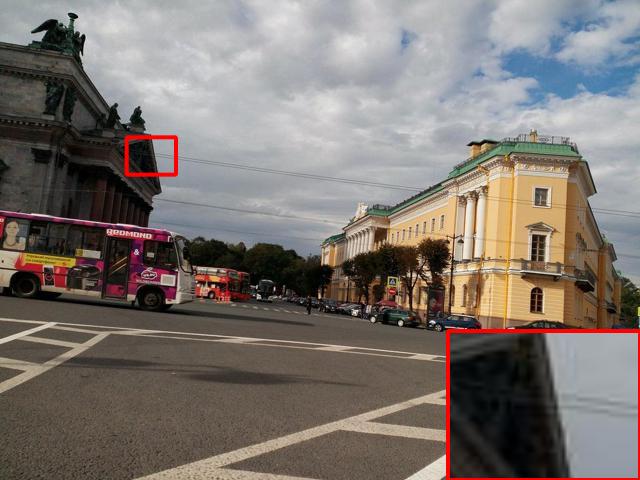} 

\\
\hspace{-1.5mm}(a) Input   &\hspace{-4.5mm} (b) Histoformer &\hspace{-4.5mm} {(c) T3-DiffWeather}&\hspace{-4.5mm} {(d) GridFormer}&\hspace{-4.5mm} (e)   Ours&\hspace{-4.5mm} (f) GT
\end{tabular}
\end{center}
\vspace{-2mm}
\caption{Visual comparison with SOTA methods on the Snow-100k test set~\cite{DesnowNet},
where red boxes indicate regions with visible improvement. 
DCMPC-Net produces cleaner results with reduced snow residues and better recovers image details compared to other competing methods.}
\label{fig:p5}
\end{figure*}

%

\subsection{Evaluation Metrics}
For simulated experiments, two widely used full-reference indicators are used: \textbf{P}eak \textbf{S}ignal-to-\textbf{N}oise \textbf{R}atio (PSNR)~\cite{PSNR} and \textbf{S}tructural \textbf{S}imilarity \textbf{I}dex \textbf{M}easurement (SSIM)~\cite{SSIM}, which are commonly used to measure the image quality in the computer vision community. 
%

\subsection{Implementation Details} 
We train models with AdamW~\cite{Adam} optimizer with initial learning rate $3 \times 10^{-4}$ gradually reduced to $1 \times 10^{-6}$ with the cosine annealing~\cite{cosine}. 
The patch size and batch size are set as $128 \times 128$ and $4$. 
For downsampling and upsampling, we adopt pixel-unshuffle and pixel-shuffle, respectively. 
For the pre-trained vision-language model, we employ LLaMA~\cite{Llama3} to generate degradation-related textual descriptions and use Multilingual E5~\cite{Text_Encoder} as the text encoder. 
The PVL and the text encoder are kept frozen and used only for semantic-prior extraction. 
For the fixed benchmark datasets used in our experiments, semantic prompt embeddings are pre-computed from degraded input images with a fixed textual query and then directly loaded during training and testing. 
Therefore, under this cached-prompt evaluation protocol, the frozen PVL and text encoder are not executed during the timed restoration forward pass.
%
All experiments are trained on four RTX 3090s. 
{Consistent with established image restoration protocols~\cite{DRSformer,Restormer,Histoformer}, we compute the PSNR and SSIM metrics on the luminance (Y) channel of the YCbCr color space.} 
To better constrain the training of DCMPC-Net, we incorporate the Fast Fourier Transform loss function with default parameters, in addition to those used in the baseline setting.

\begin{figure*}[!t]\footnotesize
\begin{center}
\begin{tabular}{cccccc}
\hspace{-1.5mm}\includegraphics[width = 0.163\linewidth]{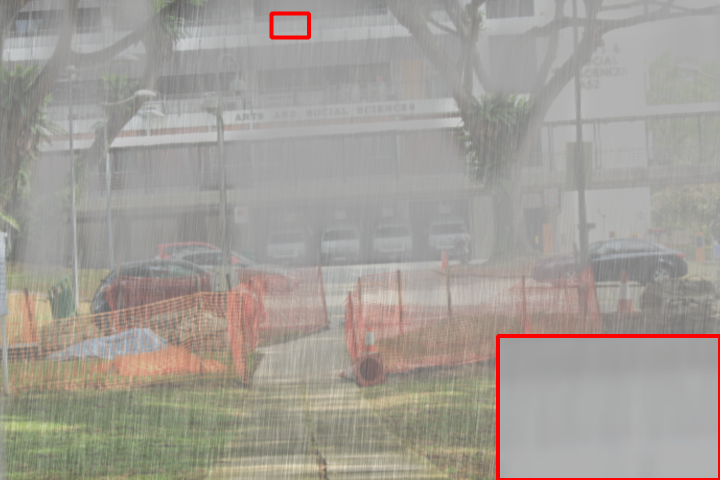} &\hspace{-4.5mm}
\includegraphics[width = 0.163\linewidth]{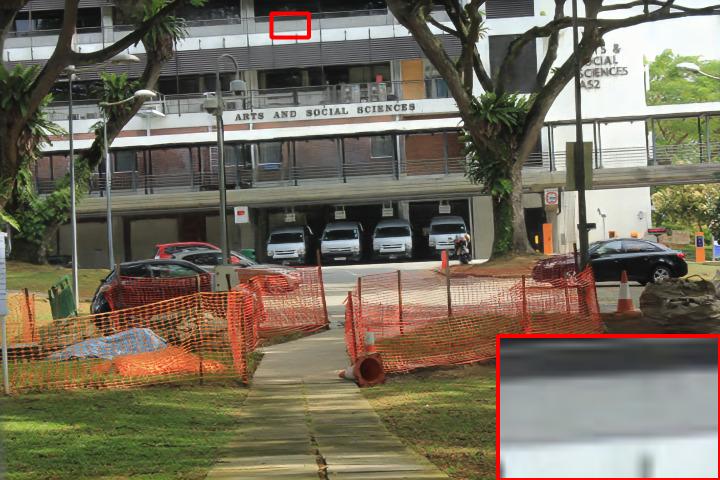}  &\hspace{-4.5mm}
\includegraphics[width = 0.163\linewidth]{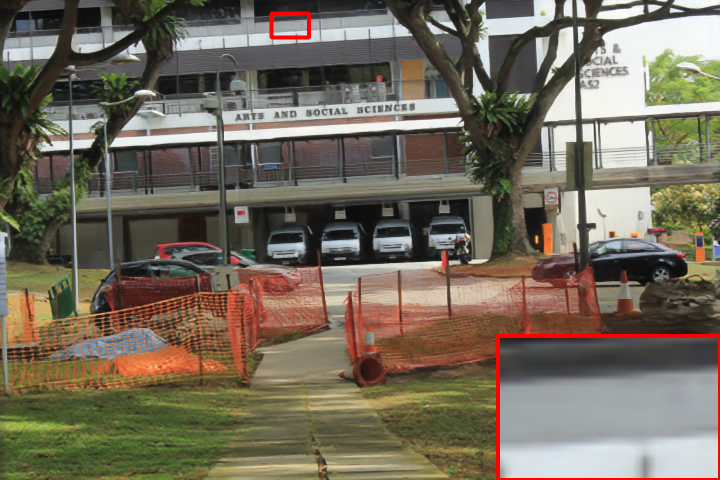} &\hspace{-4.5mm}
\includegraphics[width = 0.163\linewidth]{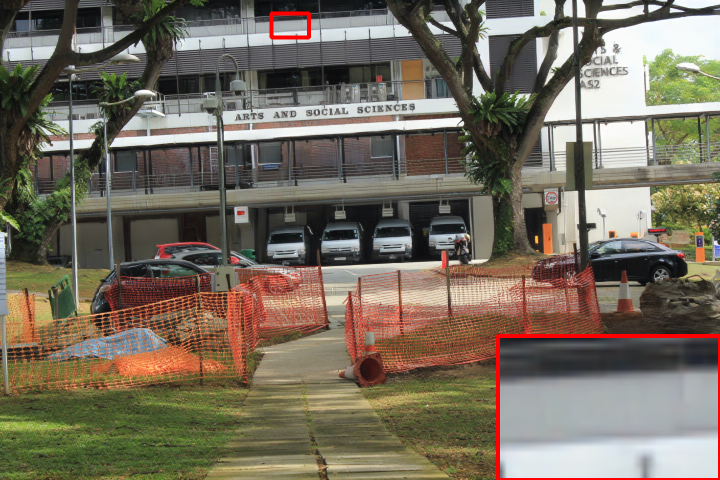} &\hspace{-4.5mm}

\includegraphics[width = 0.163\linewidth]{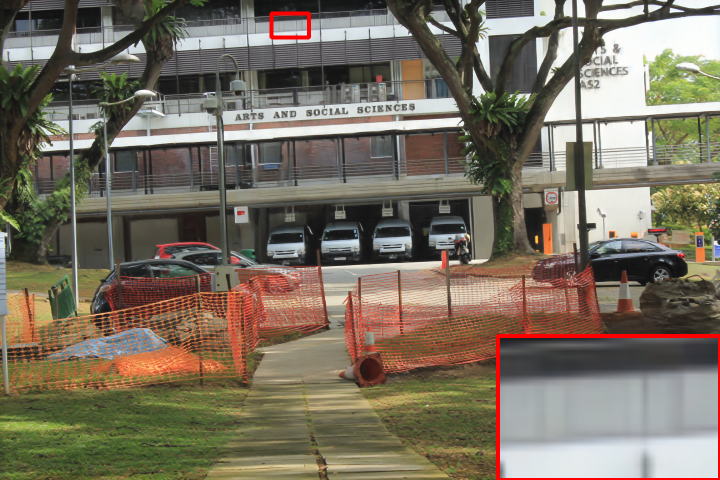} &\hspace{-4.5mm}
\includegraphics[width = 0.163\linewidth]{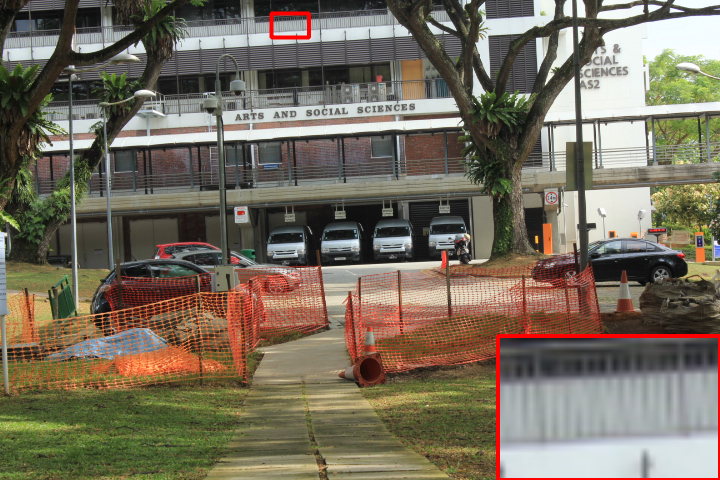} 
\\
\hspace{-1.5mm}\includegraphics[width = 0.163\linewidth]{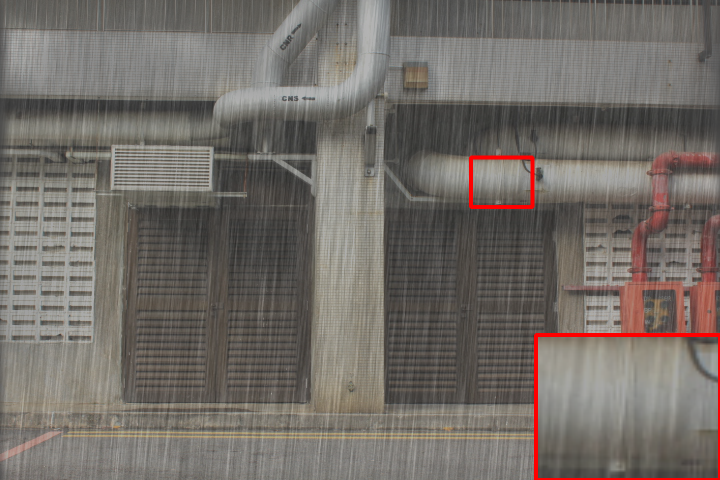} &\hspace{-4.5mm}
\includegraphics[width = 0.163\linewidth]{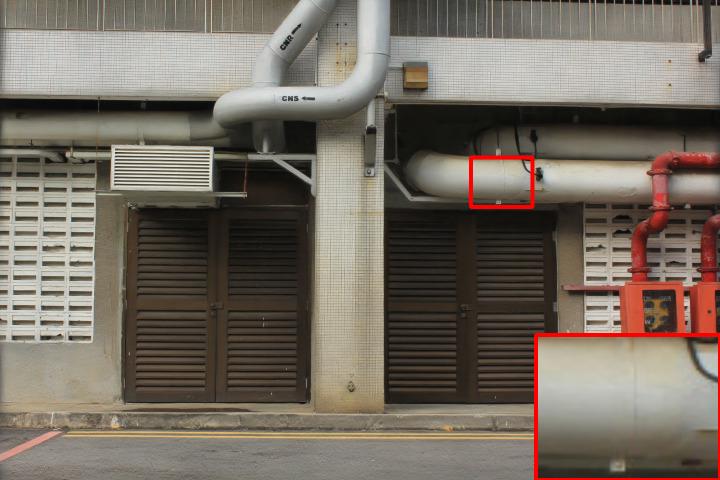} &\hspace{-4.5mm}
\includegraphics[width = 0.163\linewidth]{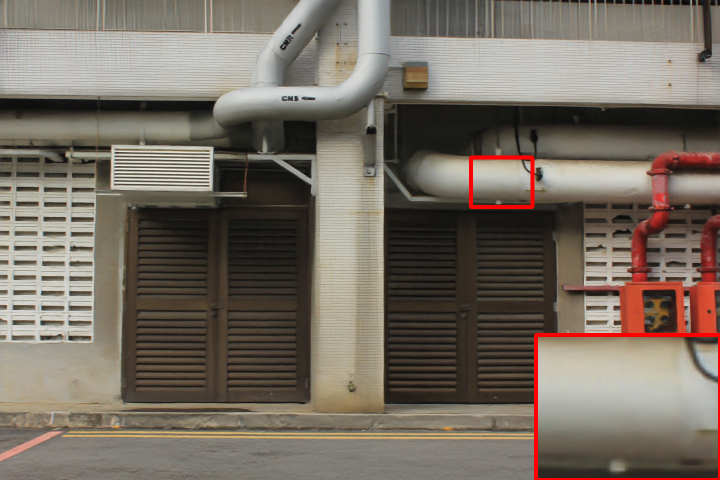} &\hspace{-4.5mm}
\includegraphics[width = 0.163\linewidth]{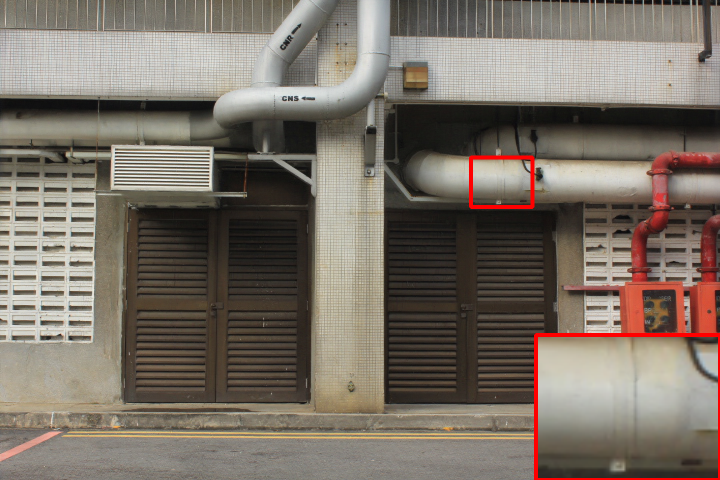} &\hspace{-4.5mm}
\includegraphics[width = 0.163\linewidth]{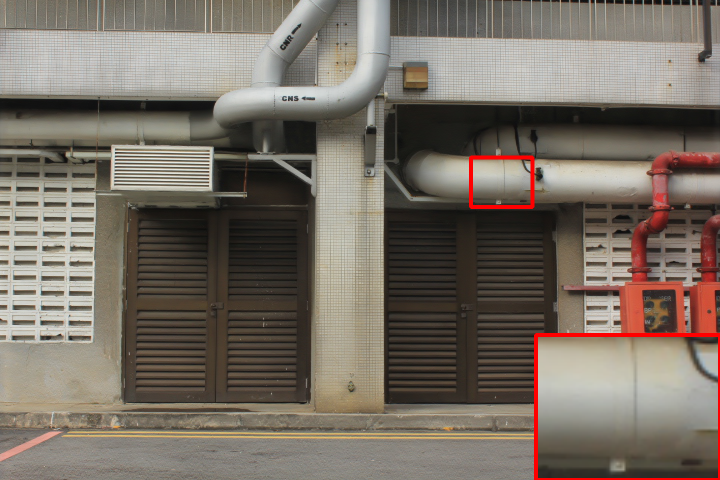}  &\hspace{-4.5mm}
\includegraphics[width = 0.163\linewidth]{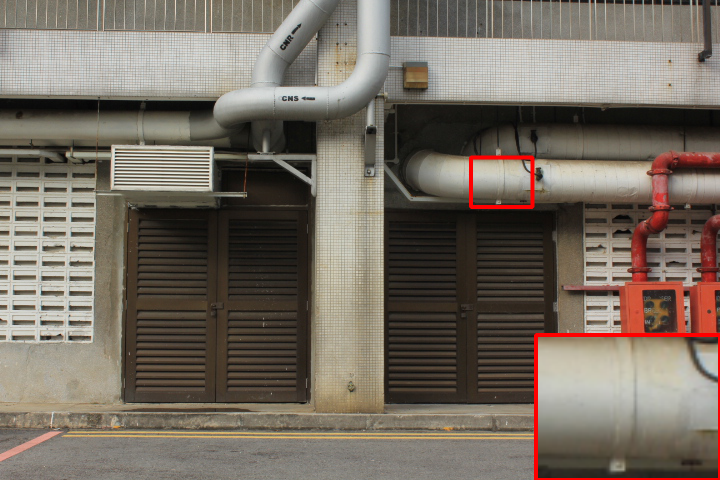} 
\\
\hspace{-1.5mm}\includegraphics[width = 0.163\linewidth]{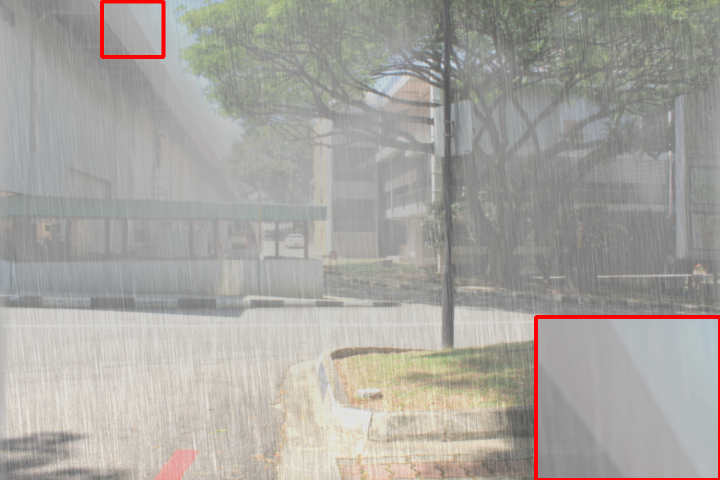} &\hspace{-4.5mm}
\includegraphics[width = 0.163\linewidth]{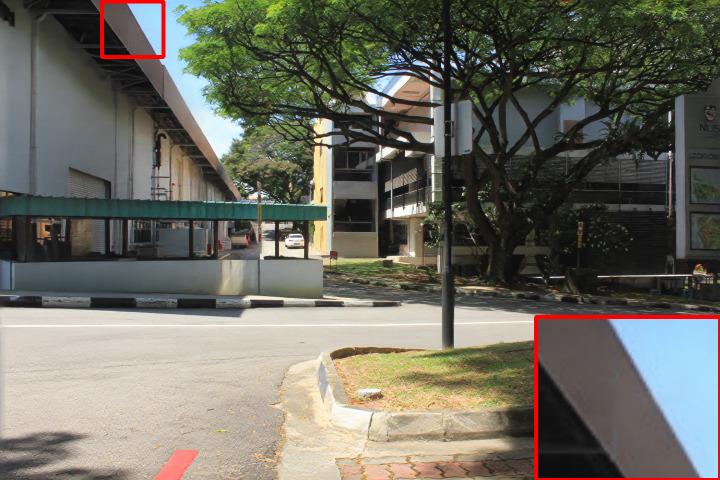} &\hspace{-4.5mm}
\includegraphics[width = 0.163\linewidth]{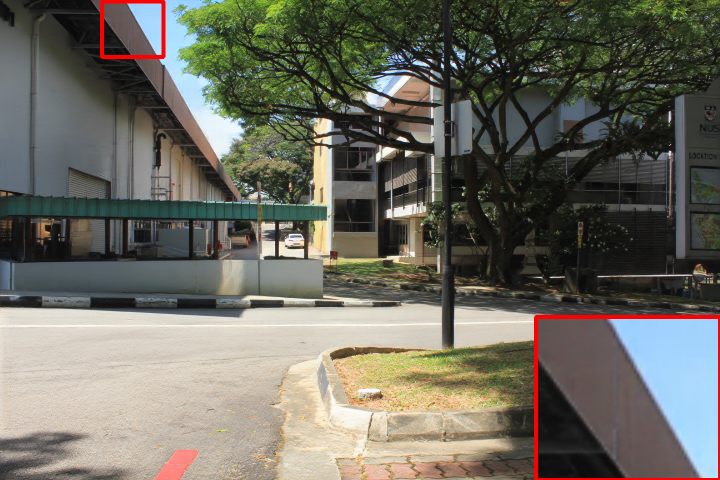} &\hspace{-4.5mm}
\includegraphics[width = 0.163\linewidth]{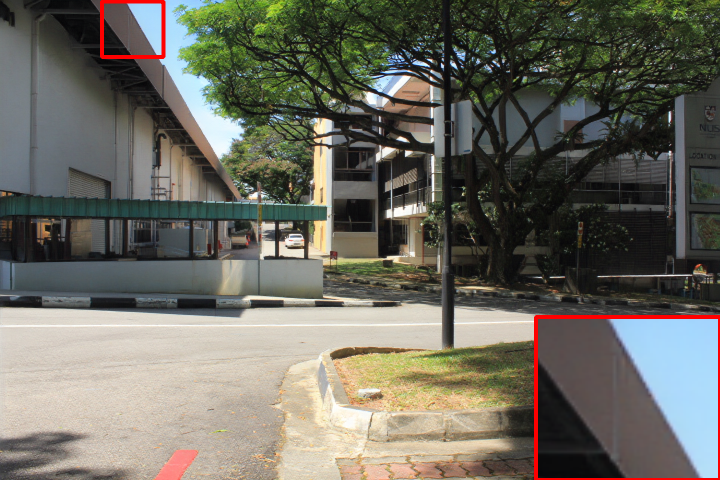} &\hspace{-4.5mm}
\includegraphics[width = 0.163\linewidth]{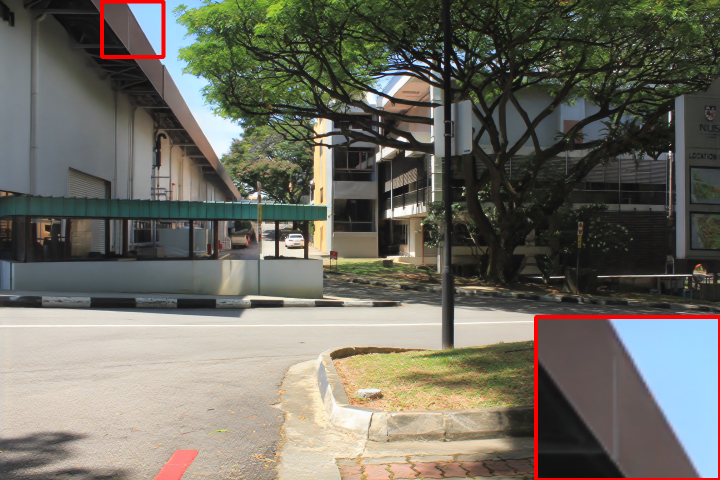}  &\hspace{-4.5mm}
\includegraphics[width = 0.163\linewidth]{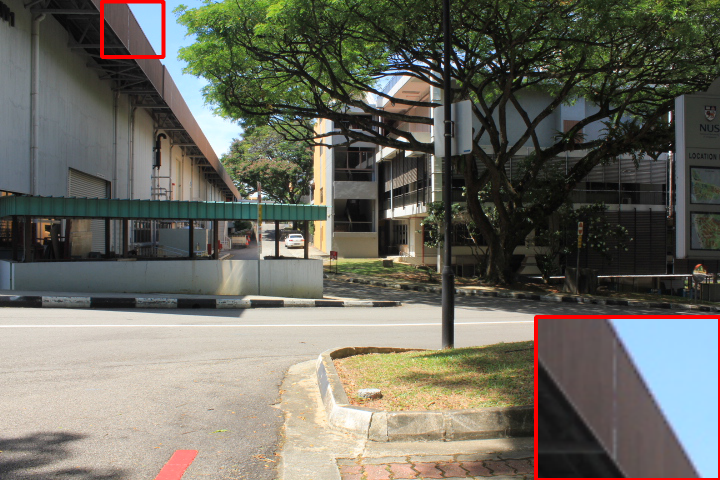} 
\\
\hspace{-1.5mm}\includegraphics[width = 0.163\linewidth]{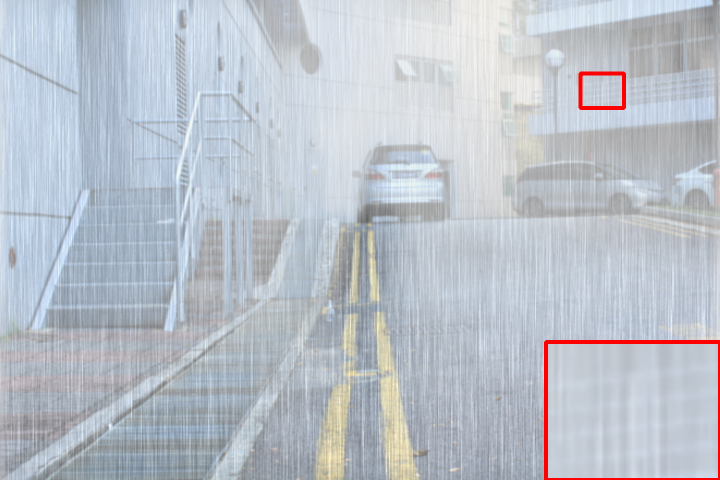} &\hspace{-4.5mm}
\includegraphics[width = 0.163\linewidth]{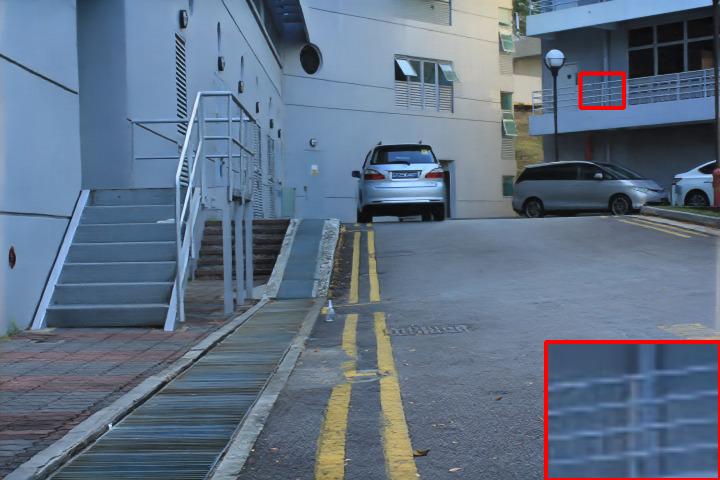} &\hspace{-4.5mm}
\includegraphics[width = 0.163\linewidth]{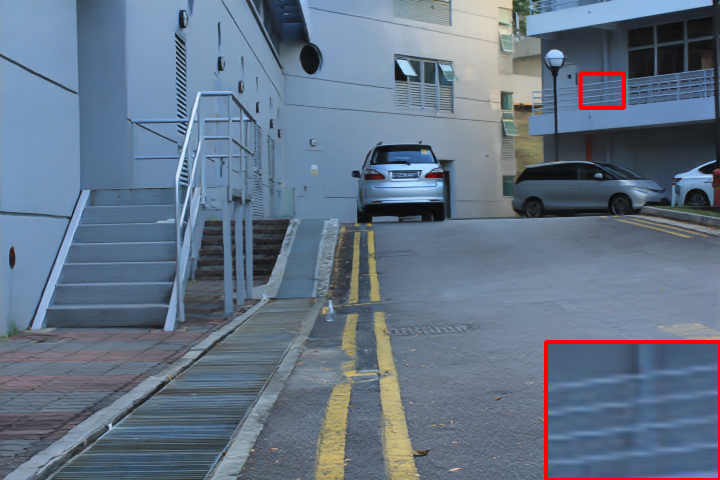} &\hspace{-4.5mm}
\includegraphics[width = 0.163\linewidth]{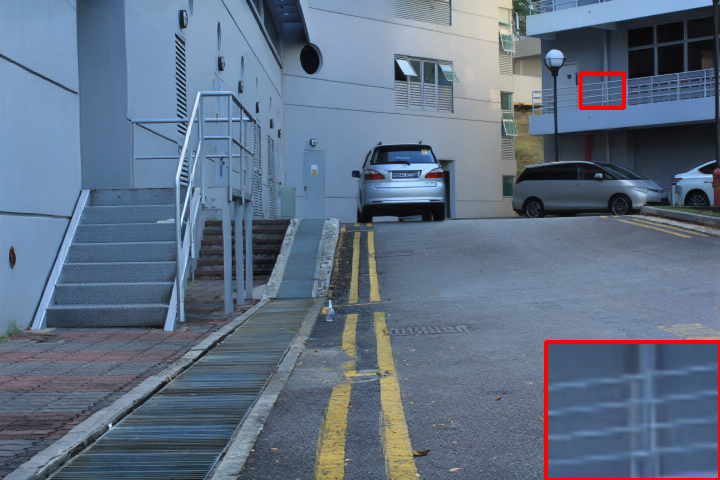} &\hspace{-4.5mm}
\includegraphics[width = 0.163\linewidth]{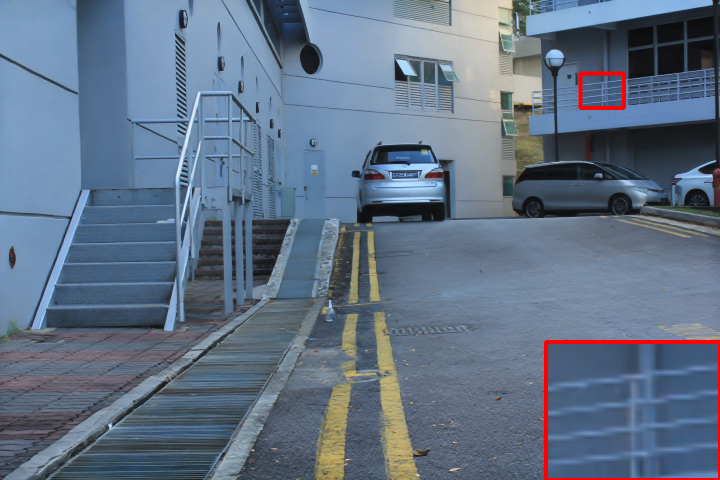}  &\hspace{-4.5mm}
\includegraphics[width = 0.163\linewidth]{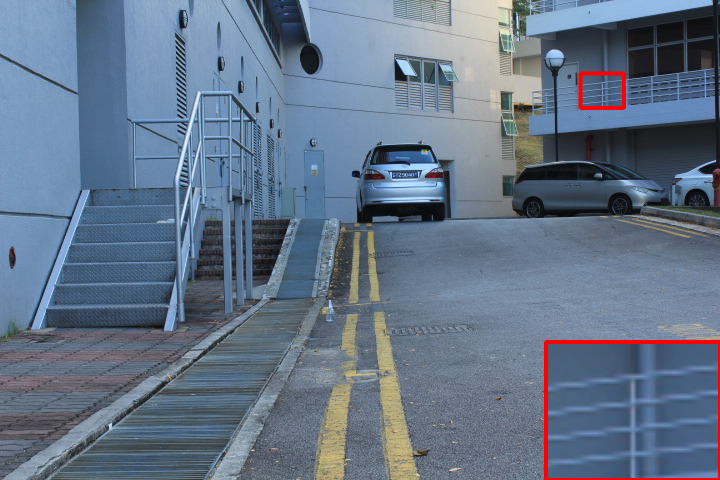} 
\\
\hspace{-1.5mm}(a) Input   &\hspace{-4.5mm} (b) Histoformer &\hspace{-4.5mm} 
{(c) T3-DiffWeather}&\hspace{-4.5mm} 
{(d) GridFormer}&\hspace{-4.5mm} (e)   Ours&\hspace{-4.5mm} (f) GT
\end{tabular}
\end{center}
\vspace{-2mm}
\caption{Visual comparison with SOTA methods on the OutdoorRain test set~\cite{Test1}. Red boxes highlight regions with notable differences. DCMPC-Net more effectively removes rain streaks and suppresses artifacts, resulting in cleaner backgrounds and improved detail preservation compared to other methods.}
\label{fig:p6}
\end{figure*}

\subsection{Experimental Results}
We evaluate our approach by conducting comprehensive comparisons with representative and state-of-the-art specialized solutions across three distinct categories of adverse weather effect removal.
For snow elimination, we compare our results against five representative snow removal networks: SPANet~\cite{SPANet}, JSTASR~\cite{JSTASR}, RESCAN~\cite{RESCAN}, DesnowNet~\cite{DesnowNet}, and DDMSNet~\cite{DDMSNet}. In addressing combined rain\&fog effects, we contrast our performance with CycleGAN~\cite{CycleGAN}, pix2pix~\cite{pix2pix}, HRGAN~\cite{Test1}, PCNet~\cite{PCNet} and MPRNet~\cite{MPRNet}. 
For handling raindrop effects, we benchmark against several dedicated solutions, including pix2pix~\cite{pix2pix}, DuRN~\cite{DuRN}, RaindropAttn~\cite{RainDropAttn}, AttentiveGAN~\cite{AttentiveGAN}, and IPT~\cite{IPT}. 
We also include comparisons with recent advanced networks that use Transformer architecture or handle multiple types of image degradation, specifically NAFNet~\cite{NAPNet}, MAXIM~\cite{MAXIM}, and Restormer~\cite{Restormer}. 
Furthermore, we conduct performance comparison with the All-in-one~\cite{All-in-one}, Transweather~\cite{TransWeather}, Chen et al.~\cite{Chen}, WGWSNet~\cite{WGWS}, AWRCP~\cite{AWRCP}, Histoformer~\cite{Histoformer}, GridFormer~\cite{GridFormer}, T3-DiffWeather~\cite{T3-DiffWeather}
and CyclicPrompt\cite{CyclicPrompt}.
\subsubsection{Quantitative Evaluation}
We compare DCMPC-Net with state-of-the-art methods on four adverse weather test sets: Snow100K-S~\cite{DesnowNet}, Snow100K-L~\cite{DesnowNet}, OutdoorRain~\cite{Test1}, and RainDrop~\cite{RainDropAttn}.
For a fair comparison, we report the quantitative results of competing methods as reported in the original papers.
%
As shown in Table~\ref{tab:t1}, DCMPC-Net achieves the best average performance across the four benchmark datasets, outperforming the second-best method by 0.27 dB in PSNR and obtaining the highest average SSIM of 0.9486.
These results demonstrate the effectiveness of DCMPC-Net across diverse adverse weather benchmarks.
We further compare DCMPC-Net with LDR~\cite{Language-driven}, a representative language-driven adverse weather restoration method. 
For LDR, PSNR and SSIM are computed from publicly available full-resolution restoration results in its official repository on Snow100K-L~\cite{DesnowNet}, Outdoor-Rain~\cite{Test1}, and RainDrop~\cite{RainDropAttn}, using the same evaluation protocol.
As shown in Table~\ref{tab:LDR}, DCMPC-Net achieves higher average PSNR and SSIM, with consistently higher SSIM across the three test sets.

\begin{table}[!t]
\centering
\caption{
Quantitative comparison with LDR~\cite{Language-driven} on Snow100K-L~\cite{DesnowNet}, Outdoor-Rain~\cite{Test1}, and RainDrop~\cite{RainDropAttn} in terms of PSNR and SSIM.
}
\label{tab:LDR}

{
\small
\setlength{\tabcolsep}{4pt}
\renewcommand{\arraystretch}{0.95}

\resizebox{\columnwidth}{!}{%
\begin{tabular}{c c c c c c}
\toprule
Method & Metric & Snow100K-L & OutdoorRain & RainDrop & Average \\
\midrule

\multirow{2}{*}{LDR~\cite{Language-driven}}
& PSNR & 32.47 & 30.92 & 33.83 & 32.41 \\
& SSIM & 0.9281 & 0.9264 & 0.9415 & 0.9320 \\

\midrule

\multirow{2}{*}{DCMPC-Net}
& PSNR & 32.35 & 32.49 & 33.08 & 32.64 \\
& SSIM & 0.9302 & 0.9477 & 0.9474 & 0.9418 \\

\bottomrule
\end{tabular}%
}
}
\vspace{-3mm}
\end{table}


 \begin{figure*}[!t]\footnotesize
\begin{center}
\begin{tabular}{cccccc}
\hspace{-1.5mm}\includegraphics[width = 0.163\linewidth]{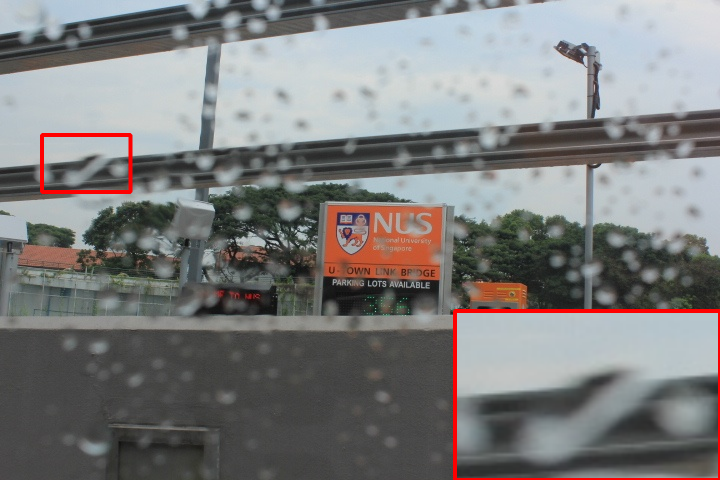} &\hspace{-4.5mm}
\includegraphics[width = 0.163\linewidth]{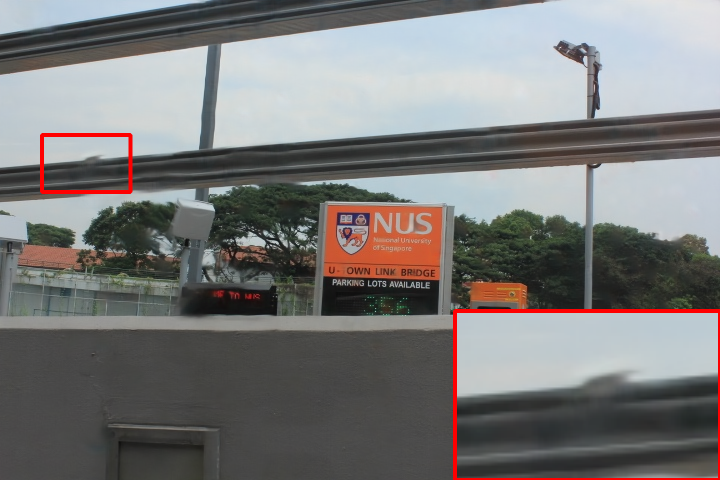}  &\hspace{-4.5mm}
\includegraphics[width = 0.163\linewidth]{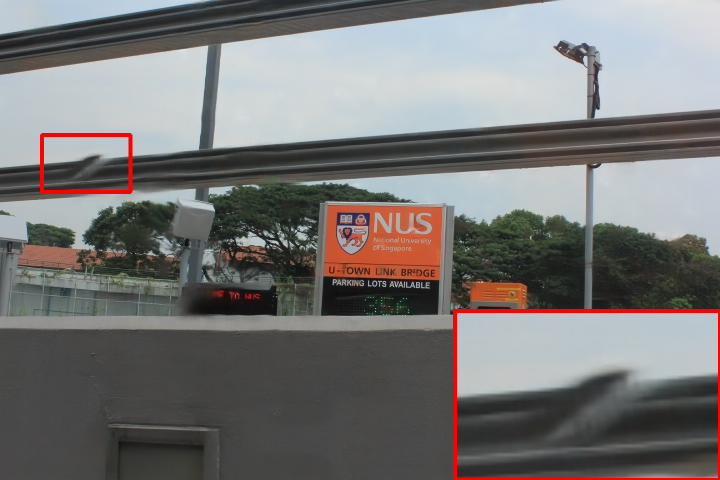} &\hspace{-4.5mm}
\includegraphics[width = 0.163\linewidth]{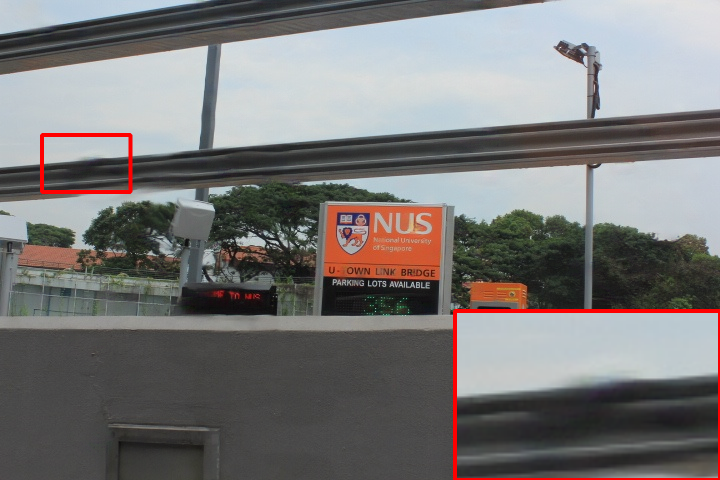} &\hspace{-4.5mm}
\includegraphics[width = 0.163\linewidth]{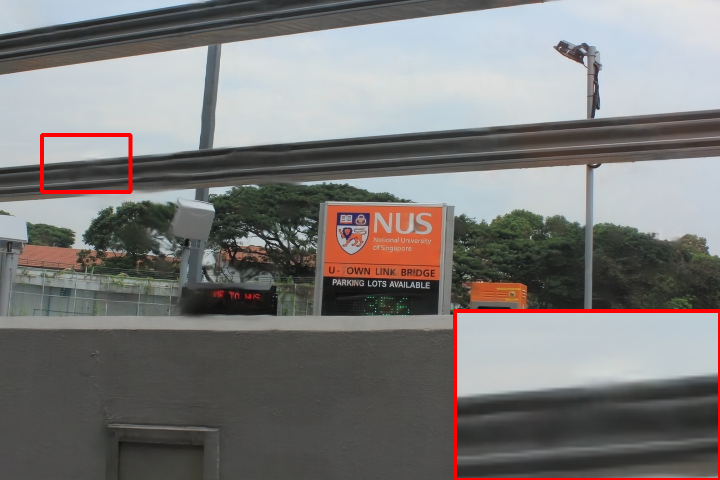} &\hspace{-4.5mm}
\includegraphics[width = 0.163\linewidth]{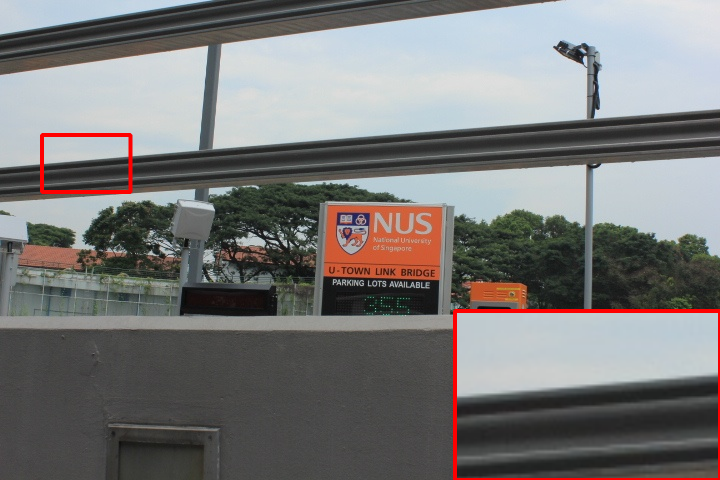} 
\\
\hspace{-1.5mm}\includegraphics[width = 0.163\linewidth]{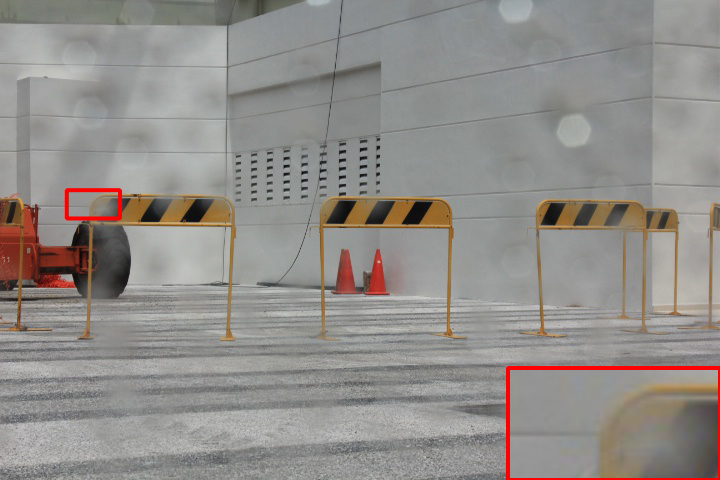} &\hspace{-4.5mm}
\includegraphics[width = 0.163\linewidth]{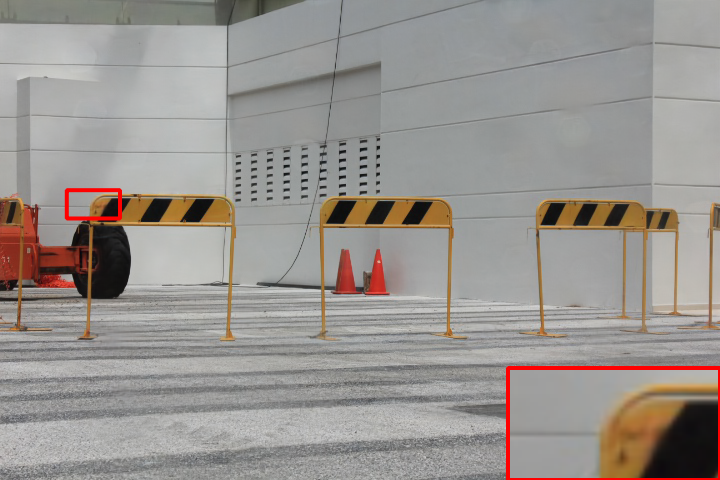}  &\hspace{-4.5mm}
\includegraphics[width = 0.163\linewidth]{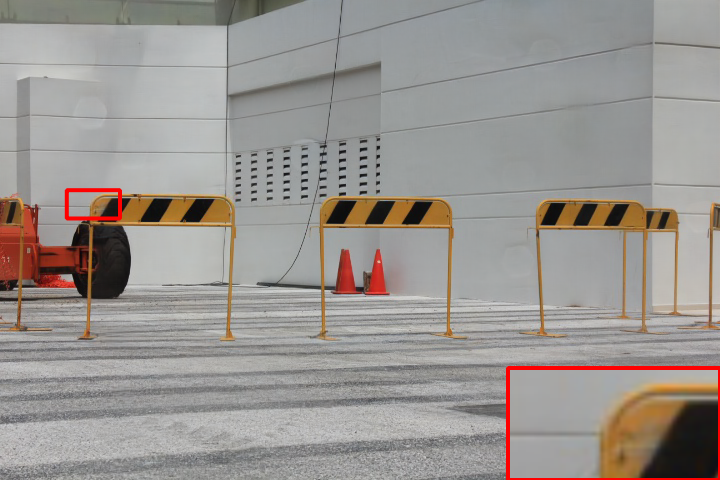} &\hspace{-4.5mm}
\includegraphics[width = 0.163\linewidth]{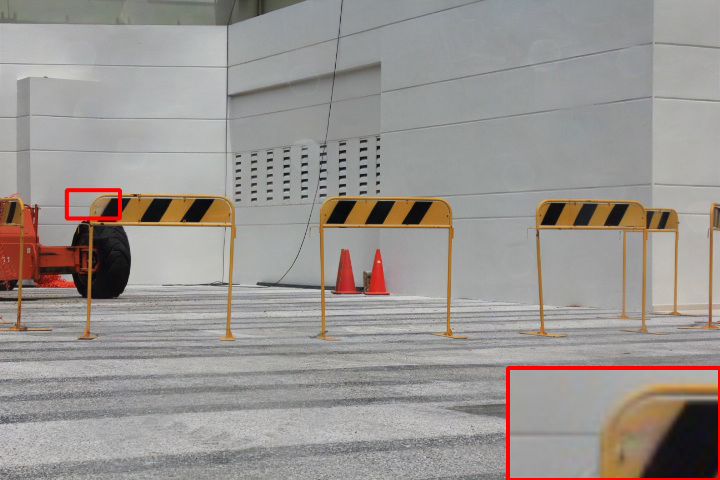} &\hspace{-4.5mm}
\includegraphics[width = 0.163\linewidth]{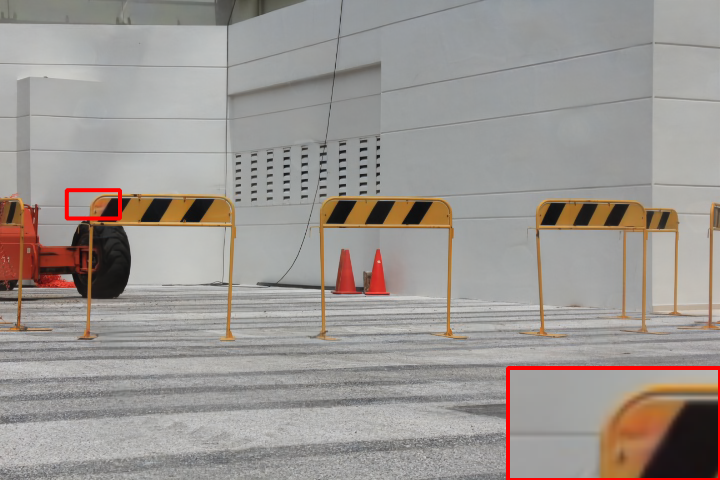} &\hspace{-4.5mm}
\includegraphics[width = 0.163\linewidth]{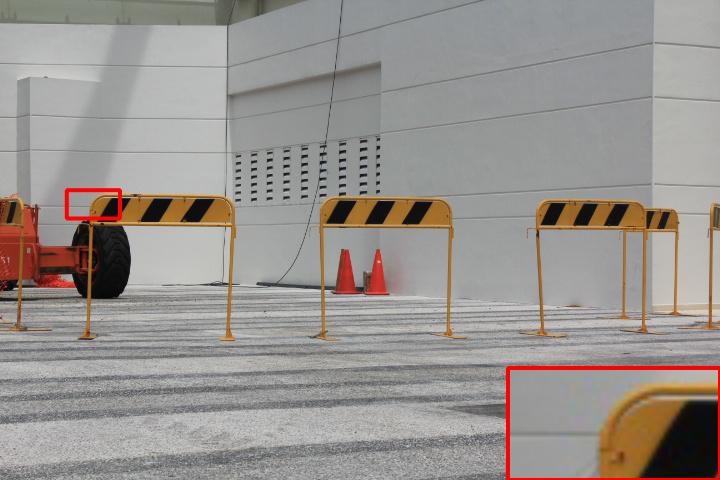} 
\\
\hspace{-1.5mm}\includegraphics[width = 0.163\linewidth]{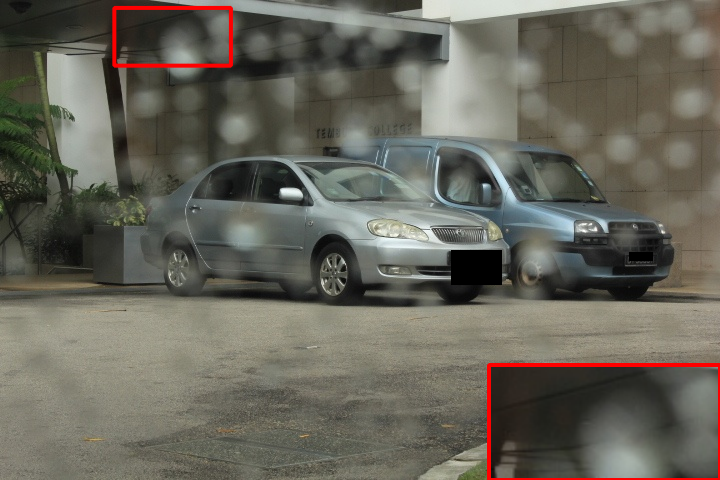} &\hspace{-4.5mm}
\includegraphics[width = 0.163\linewidth]{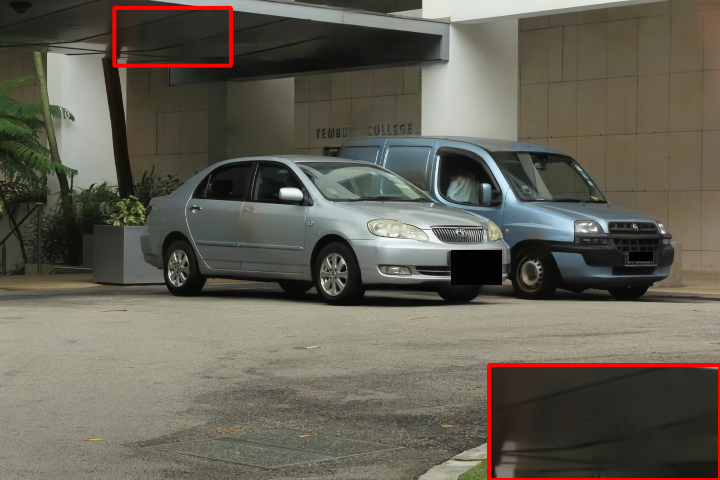}  &\hspace{-4.5mm}
\includegraphics[width = 0.163\linewidth]{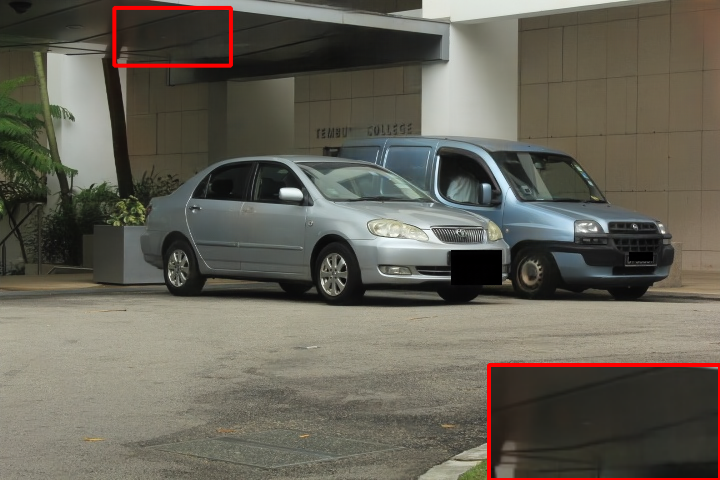} &\hspace{-4.5mm}
\includegraphics[width = 0.163\linewidth]{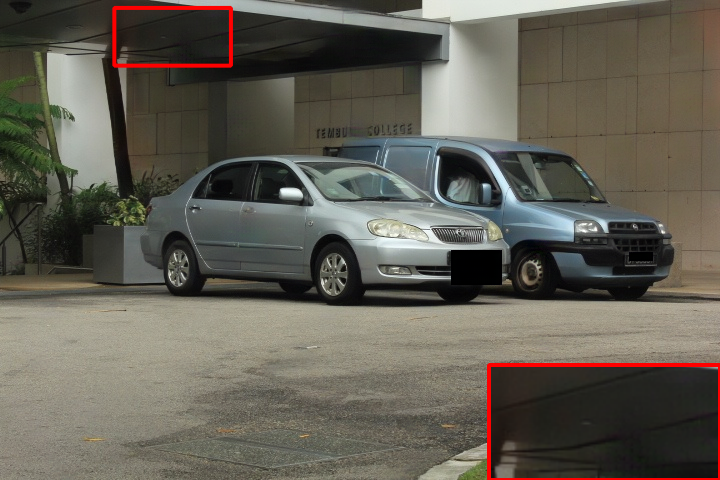} &\hspace{-4.5mm}
\includegraphics[width = 0.163\linewidth]{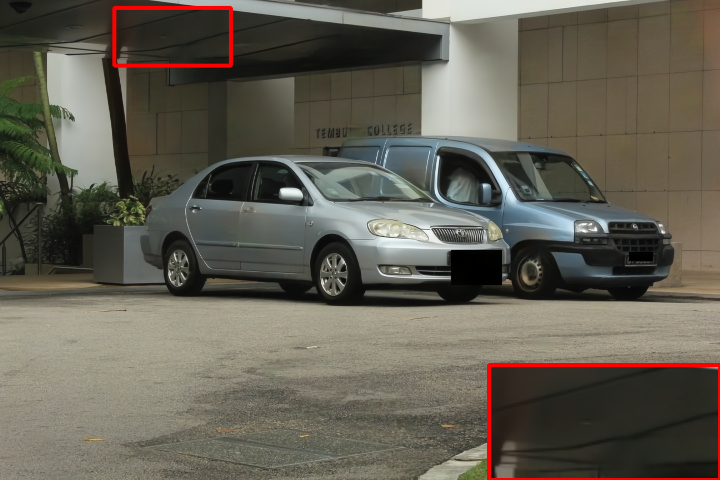} &\hspace{-4.5mm}
\includegraphics[width = 0.163\linewidth]{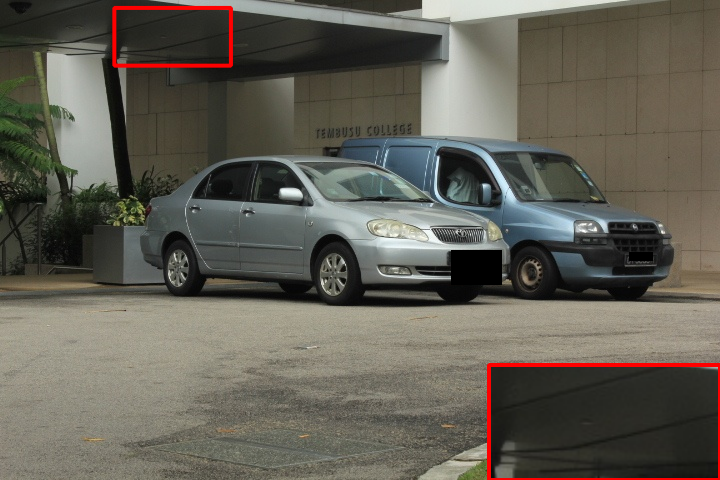} 
\\
\hspace{-1.5mm}\includegraphics[width = 0.163\linewidth]{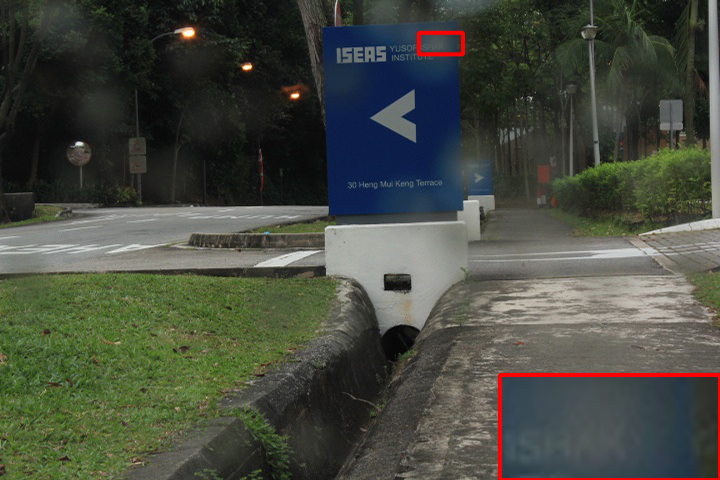} &\hspace{-4.5mm}
\includegraphics[width = 0.163\linewidth]{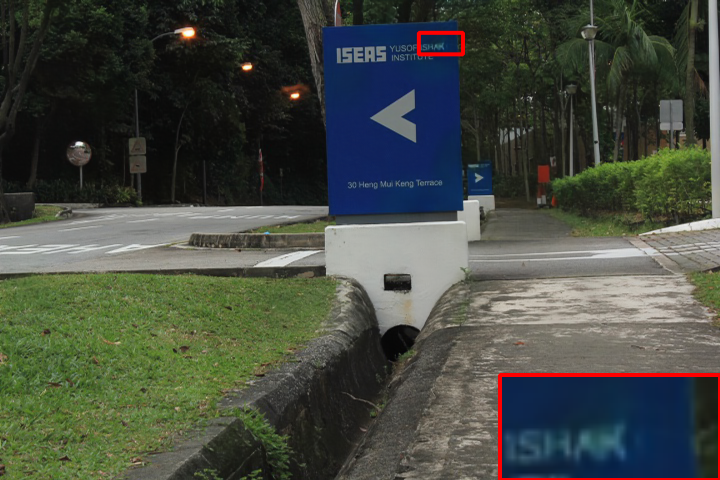}  &\hspace{-4.5mm}
\includegraphics[width = 0.163\linewidth]{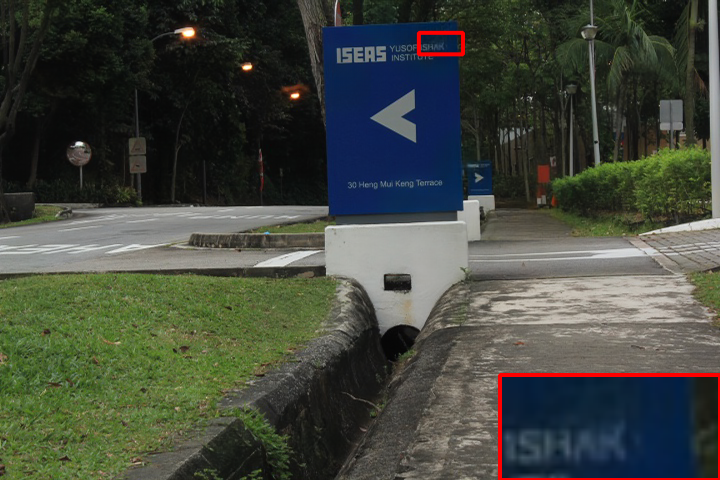} &\hspace{-4.5mm}
\includegraphics[width = 0.163\linewidth]{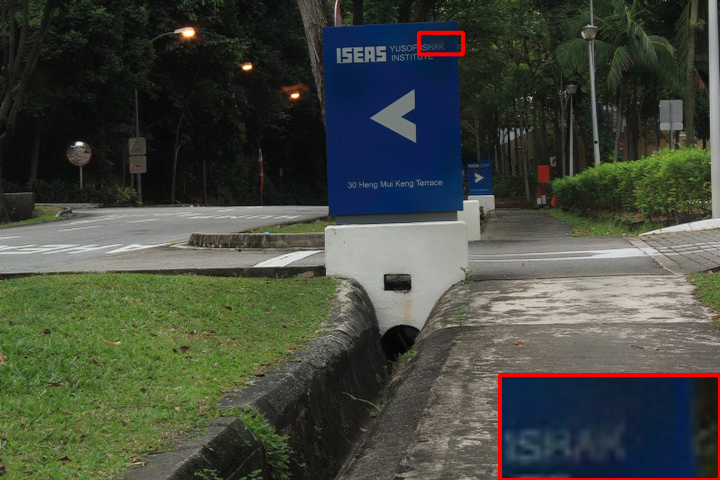} &\hspace{-4.5mm}
\includegraphics[width = 0.163\linewidth]{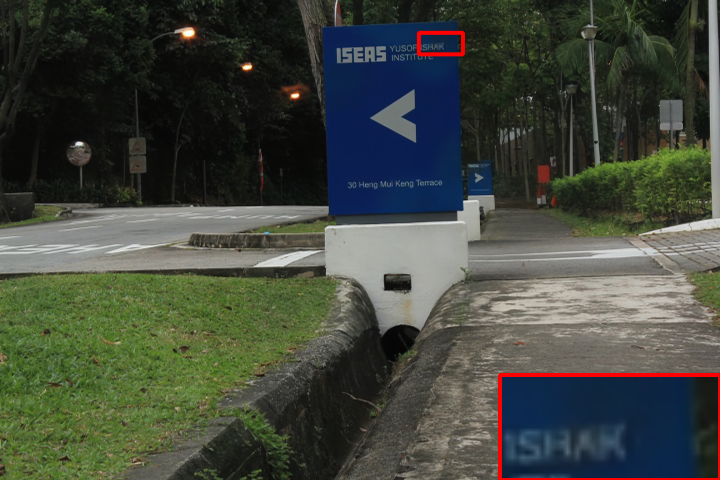} &\hspace{-4.5mm}
\includegraphics[width = 0.163\linewidth]{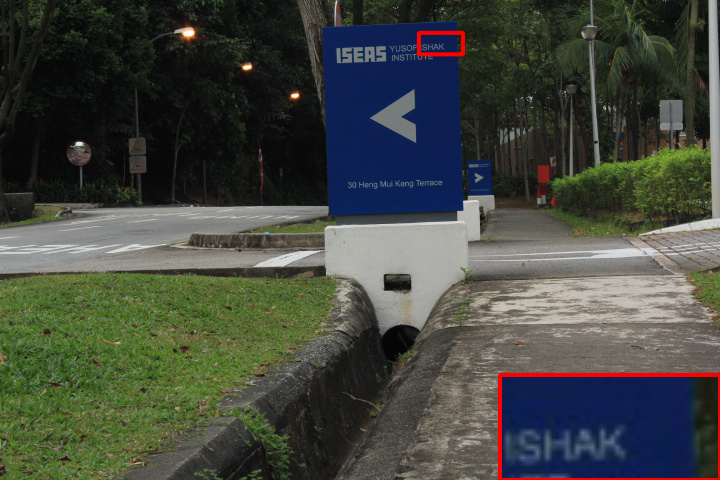} 
\\
\hspace{-1.5mm}(a) Input   &\hspace{-4.5mm} (b) Histoformer &\hspace{-4.5mm} {(c) T3-DiffWeather}&\hspace{-4.5mm}  {(d) GridFormer}&\hspace{-4.5mm} (e)   Ours&\hspace{-4.5mm} (f) GT
\end{tabular}
\end{center}
\vspace{-2mm}
\caption{Visual comparison with SOTA methods on the RainDrop test set~\cite{RainDropAttn}.
Red boxes highlight areas with visible differences. DCMPC-Net achieves more thorough raindrop removal and recovers finer image details than other competing methods.}
\label{fig:p7}
\vspace{-2mm}
\end{figure*}

\begin{figure*}[!t]\footnotesize
\centering
\setlength{\tabcolsep}{1pt} 

\begin{tabular}{cccccc}
    \includegraphics[width=0.163\linewidth]{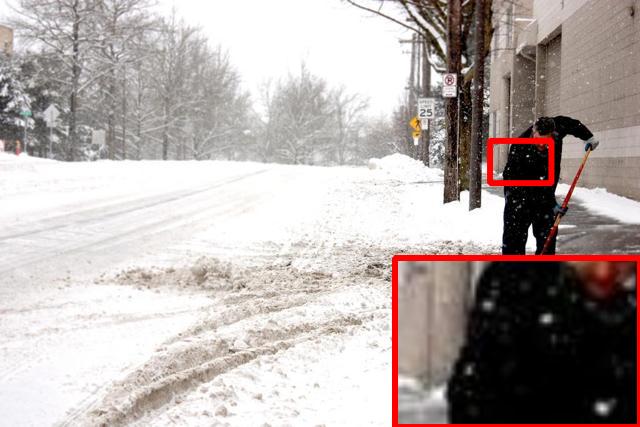} &
    \includegraphics[width=0.163\linewidth]{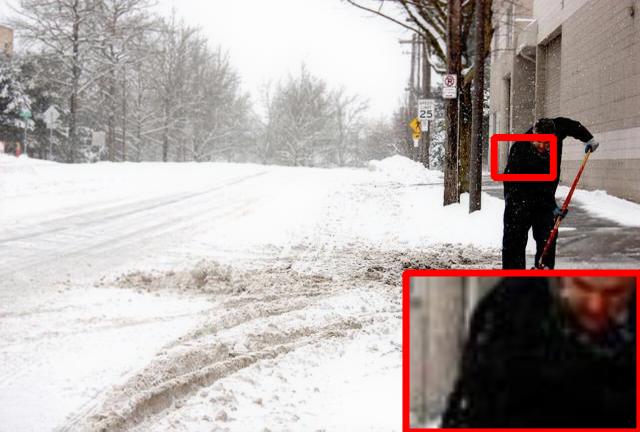} &
    \includegraphics[width=0.163\linewidth]{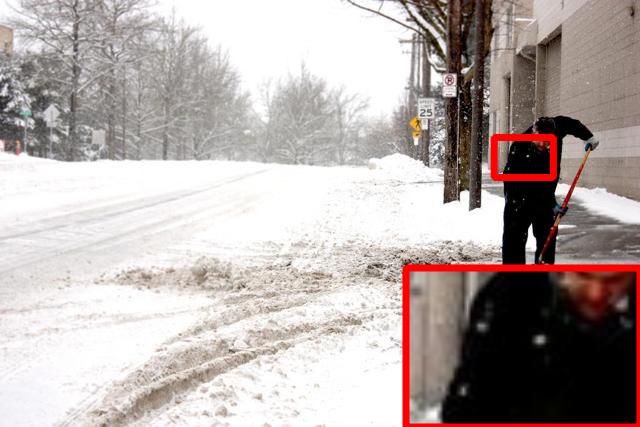} &
    \includegraphics[width=0.163\linewidth]{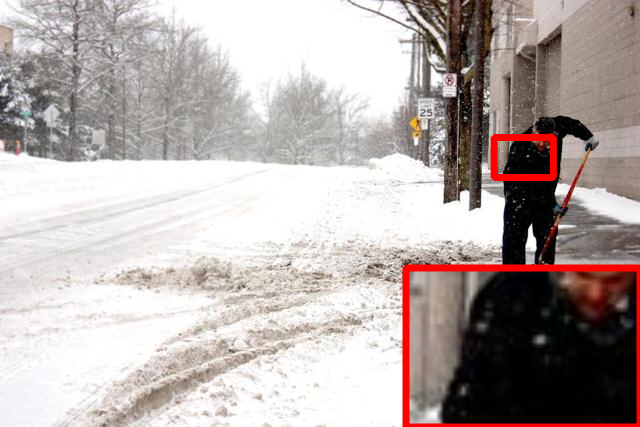} &
    \includegraphics[width=0.163\linewidth]{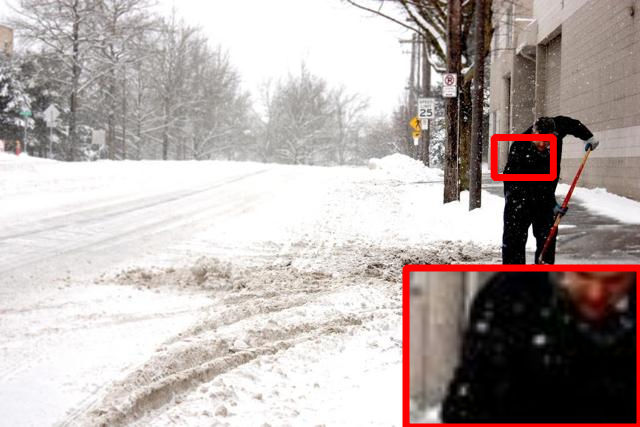} &
    \includegraphics[width=0.163\linewidth]{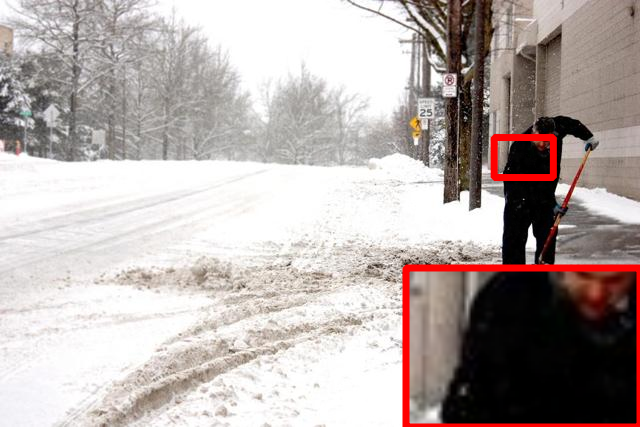} \\

    \includegraphics[width=0.163\linewidth]{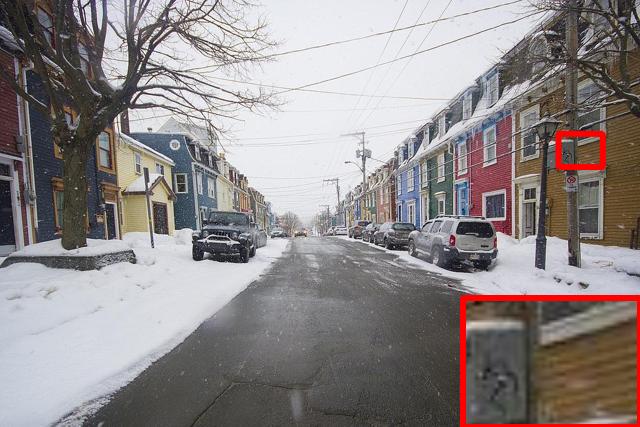} &
    \includegraphics[width=0.163\linewidth]{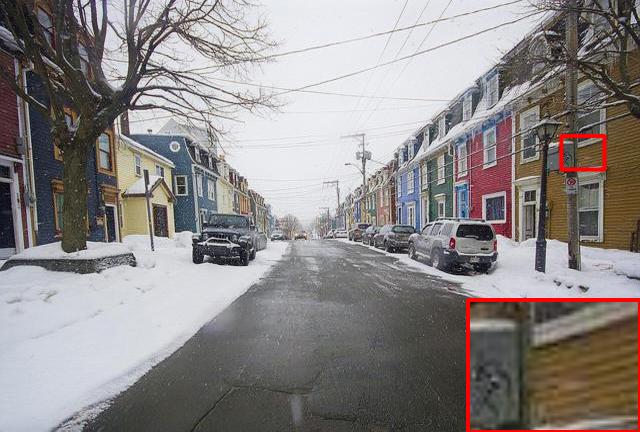} &
    \includegraphics[width=0.163\linewidth]{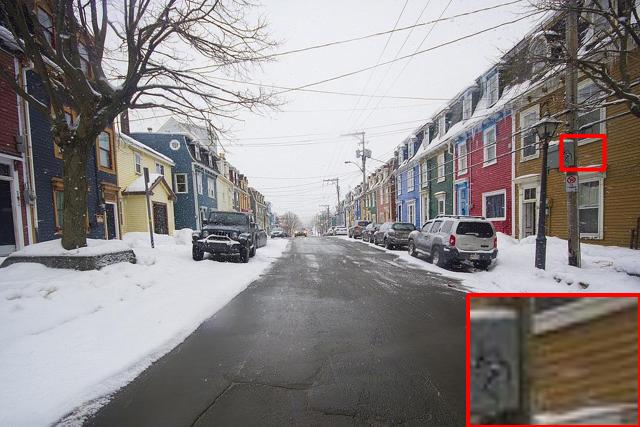} &
    \includegraphics[width=0.163\linewidth]{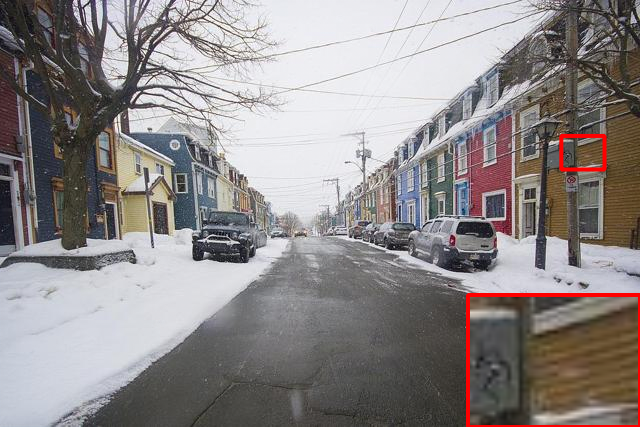} &
    \includegraphics[width=0.163\linewidth]{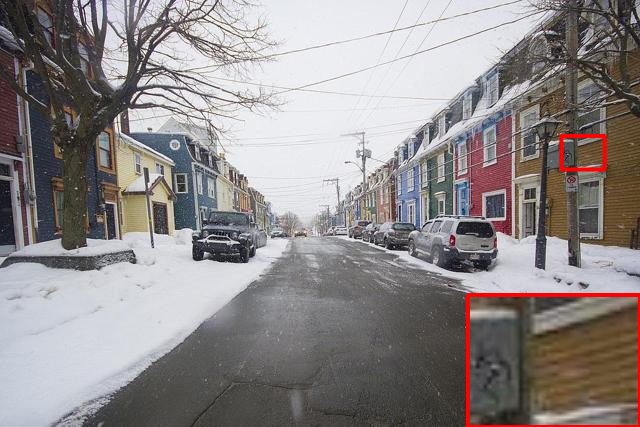} &
    \includegraphics[width=0.163\linewidth]{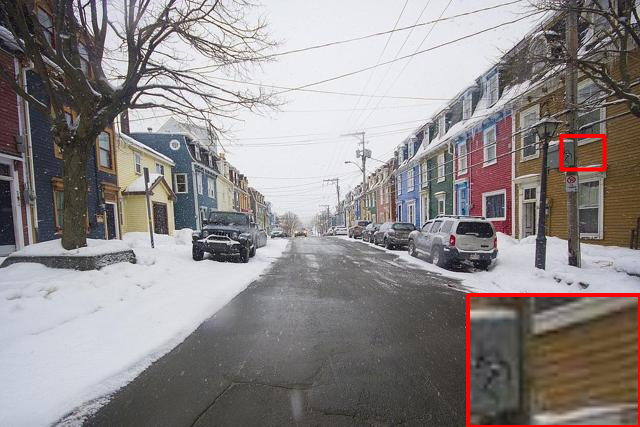} \\

    \includegraphics[width=0.163\linewidth]{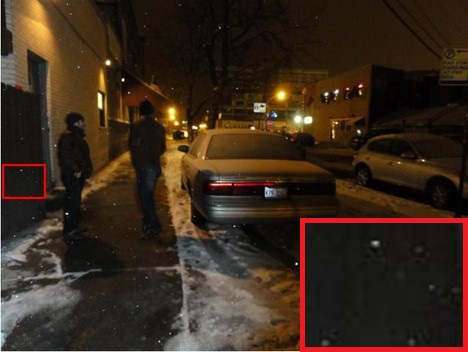} &
    \includegraphics[width=0.163\linewidth]{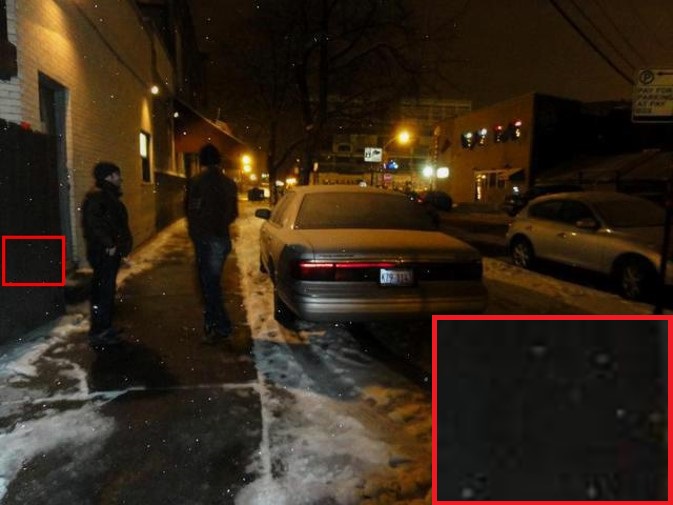} &
    \includegraphics[width=0.163\linewidth]{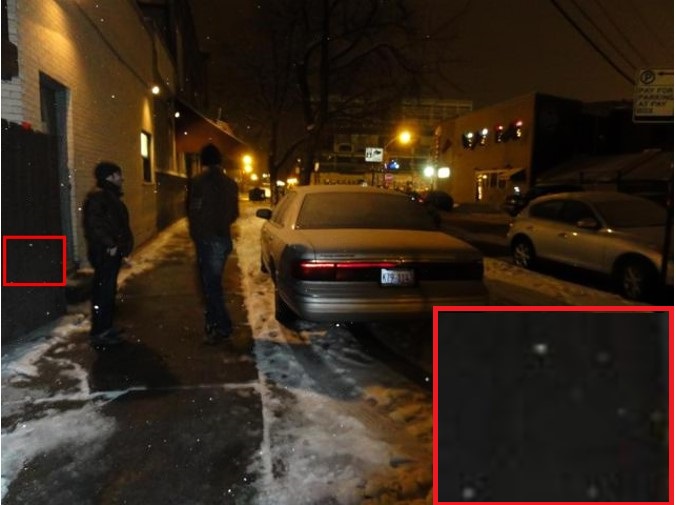} &
    \includegraphics[width=0.163\linewidth]{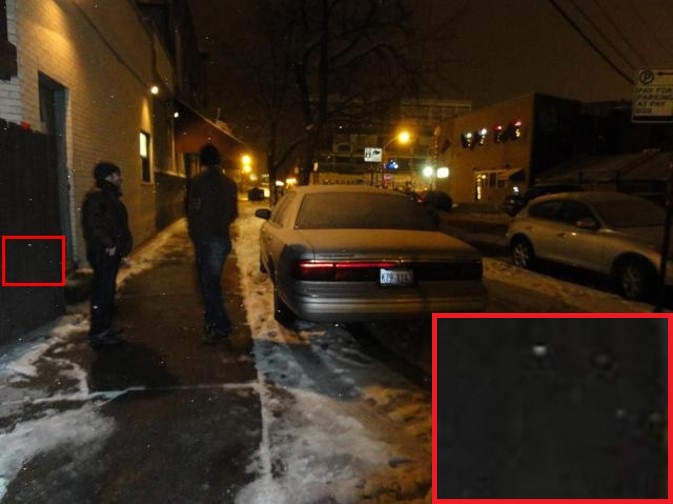} &
    \includegraphics[width=0.163\linewidth]{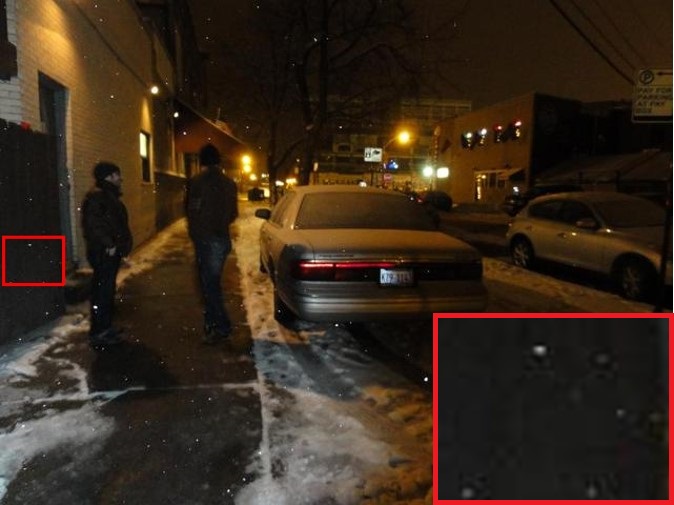} &
    \includegraphics[width=0.163\linewidth]{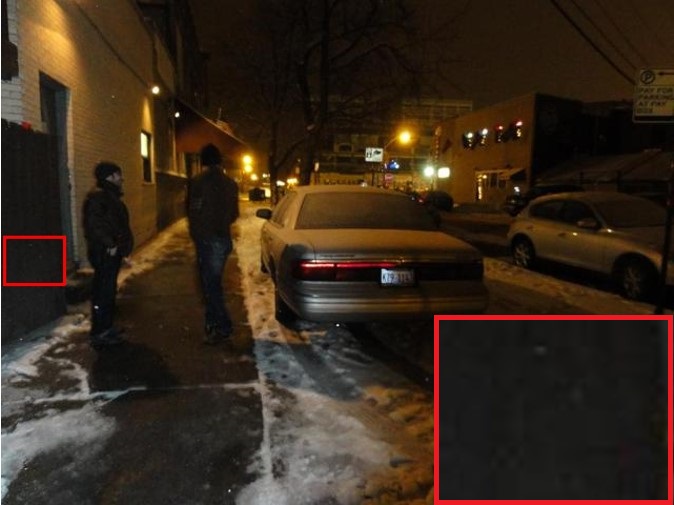} \\


    \includegraphics[width=0.163\linewidth]{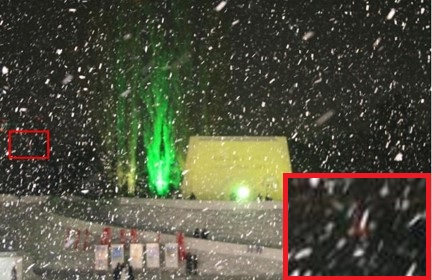} &
    \includegraphics[width=0.163\linewidth]{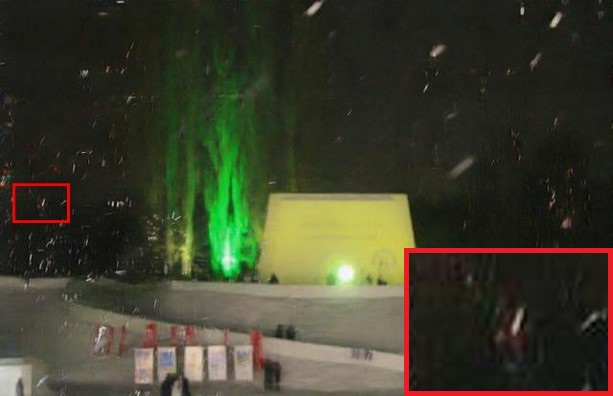} &
    \includegraphics[width=0.163\linewidth]{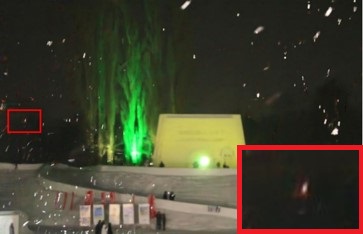} &
    \includegraphics[width=0.163\linewidth]{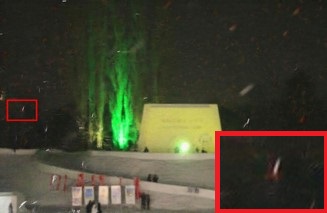} &
    \includegraphics[width=0.163\linewidth]{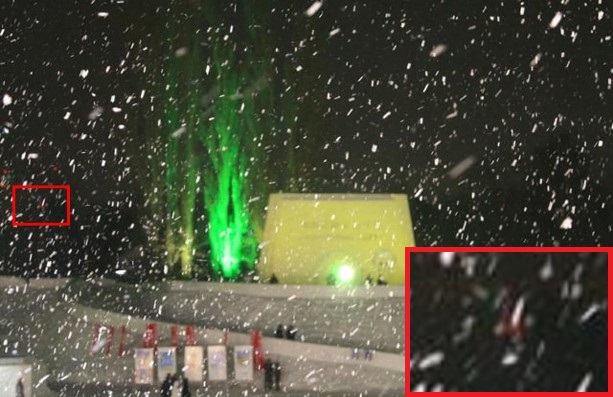} &
    \includegraphics[width=0.163\linewidth]{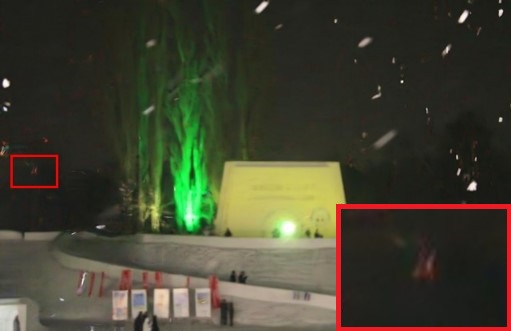} \\

    (a) Input & 
    (b) TransWeather & 
    (c) Histoformer & 
    {(d) T3-DiffWeather} & 
    {(e) GridFormer} & 
    (f) Ours
\end{tabular}
\caption{Visual comparison with SOTA methods on the Real-Snow test set~\cite{DesnowNet}.
Red boxes highlight regions with notable differences across methods. DCMPC-Net more effectively removes snow artifacts and produces cleaner backgrounds in real scenes than other competing methods.}
\label{fig:p8}
\end{figure*}

%


\subsubsection{Qualitative Evaluation}
We further provide qualitative comparisons on desnowing, rain\&fog removal, and RainDrop removal.
As shown in Fig.~\ref{fig:p5}, DCMPC-Net restores clearer content with better fine-structure preservation, while competing methods often suffer from blur or artifacts in texture-rich regions of Snow100K-real~\cite{DesnowNet}.
Fig.~\ref{fig:p6} presents qualitative comparisons on OutdoorRain~\cite{Test1}. 
Competing methods exhibit texture inconsistency and structural misalignment in background and edge regions, while DCMPC-Net produces cleaner and structurally consistent restorations.
Similarly, Fig.~\ref{fig:p7} shows that our method removes raindrops more effectively and reconstructs finer textures.
Furthermore, Snow100K~\cite{DesnowNet} includes real-world snowy scenes, offering a more challenging and realistic evaluation for desnowing.
As shown in Fig.~\ref{fig:p8}, DCMPC-Net produces more realistic restorations with fewer residual artifacts than competing methods, demonstrating improved visual fidelity across diverse adverse weather conditions.
%


%

{\subsubsection{User Study}
We conduct a user study to evaluate perceptual quality beyond objective metrics.
Restoration results on 30 real-world adverse weather images are presented in a randomized and anonymous manner.
\textcolor{blue}{Twelve} participants with computer vision backgrounds are asked to select the perceptually best result for each image.
Table~\ref{tab:user_study} summarizes the voting results, where DCMPC-Net achieves the highest preference percentage.}

\subsection{Ablation Studies}
%
We conduct ablation experiments to assess the contributions of individual components. Evaluation is performed on Snow100K-S~\cite{DesnowNet}, Snow100K-L~\cite{DesnowNet}, OutdoorRain~\cite{Test1}, and RainDrop~\cite{RainDropAttn} test sets using PSNR and SSIM metrics.
\enlargethispage{-0.6\baselineskip}
\subsubsection{Effect of Core Components}
We conduct ablation studies to assess the contributions of key modules in DCMPC-Net. 
Table~\ref{tab:ablation_3} reports quantitative results on four benchmark datasets. 
Removing CMPG, which generates degradation-aware cross-modal prompts, leads to clear drops in PSNR and SSIM.
Similarly, removing PGAAM or DFCM also degrades performance, confirming their importance to restoration quality. 
For fair comparison, all restoration-stage complexity metrics are measured under the same cached-prompt protocol and input resolution.
Fig.~\ref{fig:p9} presents qualitative comparisons of different model configurations.
Integrating CMPG, PGAAM, and DFCM improves detail preservation and structural consistency, producing cleaner and more visually faithful restorations.
These results demonstrate the complementary roles of the proposed components in adverse weather removal.


{\begin{table}[!t]
\centering
\renewcommand{\arraystretch}{1.1}
\caption{{User study preference results on real-world adverse weather images.
DCMPC-Net receives the most votes among all compared methods.}}
\label{tab:user_study}

\begin{tabular}{lcc}
\hline
Method & Frequency of being selected & Preference (\%) \\
\hline
TransWeather~\cite{TransWeather} & 12  & 3.33 \\
Histoformer~\cite{Histoformer} & 64 & 17.78 \\
T3-DiffWeather~\cite{T3-DiffWeather} & 48 & 13.33 \\
GridFormer~\cite{GridFormer}  & 55 & 15.28 \\
\textbf{DCMPC-Net (Ours)} & \textbf{181} & \textbf{50.28} \\
\hline
\end{tabular}
\vspace{-2mm}
\end{table}

\begin{table*}[t]
\centering
\setlength{\tabcolsep}{10pt} 
\renewcommand{\arraystretch}{1.1}

\caption{
Ablation study of core components in DCMPC-Net. PSNR and SSIM are reported on four benchmark datasets. 
Model complexity denotes the restoration-stage cost under the cached-prompt setting and is measured on $256 \times 256$ inputs.
}
\label{tab:ablation_3}

{%
\resizebox{\textwidth}{!}{%
\begin{tabular}{lccccccccccc}
\toprule
\multirow{3}{*}{Configuration}
& \multicolumn{8}{c}{Restoration Performance}
& \multicolumn{3}{c}{Model Complexity} \\
\cmidrule(lr){2-9}\cmidrule(lr){10-12}
& \multicolumn{2}{c}{Snow-S}
& \multicolumn{2}{c}{Snow-L}
& \multicolumn{2}{c}{OutdoorRain}
& \multicolumn{2}{c}{RainDrop}
& Params.
& FLOPs
& Runtime \\
\cmidrule(lr){2-3}\cmidrule(lr){4-5}\cmidrule(lr){6-7}\cmidrule(lr){8-9}
& PSNR & SSIM
& PSNR & SSIM
& PSNR & SSIM
& PSNR & SSIM
& (M)
& (G)
& (s) \\
\midrule
w/o CMPG
& 38.00 & 0.9683 & 32.15 & 0.9283 & 32.09 & 0.9456 & 33.06 & 0.9467
& 30.46 & 172.74 & 0.05 \\
w/o PGAAM
& 37.99 & 0.9682 & 32.12 & 0.9282 & 32.04 & 0.9452 & 32.92 & 0.9463
& 34.37 & 208.35 & 0.07 \\
w/o DFCM
& 38.12 & 0.9687 & 32.25 & 0.9293 & 32.34 & 0.9465 & 32.94 & 0.9470
& 36.99 & 251.72 & 0.08 \\
DCMPC-Net
& \textbf{38.21} & \textbf{0.9692}
& \textbf{32.35} & \textbf{0.9302}
& \textbf{32.49} & \textbf{0.9477}
& \textbf{33.08} & \textbf{0.9474}
& 37.60 & 261.00 & 0.08 \\
\bottomrule
\end{tabular}%
}%
}
\end{table*}

\begin{table}[h]
\centering
\caption{
{Quantitative comparison with baseline and variants on four benchmark datasets.}
}
\label{tab:ablation_5}

{%
    \resizebox{\columnwidth}{!}{%
    \begin{tabular}{c c c c c c}
        \toprule
        Configuration & Metric & Snow-S & Snow-L & OutdoorRain & RainDrop \\ 
        \midrule
        
        \multirow{2}{*}{Histoformer~\cite{Histoformer}} 
            & PSNR & 37.41  & 32.16  & 32.08  & 33.06  \\  
            & SSIM & 0.9658 & 0.9261 & 0.9389 & 0.9441 \\ 
        \midrule
        
        \multirow{2}{*}{\shortstack{CMPG $\rightarrow$\\Text Features}} 
            & PSNR & 38.00  & 32.15  & 31.95  & 32.96  \\ 
            & SSIM & 0.9682 & 0.9284 & 0.9452 & 0.9464 \\ 
        \midrule
        
        \multirow{2}{*}{DCMPC-Net}       
            & PSNR & \textbf{38.21} & \textbf{32.35} & \textbf{32.49} & \textbf{33.08} \\ 
            & SSIM & \textbf{0.9692} & \textbf{0.9302} & \textbf{0.9477} & \textbf{0.9474} \\ 
        \bottomrule                 
    \end{tabular}%
    }%
}

\end{table}

\begin{table*}[t]
\centering
\setlength{\tabcolsep}{6pt} 
\renewcommand{\arraystretch}{1.1}

\caption{
Ablation study on CMPG components.
We evaluate alternative designs by (1) replacing semantic text features with learnable queries,
and (2) substituting CMPG with the prompt block from PromptIR. 
Model complexity is measured on $256 \times 256$ inputs for the restoration-stage forward pass under the cached-prompt setting.
}
\label{tab:ablation_6}

{%
\resizebox{\textwidth}{!}{%
\begin{tabular}{lcccccccccccc@{\hskip 12pt}ccc}
\toprule
\multirow{3}{*}{Configuration}
& \multicolumn{12}{c}{Restoration Performance}
& \multicolumn{3}{c}{Model Complexity} \\
\cmidrule(lr){2-13}\cmidrule(lr){14-16}
& \multicolumn{3}{c}{Snow-S}
& \multicolumn{3}{c}{Snow-L}
& \multicolumn{3}{c}{OutdoorRain}
& \multicolumn{3}{c}{RainDrop}
& Params.
& FLOPs
& Runtime \\
\cmidrule(lr){2-4}\cmidrule(lr){5-7}\cmidrule(lr){8-10}\cmidrule(lr){11-13}
& PSNR~$\uparrow$ & SSIM~$\uparrow$ & LPIPS~$\downarrow$
& PSNR~$\uparrow$ & SSIM~$\uparrow$ & LPIPS~$\downarrow$
& PSNR~$\uparrow$ & SSIM~$\uparrow$ & LPIPS~$\downarrow$
& PSNR~$\uparrow$ & SSIM~$\uparrow$ & LPIPS~$\downarrow$
& (M)
& (G)
& (s) \\
\midrule
w/o CMPG
& 38.00 & 0.9683 & 0.0004
& 32.15 & 0.9283 & 0.0012
& 32.09 & 0.9456 & 0.0009
& 33.06 & 0.9467 & 0.0009
& 30.46 & 172.74 & 0.05 \\

CMPG $\rightarrow$ Learnable Queries
& 38.00 & 0.9683 & 0.0004
& 32.15 & 0.9283 & 0.0011
& 31.98 & 0.9454 & 0.0010
& 33.05 & 0.9463 & 0.0008
& 37.60 & 261.00 & 0.08 \\

CMPG $\rightarrow$ Prompt Block
& 38.00 & 0.9682 & 0.0004
& 32.14 & 0.9283 & 0.0011
& 32.05 & 0.9455 & 0.0009
& 33.00 & 0.9464 & 0.0008
& 37.77 & 263.99 & 0.08 \\

DCMPC-Net
& \textbf{38.21} & \textbf{0.9692} & \textbf{0.0004}
& \textbf{32.35} & \textbf{0.9302} & \textbf{0.0010}
& \textbf{32.49} & \textbf{0.9477} & \textbf{0.0008}
& \textbf{33.08} & \textbf{0.9474} & \textbf{0.0008}
& 37.60 & 261.00 & 0.08 \\
\bottomrule
\end{tabular}%
}%
}
\end{table*}

\begin{table}[t] 
\centering
\setlength{\tabcolsep}{4.2pt}  
\renewcommand{\arraystretch}{1}

\caption{
Ablation study evaluating the effect of different pre-trained vision-language models used in the CMPG across four benchmark datasets. Results show that LLaMA-based degradation-aware prompts outperform BLIP, CLIP, and no-prompt baselines, achieving the highest PSNR and SSIM. The best results are highlighted in bold.}
\label{tab:t3}  
\begin{tabular}{l c c c c c}
\toprule
Configuration & Metric & Snow-S & Snow-L & OutdoorRain & RainDrop \\ \midrule
\multirow{2}{*}{Histoformer~\cite{Histoformer}}       & PSNR  & 37.41    & 32.16    & 32.08    & 33.06    \\  
                            & SSIM  & 0.9689   & 0.9261   & 0.9389   & 0.9441    \\ \midrule 
\addlinespace[2pt]
\multirow{2}{*}{BLIP \& CLIP} & PSNR  & 37.99    & 32.15    & 32.03    & 32.81     \\ 
                            & SSIM  & 0.9683   & 0.9283   & 0.9454   & 0.9463     \\ \midrule 
\addlinespace[2pt]
\multirow{2}{*}{DCMPC-Net}     & PSNR  & \textbf{38.21}    & \textbf{32.35}    & \textbf{32.49}    & \textbf{33.08}     \\ 
                            & SSIM  & \textbf{0.9692}   & \textbf{0.9302}   & \textbf{0.9477}   & \textbf{0.9474}     \\ 
\bottomrule                
\end{tabular}
\end{table}

\begin{table}[t]
\centering
\setlength{\tabcolsep}{4.5pt}
\renewcommand{\arraystretch}{1}
\caption{Effect of the DFCM on model performance. The best results are highlighted in bold.}
\begin{tabular}{l c c c c c}
\toprule
Configuration & Metric & Snow-S & Snow-L & OutdoorRain & RainDrop\\ \midrule 
\multirow{2}{*}{GDFN~\cite{Restormer}}   & PSNR  & 38.14    & 32.24   & 32.40    & 33.08    \\  
                                         & SSIM  & 0.9689  & 0.9294    & 0.9470  & 0.9470     \\ \midrule 
\addlinespace[2pt]
\multirow{2}{*}{DGFF~\cite{Histoformer}} & PSNR  & 38.12     & 32.21   & 32.41    & 33.01    \\ 
                                         & SSIM  & 0.9687   & 0.9298   & 0.9469   & 0.9472     \\ \midrule 
\addlinespace[2pt]                                   
\multirow{2}{*}{DCMPC-Net}              & PSNR  & \textbf{38.21}   & \textbf{32.35}    & \textbf{32.49}  & \textbf{33.08}     \\ 
                                        & SSIM  & \textbf{0.9692}  & \textbf{0.9302}   & \textbf{0.9477} & \textbf{0.9474}     \\       
\bottomrule     
\end{tabular}
\label{tab:t4}
\end{table}

\begin{table}[t]
\centering
\setlength{\tabcolsep}{26.75pt}
\renewcommand{\arraystretch}{1}
\caption{Ablation experiments of DFCM. Note that we input the restoration branch (RB) and the compensation branch (CB) separately. Our full model, which utilizes both the restoration branch and the compensation branch, achieves better performance. The best results are highlighted in bold.}
\begin{tabular}{l c c}
\toprule
Configuration            & PSNR & SSIM  \\ \midrule

\multirow{1}{*}{w/o RB}         & 32.07    & 0.9452    \\  
                           
\multirow{1}{*}{w/o CB}         & 32.00    & 0.9453    \\     

\multirow{1}{*}{DCMPC-Net}       & \textbf{32.49}  & \textbf{0.9477}   \\      
\bottomrule     
\end{tabular}
\label{tab:t5}
\end{table}

                          
                           


\begin{table}[t]
\centering
\setlength{\tabcolsep}{5pt}
\renewcommand{\arraystretch}{1}
\caption{Effect of localization of PGAAM and DFCM. 
PSNR, SSIM evaluation of various localizations of Prompt-Guided Attention Alignment Module (PGAAM) and Dual Feature Compensation Module(DFCM). Enc. denotes the location of the encoder, and Dec. denotes the location of the decoder. }
\label{tab:t6}
\begin{tabular}{l c c c c c}
\toprule
Configuration & Metric          & Snow-S & Snow-L & OutdoorRain & RainDrop   \\ \midrule 
\multirow{2}{*}{Enc.}   & PSNR  & 38.17    & 32.31   & 32.42    & 33.11      \\  
                        & SSIM  & 0.9689  & 0.9297    & 0.9470  & 0.9471     \\ \midrule 
\addlinespace[2pt]
\multirow{2}{*}{Dec.}  & PSNR  & \textbf{38.21}   & \textbf{32.35}    & \textbf{32.49}  & \textbf{33.08}     \\ 
                      & SSIM  & \textbf{0.9692}  & \textbf{0.9302}   & \textbf{0.9477} & \textbf{0.9474}     \\  \midrule  
\addlinespace[2pt]                                   
\multirow{2}{*}{Enc.+Dec.}    & PSNR  & 37.01    & 31.41   & 30.99    & 32.57      \\  
                              & SSIM  & 0.9643   & 0.9207    & 0.9325  & 0.9412     \\      
\bottomrule     
\end{tabular}
\end{table}
\begin{figure*}[!t]\footnotesize
\begin{center}
\begin{tabular}{ccccccc}  
\hspace{-1.5mm}\includegraphics[width=0.1385\linewidth]{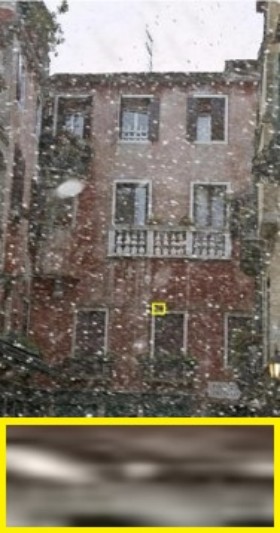} &\hspace{-4.5mm}
\includegraphics[width=0.1385\linewidth]{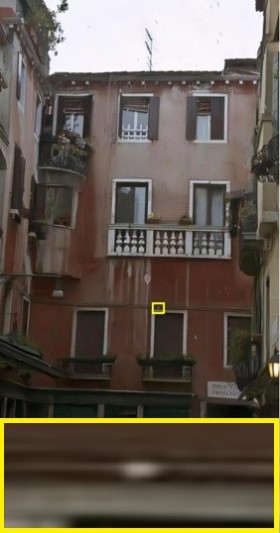}  &\hspace{-4.5mm}
\includegraphics[width=0.1385\linewidth]{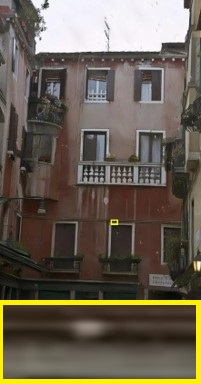} &\hspace{-4.5mm}
\includegraphics[width=0.1385\linewidth]{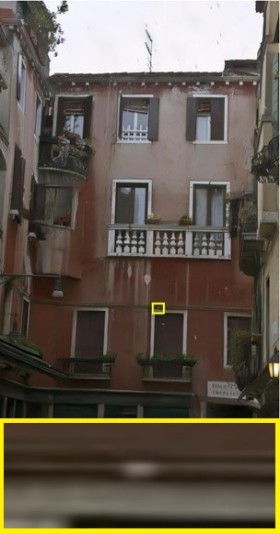} &\hspace{-4.5mm}
\includegraphics[width=0.1385\linewidth]{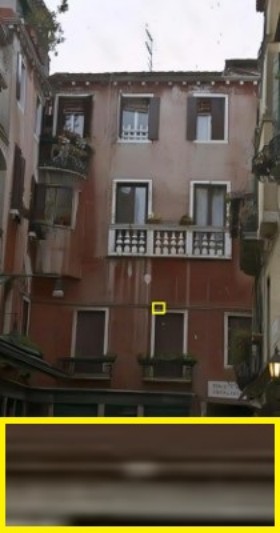} &\hspace{-4.5mm}  
\includegraphics[width=0.1385\linewidth]{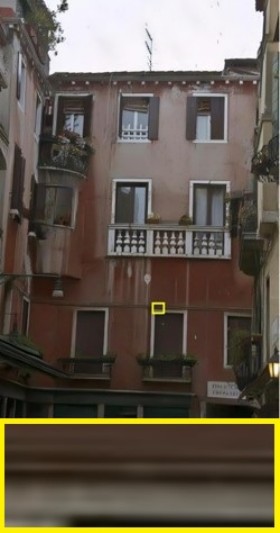} &\hspace{-4.5mm}
\includegraphics[width=0.1385\linewidth]{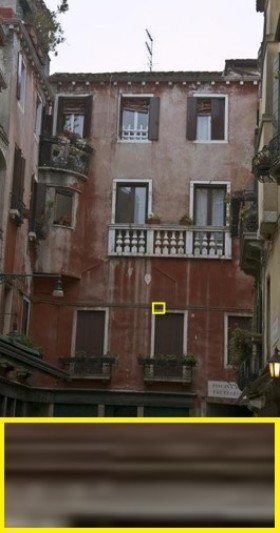} 
\\
\hspace{-1.5mm}\includegraphics[width=0.1385\linewidth]{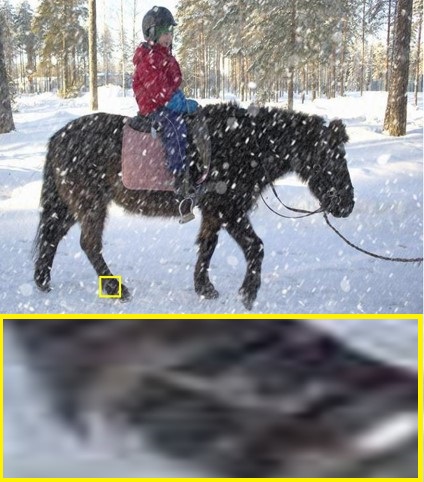} &\hspace{-4.5mm}
\includegraphics[width=0.1385\linewidth]{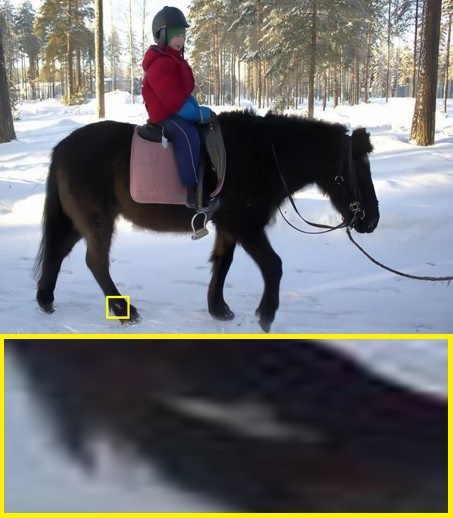} &\hspace{-4.5mm}
\includegraphics[width=0.1385\linewidth]{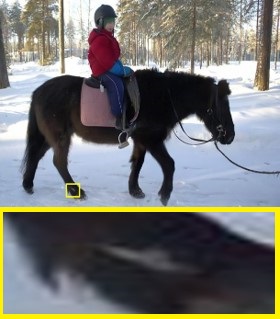} &\hspace{-4.5mm}
\includegraphics[width=0.1385\linewidth]{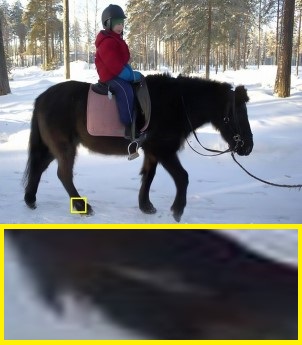} &\hspace{-4.5mm}
\includegraphics[width=0.1385\linewidth]{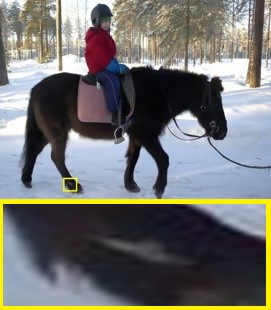} &\hspace{-4.5mm}  
\includegraphics[width=0.1385\linewidth]{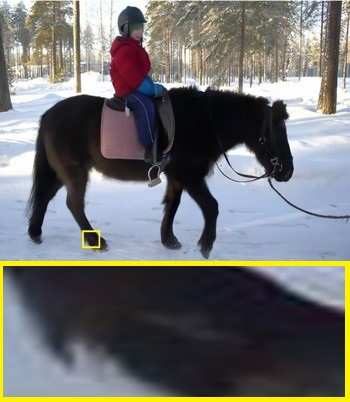}  &\hspace{-4.5mm}
\includegraphics[width=0.1385\linewidth]{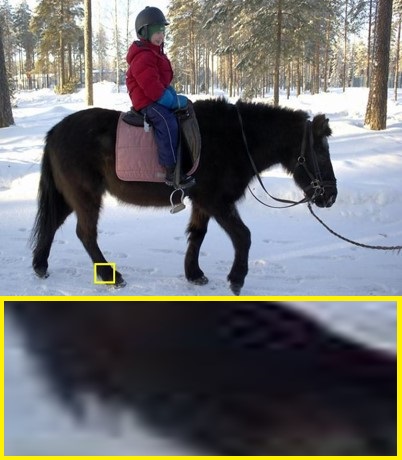} 
\\  

\hspace{-1.5mm}(a) Input &\hspace{-4.5mm} (b) Histoformer &\hspace{-4.5mm} (c) w/o CMPG &\hspace{-4.5mm} (d) w/o PGAAM &\hspace{-4.5mm} (e) w/o DFCM&\hspace{-4.5mm} (f) Ours &\hspace{-4.5mm} (g) GT  
\end{tabular}
\end{center}
\vspace{-2mm}
\caption{Visual comparison of models with different configurations. Incorporating CMPG, PGAAM, and DFCM enables DCMPC-Net to produce results with clearer details and fewer artifacts than configurations without these modules. Best viewed by zooming in.}
\label{fig:p9}
\end{figure*}

\begin{figure}[t]\footnotesize
    \centering
    \begin{minipage}[t]{0.32\linewidth}
        \centering
        \includegraphics[width=\textwidth]{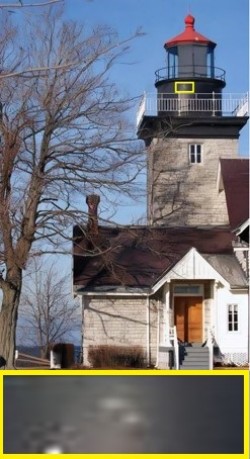}
        \PanelLabel{BLIP\&CLIP}
    \end{minipage}
    \begin{minipage}[t]{0.32\linewidth}
        \centering
        \includegraphics[width=\textwidth]{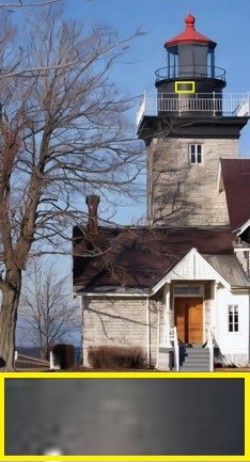}
        \PanelLabel{Ours}
    \end{minipage}
    \begin{minipage}[t]{0.32\linewidth}
        \centering
        \includegraphics[width=\textwidth]{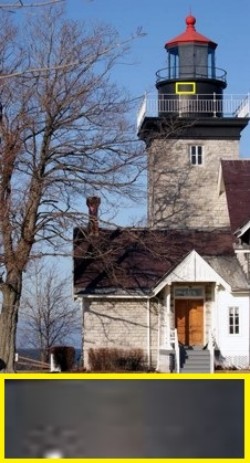}
        \PanelLabel{GT}
    \end{minipage}
    \caption{Visual comparison between different pre-trained vision language models on the Snow100K test set~\cite{DesnowNet}. Leveraging LLaMA to guide the adverse weather removal network results in sharper detail recovery compared to other models.}
    \label{fig:p10}
    \vspace{-3mm}
\end{figure}
\begin{figure}[!t]\footnotesize
\centering
\begin{minipage}[t]{0.32\linewidth}
    \centering
    \includegraphics[width=\textwidth]{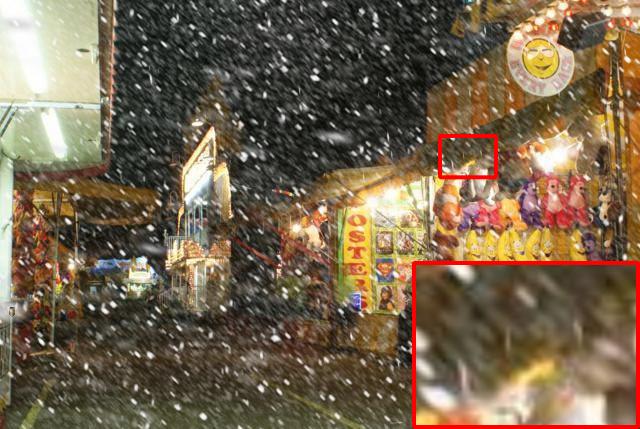}
    \PanelLabel{(a) Input}
\end{minipage}
\begin{minipage}[t]{0.32\linewidth}
    \centering
    \includegraphics[width=\textwidth]{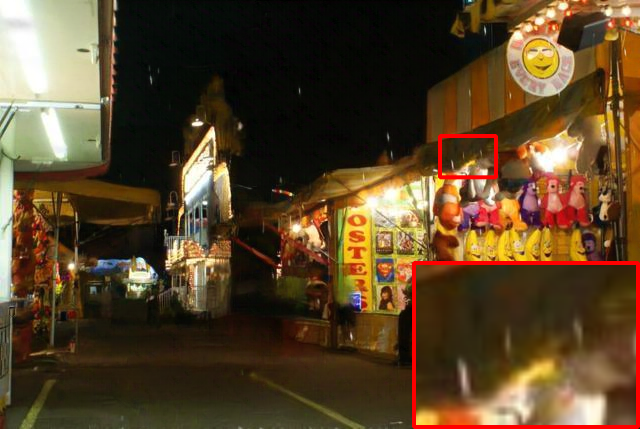}
    \PanelLabel{(b) w/o CMPG}
\end{minipage}
\begin{minipage}[t]{0.32\linewidth}
    \centering
    \includegraphics[width=\textwidth]{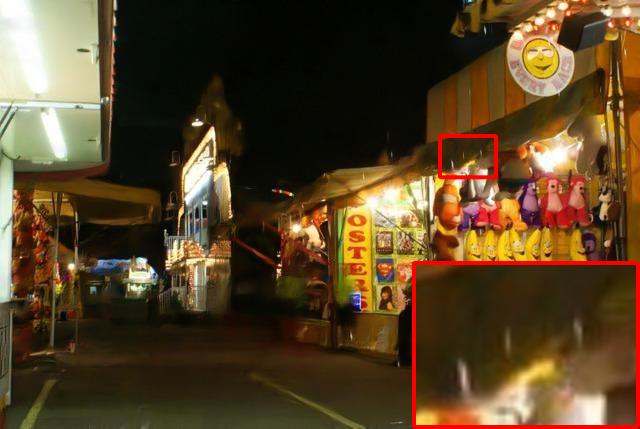}
    \PanelLabel{(c) Learnable Queries}
\end{minipage}
\begin{minipage}[t]{0.32\linewidth}
    \centering
    \includegraphics[width=\textwidth]{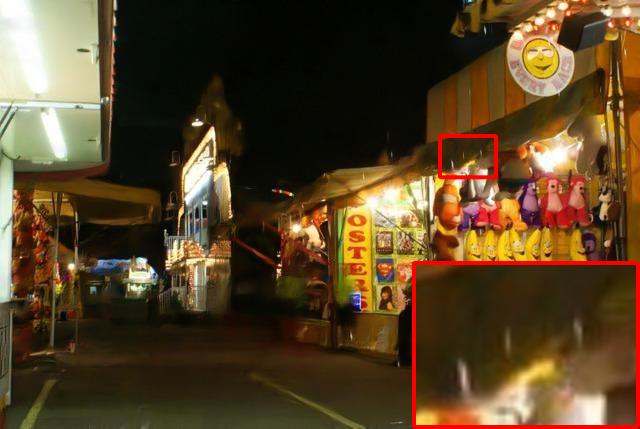}
    \PanelLabel{(d) Prompt Block}
\end{minipage}
\begin{minipage}[t]{0.32\linewidth}
    \centering
    \includegraphics[width=\textwidth]{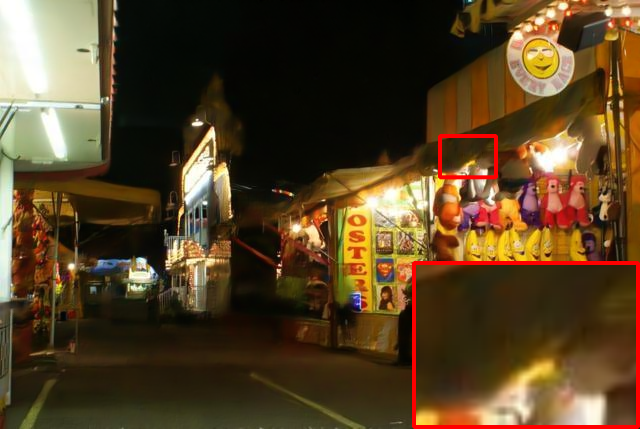}
    \PanelLabel{(e) Ours}
\end{minipage}
\begin{minipage}[t]{0.32\linewidth}
    \centering
    \includegraphics[width=\textwidth]{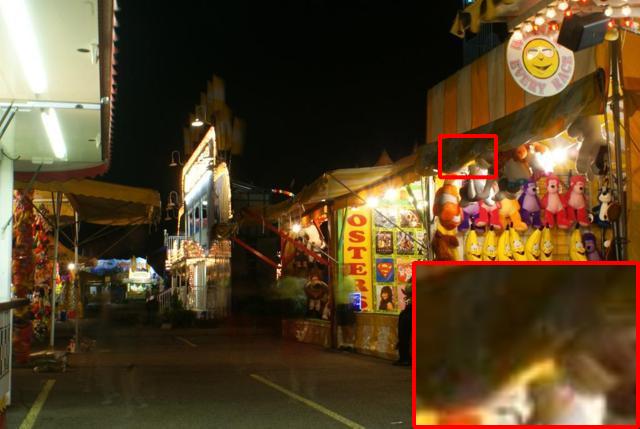}
    \PanelLabel{(f) GT}
\end{minipage}
\caption{{Visual comparison between CMPG and its lightweight prompting variants on the Snow100K test set~\cite{DesnowNet}. Incorporating CMPG in the prompting mechanism results in cleaner outputs with fewer snow artifacts.}}
\label{fig:CMPG-prompt}
\end{figure}

\begin{figure}[!t]\footnotesize
\centering
\begin{minipage}[t]{0.32\linewidth}
    \centering
    \includegraphics[width=\textwidth]{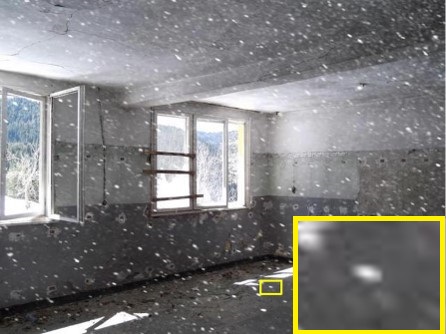}
    \PanelLabel{(a) Input}
\end{minipage}
\begin{minipage}[t]{0.32\linewidth}
    \centering
    \includegraphics[width=\textwidth]{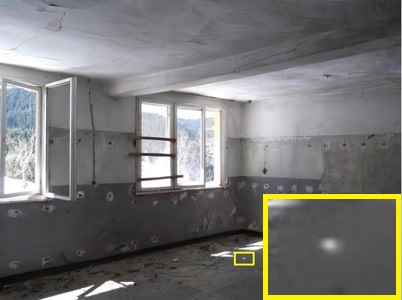}
    \PanelLabel{(b) Histoformer}
\end{minipage}
\begin{minipage}[t]{0.32\linewidth}
    \centering
    \includegraphics[width=\textwidth]{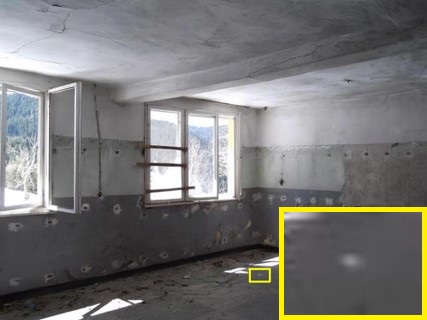}
    \PanelLabel{(c) w/o RB}
\end{minipage}
\begin{minipage}[t]{0.32\linewidth}
    \centering
    \includegraphics[width=\textwidth]{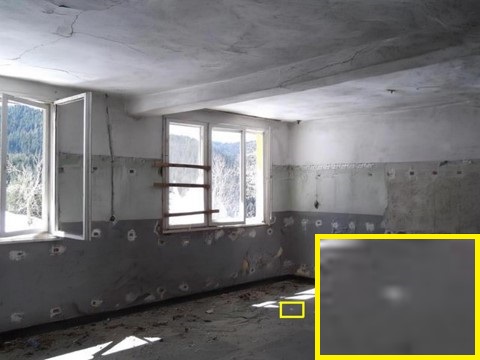}
    \PanelLabel{(d) w/o CB}
\end{minipage}
\begin{minipage}[t]{0.32\linewidth}
    \centering
    \includegraphics[width=\textwidth]{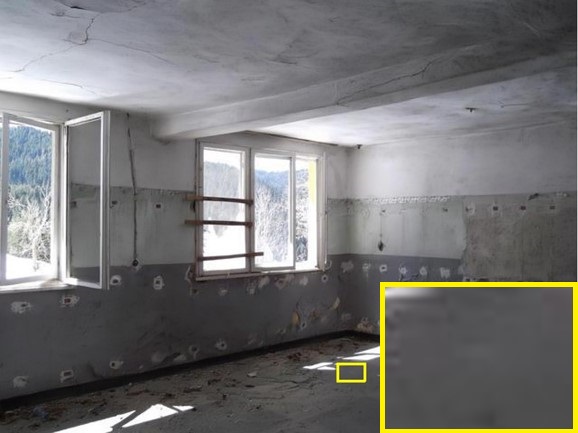}
    \PanelLabel{(e) Ours}
\end{minipage}
\begin{minipage}[t]{0.32\linewidth}
    \centering
    \includegraphics[width=\textwidth]{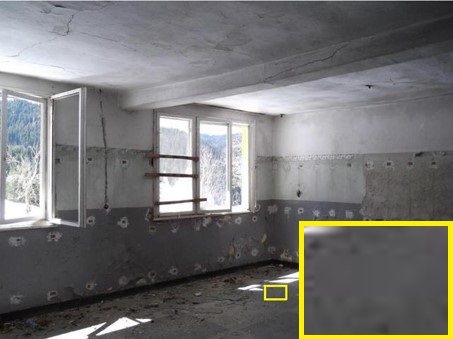}
    \PanelLabel{(f) GT}
\end{minipage}
\caption{Visual comparison between  w/o RB and w/o CB on the Snow100K test set~\cite{DesnowNet}.  Incorporating both RB and CB in the restoration network results in cleaner outputs with fewer residual snow artifacts.}
\label{fig:p11}
\end{figure}

\begin{figure}[t]\footnotesize
    \centering
    \begin{minipage}[t]{0.49\linewidth}
        \centering
        \includegraphics[width=\textwidth]{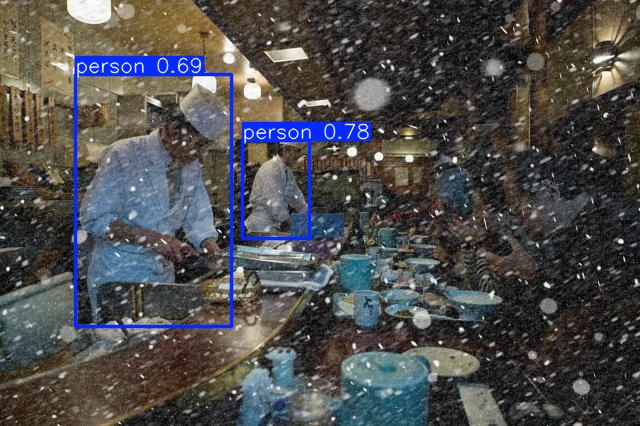}
        \PanelLabel{Input}
    \end{minipage}\hspace{0.3mm}
    \begin{minipage}[t]{0.49\linewidth}
        \centering
        \includegraphics[width=\textwidth]{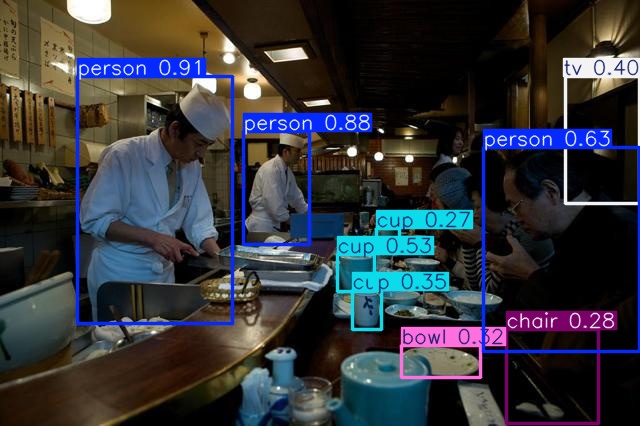}
        \PanelLabel{Deweather by ours}
    \end{minipage}\hspace{0.8mm}
    \vspace{-2mm}
    \caption{Example of object detection results before and after adverse weather removal. The proposed method enables higher detection accuracy on the restored image, demonstrating its effectiveness for downstream tasks under adverse weather conditions.}
    \label{fig:p12}
\end{figure}

\begin{figure}[h]\footnotesize
    \centering
    \begin{minipage}[t]{0.325\linewidth}
        \centering
        \includegraphics[width=\textwidth]{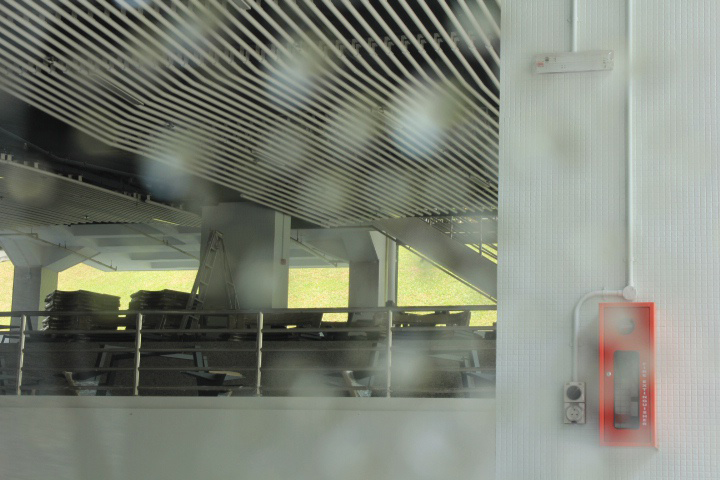}
        \PanelLabel{Input}
    \end{minipage}
    \begin{minipage}[t]{0.325\linewidth}
        \centering
        \includegraphics[width=\textwidth]{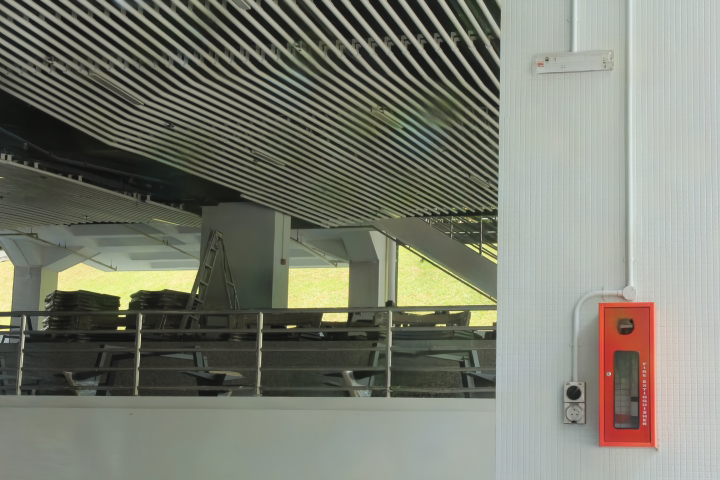}
        \PanelLabel{Ours}
    \end{minipage}
    \begin{minipage}[t]{0.325\linewidth}
        \centering
        \includegraphics[width=\textwidth]{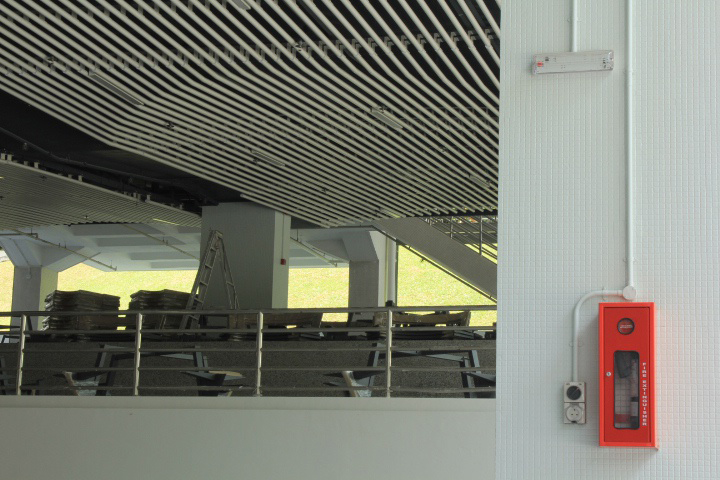}
        \PanelLabel{GT}
    \end{minipage}
    \caption{
    {Failure case of DCMPC-Net under heavy rain conditions.
    When dense and large raindrops overlap with fine-scale structural patterns in the background, the restored result still contains residual artifacts.}
    }
    \label{fig:ablation_3}
\end{figure}

\subsubsection{Effect of CMPG}

%

{
We first evaluate the effectiveness of CMPG by replacing it with standalone text features derived from image descriptions. Quantitative results are reported in Table~\ref{tab:ablation_5}.
As shown, removing cross-modal prompt guidance leads to consistent performance degradation across all benchmark datasets, indicating that unimodal textual representations are insufficient to capture the fine-grained degradation semantics required for robust image restoration.
}

{To further examine whether the performance gains of CMPG can be achieved using more lightweight prompting mechanisms, we conduct additional controlled ablation experiments by replacing CMPG with (1) learnable query parameters and (2) a PromptIR-style learnable prompt block~\cite{PromptIR}. The results are summarized in Table~\ref{tab:ablation_6}. Although these alternatives introduce comparable or lower computational complexity, they consistently underperform the full model equipped with CMPG. In particular, replacing semantic prompts with learnable queries yields performance close to the~{w/o CMPG} configuration and fails to recover the gains achieved by cross-modal semantic guidance.
Visual comparisons in Fig.~\ref{fig:CMPG-prompt} further show that lightweight prompting alternatives leave residual artifacts and produce less detailed restorations.}

\begin{table}[!t]
\centering
\caption{
Performance and complexity comparison of representative all-in-one image restoration methods.
PSNR is evaluated on the Snow100K-S dataset~\cite{DesnowNet}.
For DCMPC-Net, restoration-stage, PVL+Text Enc., and full pipeline costs are reported separately.
FLOPs and runtime are evaluated on 256$\times$256 input images.
}
\label{tab:ablation_2}
\begin{tabular}{lcccc}
\toprule
Method & PSNR & Params.& FLOPs & Runtime \\
\midrule
TransWeather~\cite{TransWeather} & 32.51 & 37.68M & 6.13G & 0.11s \\
Chen et al.~\cite{Chen} & 34.42 & 28.71M & 24.56G & 0.11s \\
Histoformer~\cite{Histoformer} & 37.41 & 29.92M & 168.06G & 0.05s \\
T3-DiffWeather~\cite{T3-DiffWeather} & 37.51 & 69.38M & 119.20G & 0.12s \\
GridFormer~\cite{GridFormer} & 37.46 & 33.98M & 367.40G & 0.20s \\
CyclicPrompt~\cite{CyclicPrompt} & 37.50 & 29.90M & 230.42G & 0.17s \\
\midrule
DCMPC-Net & 38.21 & 37.60M & 261.00G & 0.08s \\
PVL+Text Enc. & -- & 9.040B & 903.99G & 1.68s\\
Full pipeline & 38.21 & 9.078B & 1.165T & 1.76s\\
\bottomrule
\end{tabular}
\end{table}

%
In addition, we analyze the impact of different semantic sources used in CMPG.
Table~\ref{tab:t3} summarizes the quantitative results, showing that degradation-aware prompts constructed from LLaMA-based representations consistently outperform those derived from BLIP~\cite{BLIP2}, CLIP~\cite{clip}, or no-prompt configurations across all benchmark datasets.
These results indicate that the effectiveness of CMPG is primarily determined by the quality of semantic priors, rather than the mere presence of an auxiliary prompting mechanism.
Qualitative comparisons in Fig.~\ref{fig:p10} further show that our method recovers more realistic textures and structural details than variants without degradation-aware prompts or those using only image captions, especially in regions with complex spatially varying degradations.
By modulating decoder features with degradation-aware prompts, CMPG improves local detail preservation and structural reconstruction.

\subsubsection{Effect of DFCM}
DFCM explicitly incorporates structural information to more effectively guide the restoration process compared to prior local enhancement techniques~\cite{Restormer, Histoformer}.
We quantitatively evaluate DFCM against two baseline architectures: (1) Gated-Dconv Feed-Forward (GDFN)\cite{Restormer}; (2) Dual-scale Gated Feed-Forward (DGFF)\cite{Histoformer}, as summarized in Table~\ref{tab:t4}.
Although GDFN and DGFF improve local feature perception and boost restoration performance, their lack of explicit feature compensation constrains their effectiveness in adverse weather removal tasks. 
By integrating compensation information within the feed-forward network, DFCM attains additional performance gains.
%

Specifically, DFCM contains two branches: one preserves raw input features, while the other enhances degradation-related representations.
Ablation results in Table~\ref{tab:t5} verify the necessity of both branches.
This dual-branch design supports structural compensation and reduces feature interference.
Fig.~\ref{fig:p11} further shows that DFCM improves fine texture recovery and structural fidelity.

\subsubsection{Position of the PGAAM and DFCM}
Within the hierarchical architecture of DCMPC-Net, we investigate the optimal placement of PGAAM and DFCM in the decoder.
As shown in Table~\ref{tab:t6}, integrating PGAAM and DFCM in the encoder or across both the encoder and decoder results in degraded performance.  
In contrast, placing PGAAM and DFCM between consecutive decoder stages achieves the best performance, as decoder features preserve richer spatial and structural information.

\subsubsection{Comparison of Computational Complexity} {Table~\ref{tab:ablation_2} compares the parameters, FLOPs, and runtime of representative all-in-one image restoration methods under the same input resolution of $256 \times 256$. 
For DCMPC-Net, we separately report the restoration-stage cost, the additional cost of PVL+Text Enc., and the full pipeline cost. The DCMPC-Net cost includes all trainable restoration-stage modules, including the image encoder in CMPG, whereas PVL+Text Enc. refers only to the frozen semantic-prior extraction stage.  The results show that DCMPC-Net achieves favorable restoration performance with a competitive {restoration-stage} computational footprint, while the full pipeline introduces additional overhead from semantic-prior extraction.

\vspace{-3mm}

\subsection{Application}
To assess the practical applicability of DCMPC-Net and its impact on downstream detection, we present a representative example in Fig.~\ref{fig:p12}.
DCMPC-Net effectively removes snowflakes and improves target recognition, such as cups and televisions.
This result indicates its potential value for real-world applications, including safety-critical scenarios such as autonomous driving.




\subsection{{Discussion and Limitations}}
This study presents DCMPC-Net, which leverages pre-trained vision-language models for adverse weather removal. 
Despite its promising performance, DCMPC-Net still has limitations under extremely challenging conditions.
When dense or large rain degradations overlap with fine-scale scene structures, degradation cues may be difficult to distinguish from underlying content.
As shown in Fig.~\ref{fig:ablation_3}, this ambiguity can weaken degradation-aware prompting and subsequent feature modulation, leading to residual artifacts.
This reflects a common challenge in all-in-one image restoration, where severe degradations obscure both structural and semantic information, rather than a limitation specific to our method.
Additionally, although DCMPC-Net maintains competitive restoration-stage efficiency compared with several representative learning-based approaches, the full pipeline still introduces additional overhead due to external semantic-prior extraction. 
This is a common trade-off in recent vision-language-guided restoration methods that rely on external semantic priors~\cite{LDP,DACLIP}.
These limitations point to future directions, including more robust semantic cue extraction under extreme degradations and more efficient lightweight prompt-guided restoration frameworks.

\section{Conclusion}
In this paper, we have proposed the DCMPC-Net, a Degradation-Aware Cross-Modal Prompt Compensation Network for image restoration under adverse weather conditions. 
DCMPC-Net leverages high-level semantic information from a pre-trained vision-language model, LLaMA, to generate degradation-aware cross-modal prompts using a Cross-Modal Prompt Generator. 
These prompts are integrated into the restoration pipeline via a Prompt-Guided Attention Alignment Module for adaptive semantic alignment with degraded regions, while a Dual Feature Compensation Module further enhances structural fidelity and fine detail reconstruction.
Extensive experiments on multiple public benchmarks have demonstrated that DCMPC-Net outperforms state-of-the-art methods across a range of adverse weather removal tasks, achieving superior accuracy and perceptual quality.


\bibliographystyle{IEEEtran}
\bibliography{IEEEexample.bib}

@inproceedings{Gated_Fusion_Network,
  author       = {Wenqi Ren and
                  Lin Ma and
                  Jiawei Zhang and
                  Jinshan Pan and
                  Xiaochun Cao and
                  Wei Liu and
                  Minghsuan Yang},
  title        = {Gated Fusion Network for Single Image Dehazing},
  booktitle    = {CVPR},
  year         = {2018},
}

@article{Online_Rain/Snow_Removal,
  author       = {Minghan Li and
                  Xiangyong Cao and
                  Qian Zhao and
                  Lei Zhang and
                  Deyu Meng},
  title        = {Online Rain/Snow Removal From Surveillance Videos},
  journal      = {TIP},
  year         = {2021},
}

@inproceedings{DRSformer,
  author       = {Xiang Chen and
                  Hao Li and
                  Mingqiang Li and
                  Jinshan Pan},
  title        = {Learning {A} Sparse Transformer Network for Effective Image Deraining},
  booktitle    = {CVPR},
  year         = {2023},
}

@inproceedings{SelfPromer,
  author       = {Cong Wang and
                  Jinshan Pan and
                  Wanyu Lin and
                  Jiangxin Dong and
                  Wei Wang and
                  Xiaoming Wu},
  title        = {SelfPromer: Self-Prompt Dehazing Transformers with Depth-Consistency},
  booktitle    = {AAAI},
  year         = {2024},
}

@inproceedings{Restormer,
  author       = {Syed Waqas Zamir and
                  Aditya Arora and
                  Salman Khan and
                  Munawar Hayat and
                  Fahad Shahbaz Khan and
                  Minghsuan Yang},
  title        = {Restormer: Efficient Transformer for High-Resolution Image Restoration},
  booktitle    = {CVPR},
  year         = {2022},
}

@inproceedings{AdaIR_Adaptive,
  author       = {Yuning Cui and
                  Syed Waqas Zamir and
                  Salman H. Khan and
                  Alois Knoll and
                  Mubarak Shah and
                  Fahad Shahbaz Khan},
  title        = {AdaIR: Adaptive All-in-One Image Restoration via Frequency Mining
                  and Modulation},
  booktitle    = {ICLR},
  year         = {2025},
}

@inproceedings{PromptIR,
  author       = {Vaishnav Potlapalli and
                  Syed Waqas Zamir and
                  Salman H. Khan and
                  Fahad Shahbaz Khan},
  title        = {PromptIR: Prompting for All-in-One Image Restoration},
  booktitle    = {NIPS},
  year         = {2023},
}

@inproceedings{PromptRestorer,
  author       = {Cong Wang and
                  Jinshan Pan and
                  Wei Wang and
                  Jiangxin Dong and
                  Mengzhu Wang and
                  Yakun Ju and
                  Junyang Chen},
  title        = {PromptRestorer: {A} Prompting Image Restoration Method with Degradation
                  Perception},
  booktitle    = {NIPS},
  year         = {2023},
}

@inproceedings{clip,
  author       = {Alec Radford and
                  Jong Wook Kim and
                  Chris Hallacy and
                  Aditya Ramesh and
                  Gabriel Goh and
                  Sandhini Agarwal and
                  Girish Sastry and
                  Amanda Askell and
                  Pamela Mishkin and
                  Jack Clark and
                  Gretchen Krueger and
                  Ilya Sutskever},
  title        = {Learning Transferable Visual Models From Natural Language Supervision},
  booktitle    = {ICML},
  year         = {2021},
}

@inproceedings{Scaling-Up-to-Excellence,
  author       = {Fanghua Yu and
                  Jinjin Gu and
                  Zheyuan Li and
                  Jinfan Hu and
                  Xiangtao Kong and
                  Xintao Wang and
                  Jingwen He and
                  Yu Qiao and
                  Chao Dong},
  title        = {Scaling Up to Excellence: Practicing Model Scaling for Photo-Realistic
                  Image Restoration In the Wild},
  booktitle    = {CVPR},
  year         = {2024},
}

@article{Llama3,
  author       = {Abhimanyu Dubey and
                  Abhinav Jauhri and
                  et al.},
  title        = {The Llama 3 Herd of Models},
  journal      = {arXiv:2407.21783},
  year         = {2024},
}

@inproceedings{DACLIP,
  author       = {Ziwei Luo and
                  Fredrik K. Gustafsson and
                  Zheng Zhao and
                  Jens Sj{\"{o}}lund and
                  Thomas B. Sch{\"{o}}n},
  title        = {Controlling Vision-Language Models for Multi-Task Image Restoration},
  booktitle    = {ICLR},
  year         = {2024},
}

@article{DesnowNet,
  author       = {Yunfu Liu and
                  Dawei Jaw and
                  Shihchia Huang and
                  Jenqneng Hwang},
  title        = {DesnowNet: Context-Aware Deep Network for Snow Removal},
  journal      = {TIP},
  year         = {2018},
}

@inproceedings{AirNet,
  author       = {Boyun Li and
                  Xiao Liu and
                  Peng Hu and
                  Zhongqin Wu and
                  Jiancheng Lv and
                  Xi Peng},
  title        = {All-In-One Image Restoration for Unknown Corruption},
  booktitle    = {CVPR},
  year         = {2022},
}

@inproceedings{TransWeather,
  author       = {Jeya Maria Jose Valanarasu and
                  Rajeev Yasarla and
                  Vishal M. Patel},
  title        = {TransWeather: Transformer-based Restoration of Images Degraded by
                  Adverse Weather Conditions},
  booktitle    = {CVPR},
  year         = {2022},
}

@inproceedings{Histoformer,
  author       = {Shangquan Sun and
                  Wenqi Ren and
                  Xinwei Gao and
                  Rui Wang and
                  Xiaochun Cao},
  title        = {Restoring Images in Adverse Weather Conditions via Histogram Transformer},
  booktitle    = {ECCV},
  year         = {2024},
}

@inproceedings{BLIP2,
  author       = {Junnan Li and
                  Dongxu Li and
                  Silvio Savarese and
                  Steven C. H. Hoi},
  title        = {{BLIP-2:} Bootstrapping Language-Image Pre-training with Frozen Image
                  Encoders and Large Language Models},
  booktitle    = {ICML},
  year         = {2023},
}

@inproceedings{Language-driven,
  author       = {Hao Yang and
                  Liyuan Pan and
                  Yan Yang and
                  Wei Liang},
  title        = {Language-driven All-in-one Adverse Weather Removal},
  booktitle    = {CVPR},
  year         = {2024},
}

@inproceedings{InstructIR,
  author       = {Marcos V. Conde and
                  Gregor Geigle and
                  Radu Timofte},
  title        = {InstructIR: High-Quality Image Restoration Following Human Instructions},
  booktitle    = {ECCV},
  year         = {2024},
}

@inproceedings{WGWS,
  author       = {Yurui Zhu and
                  Tianyu Wang and
                  Xueyang Fu and
                  Xuanyu Yang and
                  Xin Guo and
                  Jifeng Dai and
                  Yu Qiao and
                  Xiaowei Hu},
  title        = {Learning Weather-General and Weather-Specific Features for Image Restoration
                  Under Multiple Adverse Weather Conditions},
  booktitle    = {CVPR},
  year         = {2023},
}

@inproceedings{Chen,
  author       = {Weiting Chen and
                  Zhikai Huang and
                  Chengche Tsai and
                  Haohsiang Yang and
                  Jianjiun Ding and
                  Sy{-}Yen Kuo},
  title        = {Learning Multiple Adverse Weather Removal via Two-stage Knowledge
                  Learning and Multi-contrastive Regularization: Toward a Unified Model},
  booktitle    = {CVPR},
  year         = {2022},
}

@inproceedings{SPANet,
  author       = {Tianyu Wang and
                  Xin Yang and
                  Ke Xu and
                  Shaozhe Chen and
                  Qiang Zhang and
                  Rynson W. H. Lau},
  title        = {Spatial Attentive Single-Image Deraining With a High Quality Real
                  Rain Dataset},
  booktitle    = {CVPR},
  year         = {2019},
}

@inproceedings{JSTASR,
  author       = {Weiting Chen and
                  Haoyu Fang and
                  Jianjiun Ding and
                  Chengche Tsai and
                  Syyen Kuo},
  title        = {{JSTASR:} Joint Size and Transparency-Aware Snow Removal Algorithm
                  Based on Modified Partial Convolution and Veiling Effect Removal},
  booktitle    = {ECCV},
  year         = {2020},
}

@inproceedings{RESCAN,
  author       = {Xia Li and
                  Jianlong Wu and
                  Zhouchen Lin and
                  Hong Liu and
                  Hongbin Zha},
  title        = {Recurrent Squeeze-and-Excitation Context Aggregation Net for Single
                  Image Deraining},
  booktitle    = {ECCV},
  year         = {2018},
}

@article{DDMSNet,
  author       = {Kaihao Zhang and
                  Rongqing Li and
                  Yanjiang Yu and
                  Wenhan Luo and
                  Changsheng Li},
  title        = {Deep Dense Multi-Scale Network for Snow Removal Using Semantic and
                  Depth Priors},
  journal      = {TIP},
  year         = {2021},
}

@inproceedings{NAPNet,
  author       = {Liangyu Chen and
                  Xiaojie Chu and
                  Xiangyu Zhang and
                  Jian Sun},
  title        = {Simple Baselines for Image Restoration},
  booktitle    = {ECCV},
  year         = {2022},
}

@inproceedings{All-in-one,
  author       = {Ruoteng Li and
                  Robby T. Tan and
                  Loongfah Cheong},
  title        = {All in One Bad Weather Removal Using Architectural Search},
  booktitle    = {CVPR},
  year         = {2020},
}

@inproceedings{AWRCP,
  author       = {Tian Ye and
                  Sixiang Chen and
                  Jinbin Bai and
                  Jun Shi and
                  Chenghao Xue and
                  Jingxia Jiang and
                  Junjie Yin and
                  Erkang Chen and
                  Yun Liu},
  title        = {Adverse Weather Removal with Codebook Priors},
  booktitle    = {CVPR},
  year         = {2023},
}

@inproceedings{CycleGAN,
  author       = {Junyan Zhu and
                  Taesung Park and
                  Phillip Isola and
                  Alexei A. Efros},
  title        = {Unpaired Image-to-Image Translation Using Cycle-Consistent Adversarial
                  Networks},
  booktitle    = {ICCV},
  year         = {2017},
}

@inproceedings{pix2pix,
  author       = {Phillip Isola and
                  Junyan Zhu and
                  Tinghui Zhou and
                  Alexei A. Efros},
  title        = {Image-to-Image Translation with Conditional Adversarial Networks},
  booktitle    = {CVPR},
  year         = {2017},
}

@article{PCNet,
  author       = {Kui Jiang and
                  Zhongyuan Wang and
                  Peng Yi and
                  Chen Chen and
                  Zheng Wang and
                  Xiao Wang and
                  Junjun Jiang and
                  Chiawen Lin},
  title        = {Rain-Free and Residue Hand-in-Hand: {A} Progressive Coupled Network
                  for Real-Time Image Deraining},
  journal      = {TIP},
  year         = {2021},}

@inproceedings{MPRNet,
  author       = {Syed Waqas Zamir and
                  Aditya Arora and
                  Salman H. Khan and
                  Munawar Hayat and
                  Fahad Shahbaz Khan and
                  Minghsuan Yang and
                  Ling Shao},
  title        = {Multi-Stage Progressive Image Restoration},
  booktitle    = {CVPR},
  year         = {2021},
}

@inproceedings{DuRN,
  author       = {Xing Liu and
                  Masanori Suganuma and
                  Zhun Sun and
                  Takayuki Okatani},
  title        = {Dual Residual Networks Leveraging the Potential of Paired Operations
                  for Image Restoration},
  booktitle    = {CVPR},
  year         = {2019},
}

@inproceedings{RainDropAttn,
  author       = {Yuhui Quan and
                  Shijie Deng and
                  Yixin Chen and
                  Hui Ji},
  title        = {Deep Learning for Seeing Through Window With Raindrops},
  booktitle    = {ICCV},
  year         = {2019},
}

@inproceedings{AttentiveGAN,
  author       = {Rui Qian and
                  Robby T. Tan and
                  Wenhan Yang and
                  Jiajun Su and
                  Jiaying Liu},
  title        = {Attentive Generative Adversarial Network for Raindrop Removal From
                  a Single Image},
  booktitle    = {CVPR},
  year         = {2018},
}

@article{IPT,
  author       = {Jie Xiao and
                  Xueyang Fu and
                  Aiping Liu and
                  Feng Wu and
                  Zhengjun Zha},
  title        = {Image De-Raining Transformer},
  journal      = {TPAMI},
  year         = {2023},
}

@inproceedings{T3-DiffWeather,
  author       = {Sixiang Chen and
                  Tian Ye and
                  Kai Zhang and
                  Zhaohu Xing and
                  Yunlong Lin and
                  Lei Zhu},
  title        = {Teaching Tailored to Talent: Adverse Weather Restoration via Prompt
                  Pool and Depth-Anything Constraint},
  booktitle    = {ECCV},
  year         = {2024},
}

@article{GridFormer,
  author       = {Tao Wang and
                  Kaihao Zhang and
                  Ziqian Shao and
                  Wenhan Luo and
                  Bj{\"{o}}rn Stenger and
                  Tong Lu and
                  Taekyun Kim and
                  Wei Liu and
                  Hongdong Li},
  title        = {GridFormer: Residual Dense Transformer with Grid Structure for Image
                  Restoration in Adverse Weather Conditions},
  journal      = {IJCV},
  year         = {2024},
}

@article{Text_Encoder,
  author       = {Liang Wang and
                  Nan Yang and
                  Xiaolong Huang and
                  Linjun Yang and
                  Rangan Majumder and
                  Furu Wei},
  title        = {Multilingual {E5} Text Embeddings: {A} Technical Report},
  journal      = {arXiv:2402.05672},
  year         = {2024},
}

@inproceedings{Residual_channel,
  author       = {Qiaosi Yi and
                  Juncheng Li and
                  Qinyan Dai and
                  Faming Fang and
                  Guixu Zhang and
                  Tieyong Zeng},
  title        = {Structure-Preserving Deraining with Residue Channel Prior Guidance},
  booktitle    = {ICCV},
  year         = {2021},
}

@inproceedings{Test1,
  author       = {Ruoteng Li and
                  Loongfah Cheong and
                  Robby T. Tan},
  title        = {Heavy Rain Image Restoration: Integrating Physics Model and Conditional
                  Adversarial Learning},
  booktitle    = {CVPR},
  year         = {2019},
}

@inproceedings{Adam,
  author       = {Diederik P. Kingma and
                  Jimmy Ba},
  title        = {Adam: {A} Method for Stochastic Optimization},
  booktitle    = {ICLR},
  year         = {2015},
}

@inproceedings{cosine,
  author       = {Ilya Loshchilov and
                  Frank Hutter},
  title        = {{SGDR:} Stochastic Gradient Descent with Warm Restarts},
  booktitle    = {ICLR},
  year         = {2017},
}

@inproceedings{MAXIM,
  author       = {Zhengzhong Tu and
                  Hossein Talebi and
                  Han Zhang and
                  Feng Yang and
                  Peyman Milanfar and
                  Alan C. Bovik and
                  Yinxiao Li},
  title        = {{MAXIM:} Multi-Axis {MLP} for Image Processing},
  booktitle    = {CVPR},
  year         = {2022},
}

@inproceedings{Textual_Removal,
  author={Lin, Jingbo and Zhang, Zhilu and Wei, Yuxiang and Ren, Dongwei and Jiang, Dongsheng and Tian, Qi and Zuo, Wangmeng},
  title        = {Improving image restoration through removing degradations in textual representations},
  booktitle    = {CVPR},
  year         = {2024},
}

@inproceedings{PPTformer,
  author       = {Cong Wang and
                  Jinshan Pan and
                  Liyan Wang and
                  Wei Wang},
  title        = {Intra and Inter Parser-Prompted Transformers for Effective Image Restoration},
  booktitle    = {AAAI},
  year         = {2025},
}

@inproceedings{GenIR,
  author       = {Yuang Ai and
                  Xiaoqiang Zhou and
                  Huaibo Huang and
                  Xiaotian Han and
                  Zhengyu Chen and
                  Quanzeng You and
                  Hongxia Yang},
  title        = {DreamClear: High-Capacity Real-World Image Restoration with Privacy-Safe
                  Dataset Curation},
  booktitle    = {NIPS},
  year         = {2024},
}

@inproceedings{MPerceiver,
  author       = {Yuang Ai and
                  Huaibo Huang and
                  Xiaoqiang Zhou and
                  Jiexiang Wang and
                  Ran He},
  title        = {Multimodal Prompt Perceiver: Empower Adaptiveness, Generalizability
                  and Fidelity for All-in-One Image Restoration},
  booktitle    = {CVPR},
  year         = {2024},
}

@inproceedings{GenLV,
  title={Learning a low-level vision generalist via visual task prompt},
  author={Chen, Xiangyu and Liu, Yihao and Pu, Yuandong and Zhang, Wenlong and Zhou, Jiantao and Qiao, Yu and Dong, Chao},
  booktitle={ACM MM},
  year={2024}
}

@article{weather_self_prompt,
  title={All-in-one Weather-degraded Image Restoration via Adaptive Degradation-aware Self-prompting Model},
  author={Wen, Yuanbo and Gao, Tao and Li, Ziqi and Zhang, Jing and Zhang, Kaihao and Chen, Ting},
  journal={TOM},
  year={2024}
}

@article{VQA1,
  title={A Global Visual Information Intervention Model for Medical Visual Question Answering},
  author={Peng, Peixi and Fan, Wanshu and Shen, Yue and Yang, Xin and Zhou, Dongsheng},
  journal={Comput. Biol. Med.},
  year={2025},
}

@article{Prompt-in-prompt,
  title={Prompt-in-prompt learning for universal image restoration},
  author={Li, Z and Lei, Y and Ma, C and Zhang, J and Shan, H},
  journal={arXiv:2312.05038},
  year={2023}
}

@article{TextualIR,
  title={Textual prompt guided image restoration},
  author={Yan, Qiuhai and Jiang, Aiwen and Chen, Kang and Peng, Long and Yi, Qiaosi and Zhang, Chunjie},
  journal={EAAI},
  year={2025},
}

@article{SSP-IR,
  title={MRIR: Integrating Multimodal Insights for Diffusion-based Realistic Image Restoration},
  author={Zhang, Yuhong and Zhang, Hengsheng and Chai, Xinning and Xie, Rong and Song, Li and Zhang, Wenjun},
  journal={TCSVT},
  year={2025}
}

@inproceedings{VPT,
  title={Visual prompt tuning},
  author={Jia, Menglin and Tang, Luming and Chen, Bor-Chun and Cardie, Claire and Belongie, Serge and Hariharan, Bharath and Lim, Ser-Nam},
  booktitle={ECCV},
  year={2022},
}

@inproceedings{GPP_LLIE,
  title={Low-light image enhancement via generative perceptual priors},
  author={Zhou, Han and Dong, Wei and Liu, Xiaohong and Zhang, Yulun and Zhai, Guangtao and Chen, Jun},
  booktitle={AAAI},
  year={2025}
}

@article{NLP_prompt,
  title={Improving natural language processing tasks with human gaze-guided neural attention},
  author={Sood, Ekta and Tannert, Simon and M{\"u}ller, Philipp and Bulling, Andreas},
  journal={NIPS},
  year={2020},
}

@article{AIRformer,
  title={Frequency-oriented efficient transformer for all-in-one weather-degraded image restoration},
  author={Gao, Tao and Wen, Yuanbo and Zhang, Kaihao and Zhang, Jing and Chen, Ting and Liu, Lidong and Luo, Wenhan},
  journal={TCSVT},
  year={2023},
}

@inproceedings{Bert,
  title={Bert: Pre-training of deep bidirectional transformers for language understanding},
  author={Devlin, Jacob and Chang, Ming-Wei and Lee, Kenton and Toutanova, Kristina},
  booktitle={ACL},
  year={2019}
}

@article{GPT,
  title={Improving language understanding by generative pre-training},
  author={Radford, Alec and Narasimhan, Karthik and Salimans, Tim and Sutskever, Ilya and others},
  year={2018},
}

@article{NLP1,
  title={Exploiting cloze questions for few shot text classification and natural language inference},
  author={Schick, Timo and Sch{\"u}tze, Hinrich},
  journal={arXiv:2001.07676},
  year={2020}
}

@article{NLP2,
  title={Autoprompt: Eliciting knowledge from language models with automatically generated prompts},
  author={Shin, Taylor and Razeghi, Yasaman and Logan IV, Robert L and Wallace, Eric and Singh, Sameer},
  journal={arXiv:2010.15980},
  year={2020}
}

@article{MAE-VQGAN,
  title={Visual prompting via image inpainting},
  author={Bar, Amir and Gandelsman, Yossi and Darrell, Trevor and Globerson, Amir and Efros, Alexei},
  journal={NIPS},
  year={2022}
}

@inproceedings{caption,
  title={Deep visual-semantic alignments for generating image descriptions},
  author={Karpathy, Andrej and Fei-Fei, Li},
  booktitle={CVPR},
  year={2015}
}

@article{SSIM,
  title={Image quality assessment: from error visibility to structural similarity},
  author={Wang, Zhou and Bovik, Alan C and Sheikh, Hamid R and Simoncelli, Eero P},
  journal={TIP},
  year={2004},
}

@article{PSNR,
  title={Scope of validity of PSNR in image/video quality assessment},
  author={Huynh-Thu, Quan and Ghanbari, Mohammed},
  journal={Electronics letters},
  year={2008},
}

@article{layernorm,
  author       = {Lei Jimmy Ba and
                  Jamie Ryan Kiros and
                  Geoffrey E. Hinton},
  journal      = {arXiv:1607.06450},
  year         = {2016},
}

@article{pearson,
  title={Pearson correlation coefficient},
  author={Cohen, Israel and Huang, Yiteng and Chen, Jingdong and Benesty, Jacob and Benesty, Jacob and Chen, Jingdong and Huang, Yiteng and Cohen, Israel},
  journal={Noise reduction in speech processing},
  year={2009},
}

@inproceedings{FFTloss,
  title={Rethinking coarse-to-fine approach in single image deblurring},
  author={Cho, Sung-Jin and Ji, Seo-Won and Hong, Jun-Pyo and Jung, Seung-Won and Ko, Sung-Jea},
  booktitle={ICCV},
  year={2021}
}

@inproceedings{MOERL,
  title={MOERL: When Mixture-of-Experts Meet Reinforcement Learning for Adverse Weather Image Restoration},
  author={Wang, Tao and Xia, Peiwen and Li, Bo and Jiang, Peng-Tao and Kong, Zhe and Zhang, Kaihao and Lu, Tong and Luo, Wenhan},
  booktitle={ICCV},
  year={2025}
}

@article{CyclicPrompt,
  author       = {Rongxin Liao and
                  Feng Li and
                  Yanyan Wei and
                  Zenglin Shi and
                  Le Zhang and
                  Huihui Bai and
                  Meng Wang},
  title        = {Prompt to Restore, Restore to Prompt: Cyclic Prompting for Universal Adverse Weather Removal},
  journal      = {TIP},
  year         = {2025},
}

@inproceedings{dcsfn,
  title={Dcsfn: Deep cross-scale fusion network for single image rain removal},
  author={Wang, Cong and Xing, Xiaoying and Wu, Yutong and Su, Zhixun and Chen, Junyang},
  booktitle={ACM MM},
  pages={1643--1651},
  year={2020}
}

@inproceedings{jdnet,
  author       = {Cong Wang and
                  Yutong Wu and
                  Zhixun Su and
                  Junyang Chen},
  title        = {Joint Self-Attention and Scale-Aggregation for Self-Calibrated Deraining
                  Network},
  booktitle    = {ACM MM},
  pages        = {2517--2525},
  year         = {2020}
}

@inproceedings{online_derain_aaai22,
  author       = {Cong Wang and
                  Jinshan Pan and
                  Xiao{-}Ming Wu},
  title        = {Online-Updated High-Order Collaborative Networks for Single Image
                  Deraining},
  booktitle    = {AAAI},
  pages        = {2406--2413},
  year         = {2022}
}

@inproceedings{wang_acmmm24_derain,
  author       = {Cong Wang and
                  Liyan Wang and
                  Jie Mu and
                  Chengjin Yu and
                  Wei Wang},
  title        = {Progressive Local and Non-Local Interactive Networks with Deeply Discriminative
                  Training for Image Deraining},
  booktitle    = {ACM MM},
  pages        = {10326--10335},
  year         = {2024}
}

@inproceedings{wang_gragh_derain_ijcai24,
  author       = {Cong Wang and
                  Wei Wang and
                  Chengjin Yu and
                  Jie Mu},
  title        = {Explore Internal and External Similarity for Single Image Deraining
                  with Graph Neural Networks},
  booktitle    = {IJCAI},
  pages        = {1371--1379},
  year         = {2024}
}

@inproceedings{uhdformer,
  author       = {Cong Wang and
                  Jinshan Pan and
                  Wei Wang and
                  Gang Fu and
                  Siyuan Liang and
                  Mengzhu Wang and
                  Xiao{-}Ming Wu and
                  Jun Liu},
  title        = {Correlation Matching Transformation Transformers for {UHD} Image Restoration},
  booktitle    = {AAAI},
  pages        = {5336--5344},
  year         = {2024}
}

@article{wang2026neural,
  title={Neural Discrimination-Prompted Transformers for Efficient UHD Image Restoration and Enhancement},
  author={Wang, Cong and Pan, Jinshan and Wang, Liyan and Wang, Wei and Yang, Yang},
  journal={IJCV},
  volume={134},
  number={3},
  pages={84},
  year={2026}
}

@inproceedings{LDP,
  author       = {Hao Yang and
                  Liyuan Pan and
                  Yan Yang and
                  Richard I. Hartley and
                  Miaomiao Liu},
  title        = {{LDP:} Language-driven Dual-Pixel Image Defocus Deblurring Network},
  booktitle    = {CVPR},
  year         = {2024},
}

\vspace{-5mm}
\begin{IEEEbiography}
[{\includegraphics[width=1in,height=1.25in,clip,keepaspectratio]{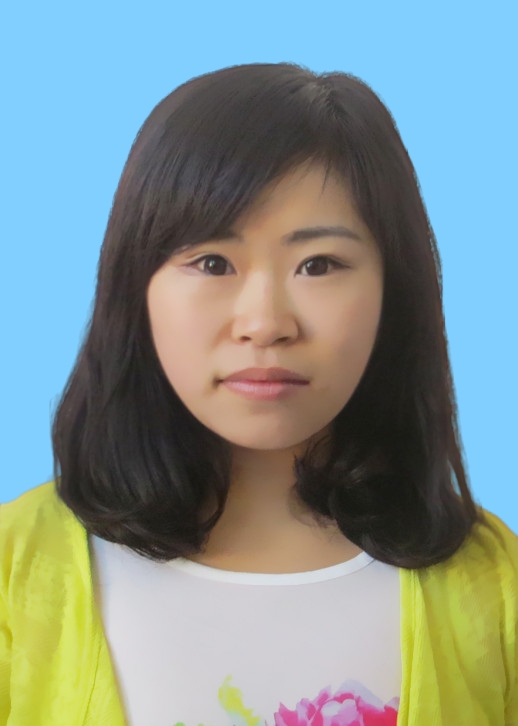}}]{Wanshu Fan} (Member, IEEE) was born in Heilongjiang, China. She received the Ph.D. degree in computational mathematics from Dalian University of Technology in 2020. Currently, she is working in the School of Software Engineering at Dalian University. Her research interests include computer vision and deep learning. She is a member of IEEE and CCF.
\end{IEEEbiography}

\vspace{-5mm}
\begin{IEEEbiography}[{\includegraphics[width=1in,height=1.25in,clip,keepaspectratio]{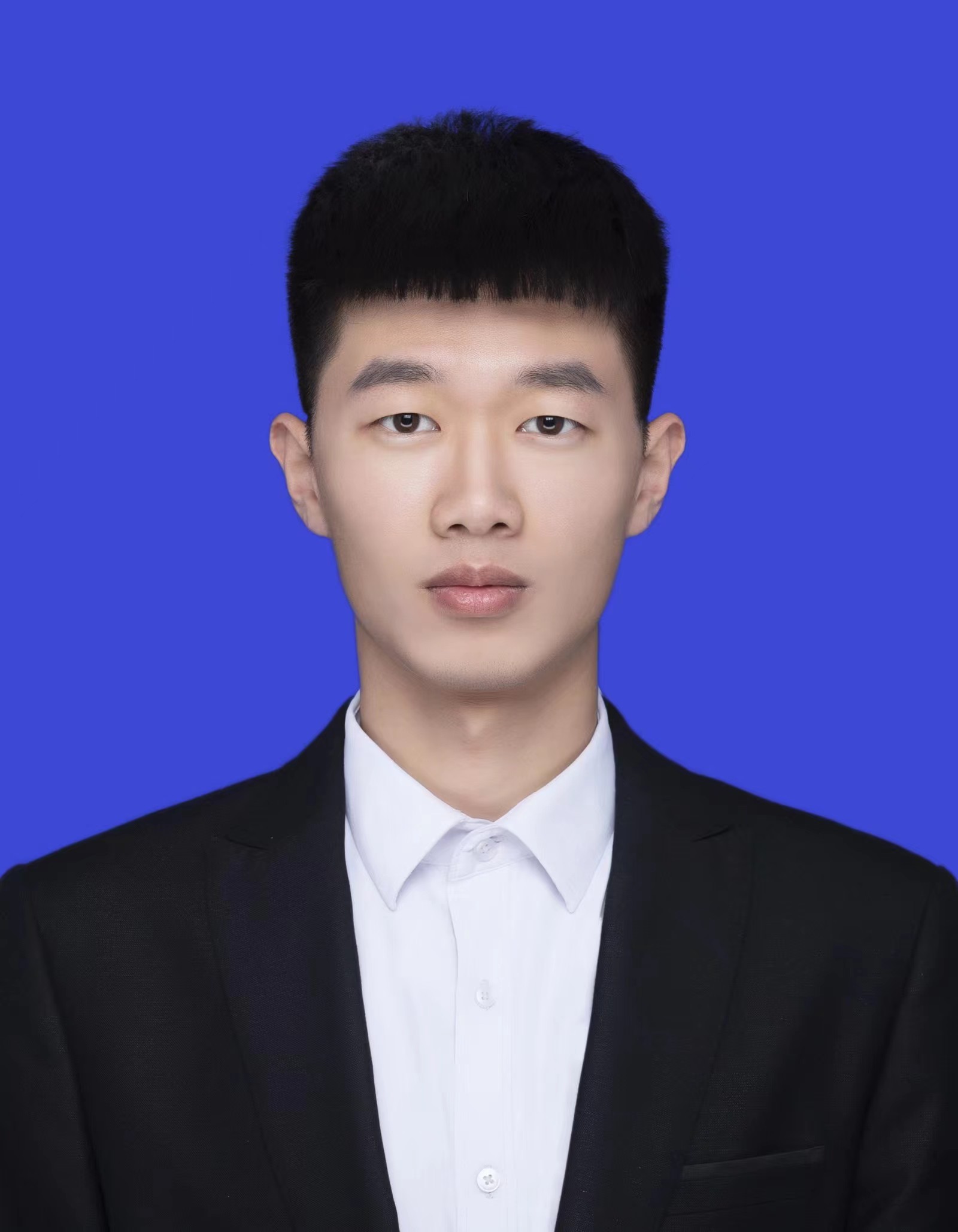}}]{Yunzhe Zhang} Yunzhe Zhang was born in Changchun, China. He received his B.S. degree in information and computing science from Changchun University of Technology in 2022. Now he is pursuing software engineering at Dalian University and is working hard to pursue a master's degree. His research interests include deep learning and computer vision.
\end{IEEEbiography}

\vspace{-5mm}
\begin{IEEEbiography}[{\includegraphics[width=1in,height=1.25in,clip,keepaspectratio]{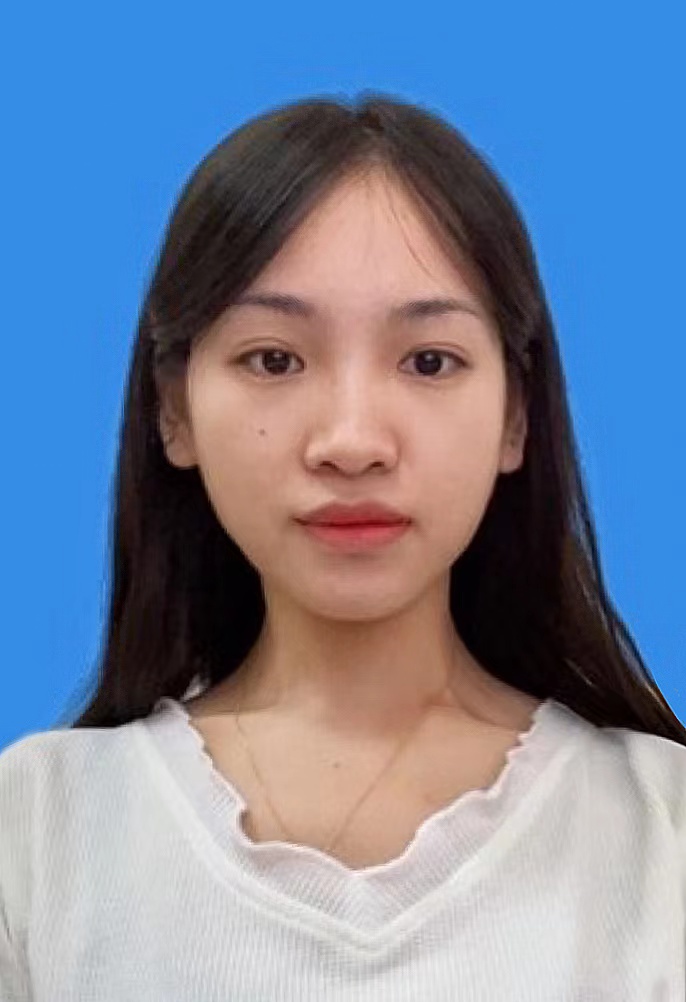}}]{Yue Shen} was born in Chongqing, China. She received her B.S. degree in software engineering from Chongqing University of Arts and Sciences in 2019. Now she is pursuing software engineering at Dalian University and is working hard to pursue a master's degree. Her research interests include deep learning and computer vision.
\end{IEEEbiography}

\begin{IEEEbiography}[{\includegraphics[width=1in,height=1.25in,clip,keepaspectratio]{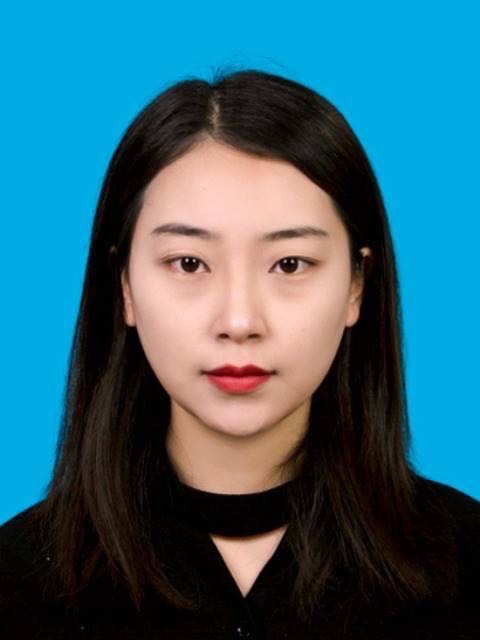}}]{Liyan Wang} is currently a Ph.D. student at the School of Mathematical Sciences, Dalian University of Technology. She received the Master’s degree and the Bachelor’s Degree in computer science and technology from the School of Computer and Information Technology, Liaoning Normal University, Dalian,  China. Her research interests include computer vision and deep learning.
\end{IEEEbiography}

\begin{IEEEbiography}[{\includegraphics[width=1in,height=1.25in,clip,keepaspectratio]{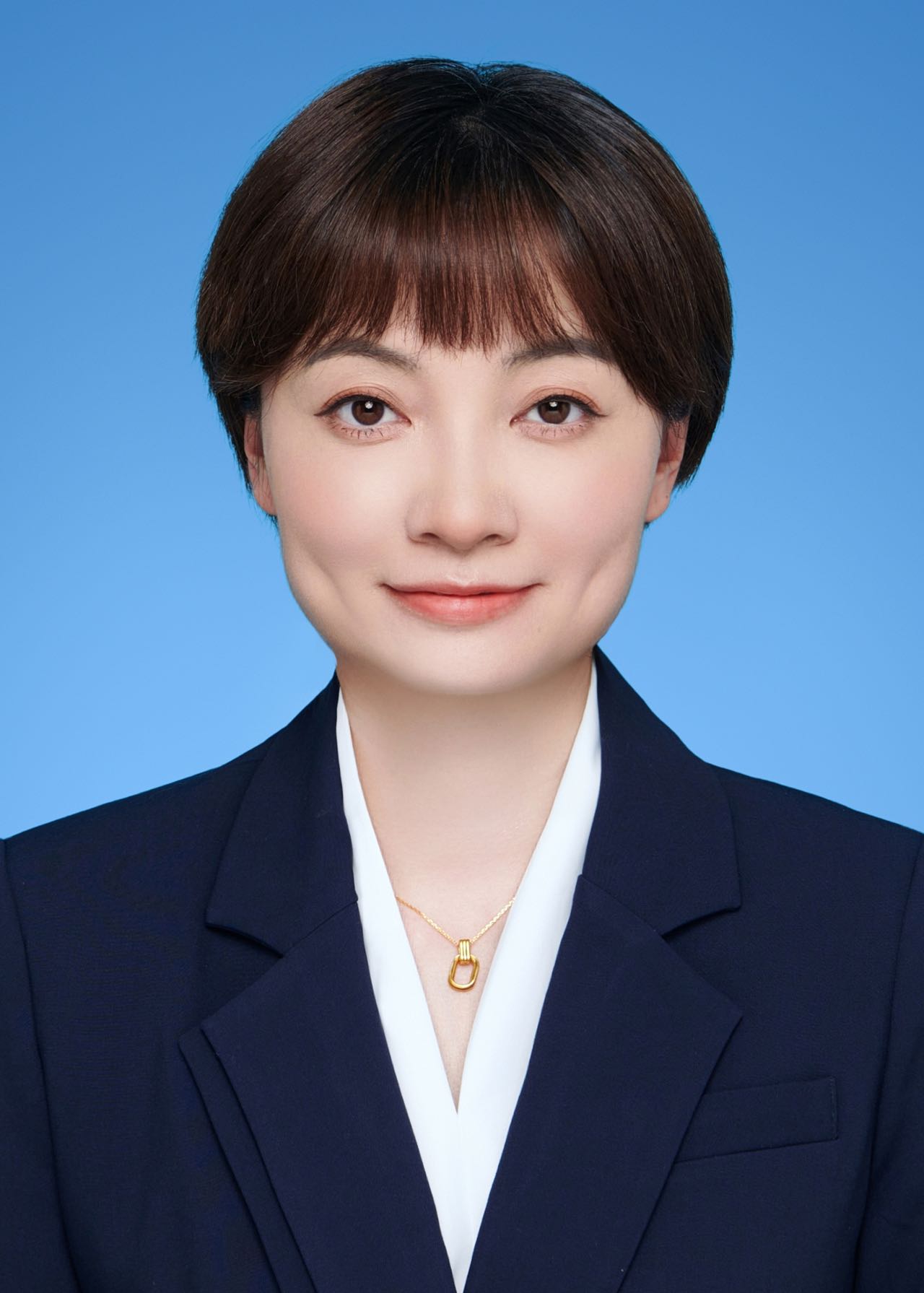}}]{Jing Qin} received the PhD degree in the School of Computer Science and Technology, Dalian University of Technology, Dalian, China, in 2018. She is an associate professor in the School of Software Technology, Dalian University, Dalian, China.  Her research interests lie in Smart Healthcare and Healthcare Information Management, with a focus on AI-driven medical solutions, data analytics, and machine learning.
\end{IEEEbiography}

\vspace{-5mm}
\begin{IEEEbiography} [{\includegraphics[width=1in,height=1.25in,clip,keepaspectratio]{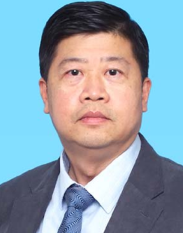}}]{Kin-Man Lam } (Senior Member, IEEE) received the associateship in electronic engineering with distinction from Hong Kong Polytechnic University, in 1986, the MSc degree in communication engineering from the Department of Electrical Engineering, Imperial College, U.K., in 1987, and the PhD degree from the Department of Electrical Engineering, University of Sydney, Australia, in 1996. From 1990 to 1993, he was a lecturer with the Department of Electronic Engineering, Hong Kong Polytechnic University. He joined the Department of Electronic and Information Engineering, The Hong Kong Polytechnic University again as an assistant professor in 1996. He became an associate professor in 1999 and has been a professor since 2010. Currently, he is also an associate dean with the Faculty of Engineering. He was actively involved in professional activities. He was the Chairman of the IEEE Hong Kong Chapter of Signal Processing between 2006 and 2008, and was the
Director-Student Services and the Director-Membership Services of the IEEE SPS between 2012 and 2014, and between 2015 and 2017, respectively. He was also the VP-Member Relations and Development and VP-Publications of the Asia-Paciﬁc Signal and Information Processing Association (APSIPA) between 2014 and 2017, and between 2017 and 2021, respectively. He was an associate editor of the IEEE Transactions on Image Processing between 2009 and 2014, and Digital Signal Processing between 2014 and 2018. He was an Editor of HKIE Transactions between 2013 and 2018, and an Area Editor of the IEEE Signal Processing Magazine between 2015 and 2017. Currently, he is the
IEEE SPS VP-Membership and the Member-at-Large of APSIPA. Prof. Lam also serves as a Senior Editorial Board member of APSIPA Trans. on Signal and Information Processing and an Associate Editor of EURASIP International Journal on Image and Video Processing. His current research interests include image and video processing, computer vision, and human face analysis and recognition.
\end{IEEEbiography}

\begin{IEEEbiography}[{\includegraphics[width=1in,height=1.25in,clip,keepaspectratio]{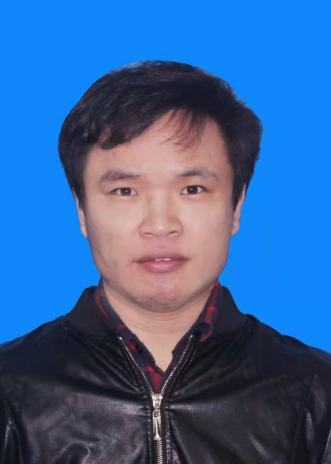}}]{Cong Wang} received the Ph.D. degree from the Department of Computing, The Hong Kong Polytechnic University. He is currently a Postdoctoral Fellow at the Department of Radiology and Biomedical Imaging, University of California, San Francisco, CA, USA. His research interests include computer vision, deep learning, and AI for healthcare.
\end{IEEEbiography}

\begin{IEEEbiography}[{\includegraphics[width=1in,height=1.25in,clip,keepaspectratio]{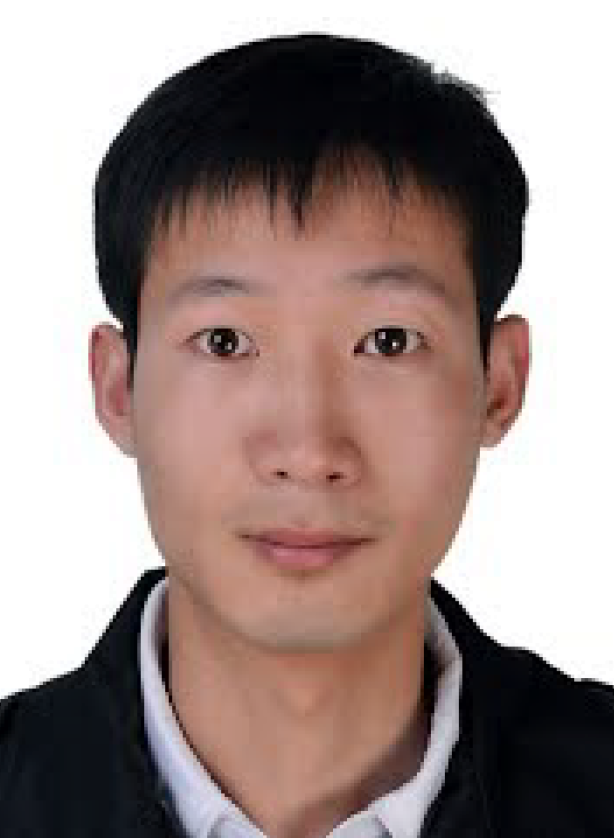}}]{Jinshan Pan} (Senior Member, IEEE) is a professor of the School of Computer Science and Engineering, Nanjing University of Science and Technology. He received the Ph.D. degree in computational mathematics from Dalian University of Technology, China, in 2017. He was a joint training Ph.D. student in Electrical Engineering and Computer Science at the University of California, Merced, from 2014 to 2016. His research interest includes image analysis and enhancement, and related vision problems.
\end{IEEEbiography}

\end{document}